\documentclass{article}
\usepackage{iclr2027_conference,times}
\usepackage{graphicx}
\usepackage{hyperref}
\usepackage{url}

\usepackage{amsmath}
\usepackage{amssymb}
\usepackage{booktabs}
\usepackage{makecell}
\usepackage{multirow}
\usepackage{tabularx}
\usepackage{longtable}
\usepackage{listings}
\usepackage[most]{tcolorbox}

\lstdefinestyle{artifactlisting}{
  basicstyle=\ttfamily\footnotesize,
  breaklines=true,
  breakatwhitespace=false,
  columns=fullflexible,
  keepspaces=true,
  showstringspaces=false,
  numbers=none,
  frame=none,
  tabsize=2
}

\newtcblisting{artifactbox}[1][]{
  enhanced,
  listing only,
  listing engine=listings,
  listing options={style=artifactlisting},
  colback=white,
  colframe=black,
  colbacktitle=black!5,
  coltitle=black,
  fonttitle=\bfseries\small,
  title={#1},
  boxrule=0.7pt,
  arc=0pt,
  left=5pt,
  right=5pt,
  top=5pt,
  bottom=5pt,
  width=\linewidth
}

\newcommand{\apptrajframefour}[3]{%
\begin{minipage}[t]{0.235\textwidth}
\vspace{0pt}
\centering
\includegraphics[width=\linewidth]{#1}\\[-0.2em]
{\scriptsize \textbf{Step #2.} #3}
\end{minipage}%
}

\title{
Embodied-BenchForge:
A Closed-Loop Agentic Workflow for Embodied Benchmark Construction
}

\author{
Baoyang Jiang$^1$, Fengchun Zhang$^2$, Leyuan Wang$^3$,
Haotian Li$^3$, Yida Wang$^3$, Zhe Ji$^4$ \\
Jinshan Lai$^2$, Xi Ren$^5$, Danyang Li$^6$, Zheng Yang$^6$,
Jianwei Hu$^1$, Qiang Ma$^1$\thanks{Corresponding author.} \\
$^1$QiYuan Lab \\
$^2$University of Electronic Science and Technology of China \\
$^3$Beijing University of Posts and Telecommunications \\
$^4$Northeastern University \quad
$^5$Beihang University \quad
$^6$Tsinghua University
}

 \iclrfinalcopy
\begin{document}

\maketitle
\lhead{}

\begin{abstract}
Agentic systems offer a promising way to automate embodied benchmark
construction, but existing approaches typically cover isolated stages
or remain specialized to predefined environments and task families.
More importantly, multi-step construction produces dependent
intermediate artifacts that are often passed downstream without
artifact-specific verification, allowing local defects to propagate
into the final benchmark.
We present \textbf{Embodied-BenchForge}, an agentic framework that
transforms user-specified evaluation intents into complete embodied
benchmark artifacts.
It formulates construction as
\textbf{Closed-Loop Benchmark Synthesis}, integrating forward artifact
synthesis with backward verification and repair.
\textbf{Skill-Orchestrated Artifact Synthesis} composes typed and
reusable skills into executable workflows, while an artifact dependency
graph records intermediate outputs and their dependencies.
\textbf{Requirement-Guided Verification and Repair} applies
artifact-specific contracts throughout construction and uses provenance
to trigger local re-execution or upstream rollback when verification
fails.
Embodied-BenchForge constructs six benchmarks covering diverse embodied
scenarios in the \textit{Offline EQA Track}, together with one
interactive benchmark containing 220 executable tasks in the
\textit{Interactive Embodied Track}.
Evaluations of representative MLLMs and embodied agents show that the benchmarks distinguish model capabilities in both
observation-based understanding and closed-loop execution.
Quality assessment and ablations validate benchmark quality and the
effectiveness of verification and repair, while repair and skill-reuse
analyses demonstrate efficient localized recovery and cross-benchmark
reusability.
\end{abstract}

\section{Introduction}

Recent advances in multimodal large language models (MLLMs),
vision--language--action models, and embodied agents are extending
embodied intelligence from passive perception to spatial reasoning,
planning, and closed-loop interaction
~\cite{embodied_vlm_survey,vla_survey,duan2022survey}.
Accordingly, embodied benchmarks now cover question answering,
navigation, household activities, and robotic manipulation
~\cite{das2018embodied,r2r,alfred,behavior1k,calvin,libero,openeqa,
vlabench}.
Reliable evaluation is essential for identifying model capability
boundaries, yet benchmark construction remains labor-intensive.
Most existing benchmarks rely on substantial manual design, while
procedural and model-assisted methods typically automate only selected
operations, such as scene generation, task synthesis, demonstration
collection, or evaluation.

Agentic systems offer a promising direction for automating embodied
benchmark construction, but two challenges remain.
First, existing automated procedures are commonly specialized to
predefined environments, resources, or task families
~\cite{procthor,mimicgen,gensim,gensim2,robogen,holodeck}.
Embodied benchmarks may consume images, videos, trajectories,
environment states, simulator interfaces, and real-world observations,
while producing grounded questions, reference answers, task
instructions, scoring protocols, and executable verifiers
~\cite{ego4d,embodiedscan,openxembodiment,droid,habitat,carla}.
Transforming a new evaluation intent into these heterogeneous yet
consistent artifacts therefore still requires considerable manual
configuration.

Second, agentic construction introduces a
\textbf{process-level reliability problem}.
Benchmark artifacts are generated through dependent operations,
including source processing, evidence grounding, state structuring,
task synthesis, answer derivation, and verifier construction.
Without artifact-specific verification, defects in intermediate outputs
may propagate into downstream artifacts.
Final-stage inspection can remove invalid items, but delayed validation
wastes computation, while complete regeneration may discard already
verified artifacts.
Reliable automation therefore requires both end-to-end synthesis and
process-level verification with selective reconstruction.

To address these challenges, we present
\textbf{Embodied-BenchForge}, an agentic framework that transforms
user-specified evaluation intents into complete embodied benchmark
artifacts.
Embodied-BenchForge formulates construction as
\textbf{Closed-Loop Benchmark Synthesis}, integrating forward artifact
synthesis with backward verification and repair.
\textbf{Skill-Orchestrated Artifact Synthesis} represents construction
operations as typed and reusable skills and composes them into executable
workflows according to the evaluation intent, available resources, and
target artifacts.
A typed artifact dependency graph records intermediate outputs and their
dependencies, enabling heterogeneous workflows to share a common
construction backbone.
\textbf{Requirement-Guided Verification and Repair} associates each
artifact type with structural, grounding, semantic, and execution
requirements.
When verification fails, recorded provenance triggers local
re-execution or upstream rollback, after which only the affected
downstream artifacts are reconstructed.

Using Embodied-BenchForge, we construct two evaluation tracks.
The \textit{Offline EQA Track (OE-Track)} contains six benchmarks
covering household robots, mobile robots, autonomous vehicles, robotic
arms, UAVs, and quadruped robots.
The \textit{Interactive Embodied Track (IE-Track)} contains one
interactive benchmark with 220 executable tasks, each providing a
restorable initial state, action interface, goal condition, and
terminal-state verifier, while retaining trajectories for
trajectory-grounded question answering.
Evaluations of representative MLLMs and VLM-based embodied agents show
that the resulting benchmarks distinguish model capabilities and reveal
a substantial gap between observation-based understanding and
closed-loop task execution.
Artifact-quality assessment, component ablations, repair analysis,
construction-efficiency measurements, and skill-reuse statistics further
evaluate the reliability and reusability of the framework.

Our contributions are summarized as follows:
\begin{itemize}
    \item We present \textbf{Embodied-BenchForge}, an agentic framework
    for transforming user-specified evaluation intents into complete
    embodied benchmark artifacts across heterogeneous resources and
    observation-based or interactive task settings.

    \item We formulate \textbf{Closed-Loop Benchmark Synthesis}, which
    combines typed, skill-orchestrated artifact synthesis with
    artifact-specific requirement contracts and provenance-guided
    re-execution or rollback.

 \item We construct an Offline EQA Track containing six embodied
benchmarks and an Interactive Embodied Track, and demonstrate
that they provide reliable and diagnostically meaningful
evaluation signals through systematic analyses of artifact quality,
construction efficiency, skill reuse, repair behavior, and model
performance.
\end{itemize}

\begin{table}[h]
	\centering
	\footnotesize
	\setlength{\tabcolsep}{2.4pt}
	\renewcommand{\arraystretch}{1.08}
	
	\begin{tabular}{@{}lccccc@{}}
		\toprule
		\textbf{Method}
		& \textbf{\makecell{End-to-End\\Synthesis}}
		& \textbf{\makecell{Reusable \\Modules}}
		& \textbf{\makecell{Heterogeneous\\Resources}}
		& \textbf{\makecell{Executable\\Artifacts}}
		& \textbf{\makecell{Process-Level \\Verification and Repair}} \\
		\midrule
		
		\multicolumn{6}{@{}l}{\textit{General and domain-specific benchmark construction}} \\
		\midrule
		
		AutoBencher~\citeyearpar{autobencher}
		& $\checkmark$
		& --
		& --
		& --
		& $\triangle$ \\
		
		BenchAgents~\citeyearpar{benchagents}
		& $\checkmark$
		& $\triangle$
		& $\triangle$
		& --
		& $\triangle$ \\
		
		Code2Bench~\citeyearpar{code2bench}
		& $\checkmark$
		& $\triangle$
		& --
		& $\checkmark$
		& $\triangle$ \\
		
		\midrule
		\multicolumn{6}{@{}l}{\textit{Embodied task generation and evaluation}} \\
		\midrule
		
		RoboGen~\citeyearpar{robogen}
		& $\triangle$
		& $\triangle$
		& --
		& $\checkmark$
		& $\triangle$ \\
		
		A2Eval~\citeyearpar{a2eval}
		& $\triangle$
		& $\triangle$
		& $\triangle$
		& $\triangle$
		& -- \\
		
		\midrule
		
		\textbf{Embodied-BenchForge}
		& $\checkmark$
		& $\checkmark$
		& $\checkmark$
		& $\checkmark$
		& $\checkmark$ \\
		
		\bottomrule
	\end{tabular}
	
	\caption{
		Comparison of automated benchmark-construction and embodied
		task-generation systems.
		Columns indicate support for end-to-end synthesis, reusable modules,
		heterogeneous resources, executable artifacts, and process-level
		verification and repair.
		$\checkmark$, $\triangle$, and -- denote explicit, partial, and no or
		out-of-scope support, respectively.
		In the final column, $\triangle$ indicates verification without selective
		artifact reconstruction.
		The comparison focuses on mechanisms rather than performance.
	}
	\label{tab:related_systems}
\end{table}

\section{Related Work}
\label{sec:related_work}

\paragraph{Embodied Benchmarks and Task Generation.}
Existing embodied benchmarks cover scene understanding, navigation,
manipulation, planning, and interactive behavior.
Representative examples include ALFRED, LIBERO, EmbodiedScan, and
EmbodiedBench~\citep{alfred,libero,embodiedscan,embodiedbench}, while
OpenEQA evaluates question answering over recorded experience and active
exploration~\citep{openeqa}.
Although these benchmarks provide valuable task definitions and
evaluation protocols, their construction is usually tied to specific
resources, environments, embodiments, and artifact schemas.
Related work automates individual components of benchmark production.
ProcTHOR and RoboCasa generate scenes and environments, MimicGen expands
robot demonstrations, and Eureka and RoboGen use language models to
design rewards, tasks, or environments
~\citep{procthor,robocasa,mimicgen,eureka,robogen}.
These methods reduce data- and task-generation costs, but generally
operate within predefined simulators or task families rather than
constructing complete benchmark packages from user-specified evaluation
intents.

\paragraph{Automated Benchmark Construction and Verification.}
Automated benchmark construction has also been explored in language,
multimodal, and software evaluation.
Dynabench, BenchBuilder, and AutoBencher generate or refresh evaluation
data through human--model interaction, declarative requirements, search,
and quality filtering
~\citep{dynabench,benchbuilder,autobencher}.
BenchAgents organizes construction into staged agent workflows,
BenchBench introduces psychometric analysis, and Code2Bench combines
software-task generation with dependency and coverage checks
~\citep{benchagents,benchbench,code2bench}.
In the embodied domain, A2Eval selects capability-balanced subsets from
existing benchmarks rather than constructing new grounded and executable
tasks~\citep{a2eval}.
Embodied benchmark construction additionally requires consistency among
observations, privileged states, grounded evidence, feasible actions,
goals, and executable verifiers.
Embodied-BenchForge addresses this setting through reusable construction
skills, artifact-specific verification, and provenance-guided
re-execution or rollback.
Table~\ref{tab:related_systems} summarizes the differences in scope and
construction mechanisms.

\begin{figure*}[t]
	\centering
	\includegraphics[width=0.7\textwidth]{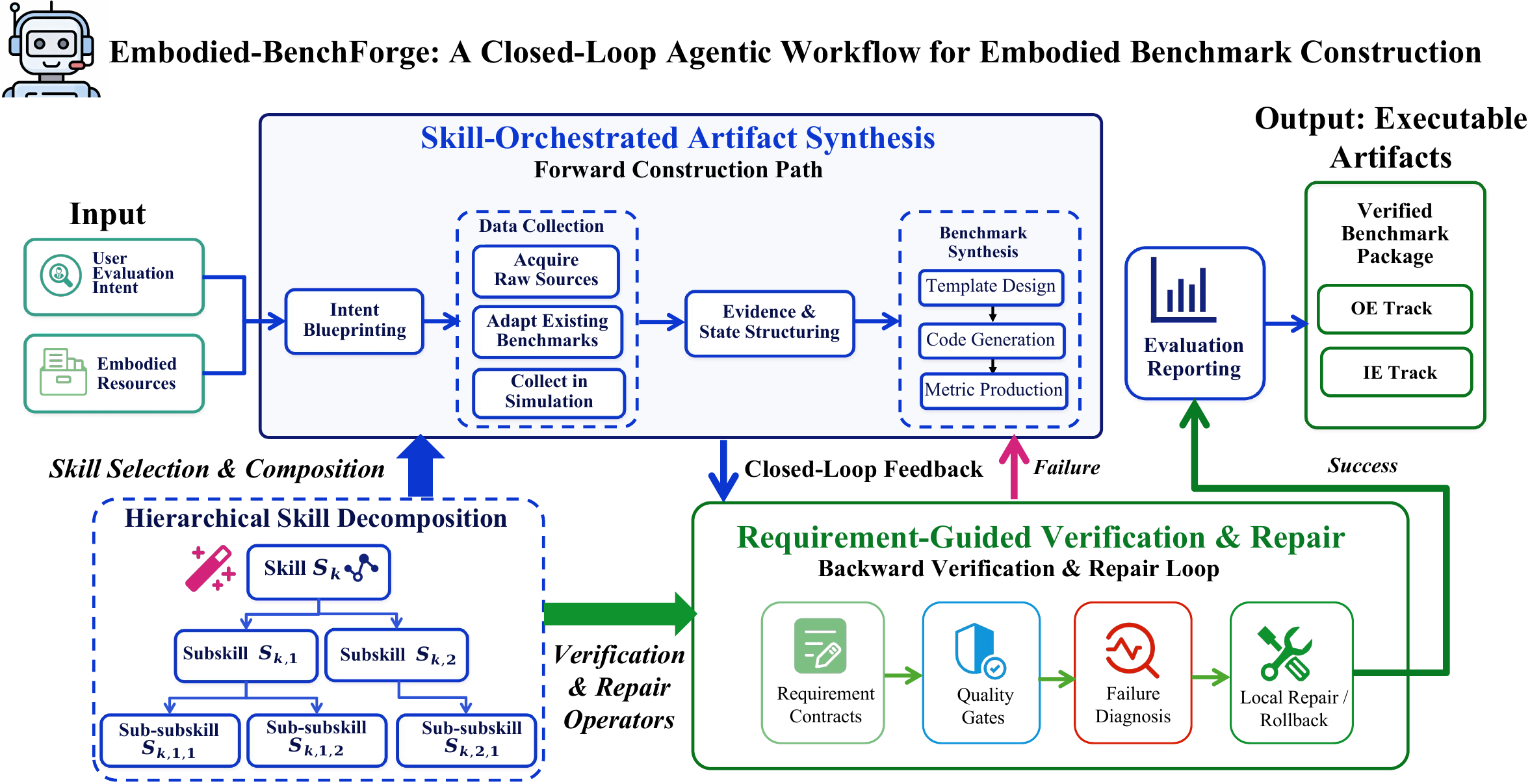}
	\caption{
		Overview of \textbf{Embodied-BenchForge} as a closed-loop agentic workflow for
		embodied benchmark construction.
		\textbf{Skill-Orchestrated Artifact Synthesis} selects and composes
		hierarchical skills to transform evaluation intents and embodied
		resources into structured benchmark artifacts.
		\textbf{Requirement-Guided Verification and Repair} applies requirement
		contracts and quality gates, traces failures through provenance, and
		triggers local repair, selective re-execution, or rollback.
		Verified artifacts proceed to evaluation reporting and are compiled into
		the Offline EQA and Interactive Embodied tracks, while failed artifacts
		are returned to the affected construction path.
	}
	\label{fig:overview0727}
\end{figure*}

\section{Embodied-BenchForge}
\label{sec:method}

\paragraph{Overview.}
Embodied-BenchForge instantiates
\textbf{Closed-Loop Benchmark Synthesis} as a coupled process of forward
construction and backward verification and repair.
As shown in Figure~\ref{fig:overview0727}, the framework receives a
\textit{User Evaluation Intent} and a set of \textit{Embodied Resources}
as input.
A hierarchical skill library supplies both
\textit{Skill Selection \& Composition} for the forward construction
path and \textit{Verification \& Repair Operators} for the backward
loop.
The forward path,
\textbf{Skill-Orchestrated Artifact Synthesis}, proceeds through
\textit{Intent Blueprinting}, \textit{Data Collection},
\textit{Evidence \& State Structuring}, \textit{Benchmark Synthesis},
and \textit{Evaluation Reporting}.
The artifacts enter
\textbf{Requirement-Guided Verification and Repair} through
\textit{Closed-Loop Feedback}.
Requirement contracts and quality gates determine whether an artifact
can proceed.
Successful artifacts are returned to \textit{Evaluation Reporting} and
compiled into a \textit{Verified Benchmark Package} containing the
OE-Track and IE-Track.
Failed artifacts enter \textit{Failure Diagnosis}, followed by
\textit{Local Repair / Rollback} and reconstruction of the affected
path.

\paragraph{Skill-Orchestrated Artifact Synthesis.}
Embodied benchmark construction produces multiple dependent artifacts
rather than directly converting raw resources into final benchmark
items.
We represent each artifact as
\begin{equation}
a_i =
\left\langle
\tau_i, x_i, m_i, p_i
\right\rangle,
\end{equation}
where $\tau_i$ is the artifact type, $x_i$ is its content, $m_i$
contains its metadata and verification state, and $p_i$ records its
construction provenance.
The principal artifact types correspond to the forward construction
path, including intent blueprints, source records, evidence or
environment-state records, benchmark items, scoring or verification
artifacts, and evaluation reports.
Explicit typing makes their dependencies inspectable and prevents
incompatible artifacts from being connected.

Construction operations are encapsulated as typed skills:
\begin{equation}
s_k =
\left\langle
I_k, O_k, P_k, F_k
\right\rangle,
\end{equation}
where $I_k$ and $O_k$ denote the input and output artifact types,
$P_k$ specifies execution preconditions, and $F_k$ is the executable
backend.
A backend may be a model, deterministic program, simulator interface,
resource adapter, or scoring and verification compiler.
Through \textit{Hierarchical Skill Decomposition}, a complex
skill $S_k$ is decomposed into reusable subskills
$S_{k,1},S_{k,2},\ldots$, which may be further decomposed into
sub-subskills such as $S_{k,1,1}$ and $S_{k,1,2}$.
The same hierarchy provides construction operators for the forward path
and verification or repair operators for the backward loop.

Given the evaluation intent, \textit{Skill Selection \& Composition}
retrieves compatible skills and binds them to the available resources,
templates, and target artifact types.
The selected skills and artifacts form a typed construction workflow,
where each connection requires matching output types and downstream
preconditions.
Although the instantiated workflow varies across benchmarks, it follows
the stable forward construction path illustrated in
Figure~\ref{fig:overview0727}.

\textit{Intent Blueprinting} converts the user evaluation intent into
capability targets, resource requirements, task specifications, and
output schemas.
\textit{Data Collection} then selects one or more resource operations:
\textit{Acquire Raw Sources}, \textit{Adapt Existing Benchmarks}, or
\textit{Collect in Simulation}.
Heterogeneous resources are accessed through unified interfaces that
describe their available fields, supported operations, and
input--output schemas.
Interactive simulators additionally expose scene initialization, state
restoration, action constraints, and terminal-state inspection.

The collected resources are transformed by
\textit{Evidence \& State Structuring}.
For Offline EQA, this operation produces registered
evidence records with spatial, semantic, temporal, and source
information.
For Interactive Embodied, it additionally organizes
restorable environment states, valid actions, task preconditions, and
goal-relevant variables.
These structured artifacts are consumed by
\textit{Benchmark Synthesis}, which consists of
\textit{Template Design}, \textit{Code Generation}, and
\textit{Metric Production}.
Template Design specifies question or instruction patterns and their
required evidence fields; Code Generation produces questions, answers,
task instructions, action interfaces, goals, and verifiers; and Metric
Production creates automatic scoring protocols and task-level
metrics.
After verification, \textit{Evaluation Reporting} summarizes benchmark
composition, model results, and diagnostic statistics for the final
package.

\begin{figure*}[!ht]
	\centering
	\includegraphics[width=0.7\linewidth]{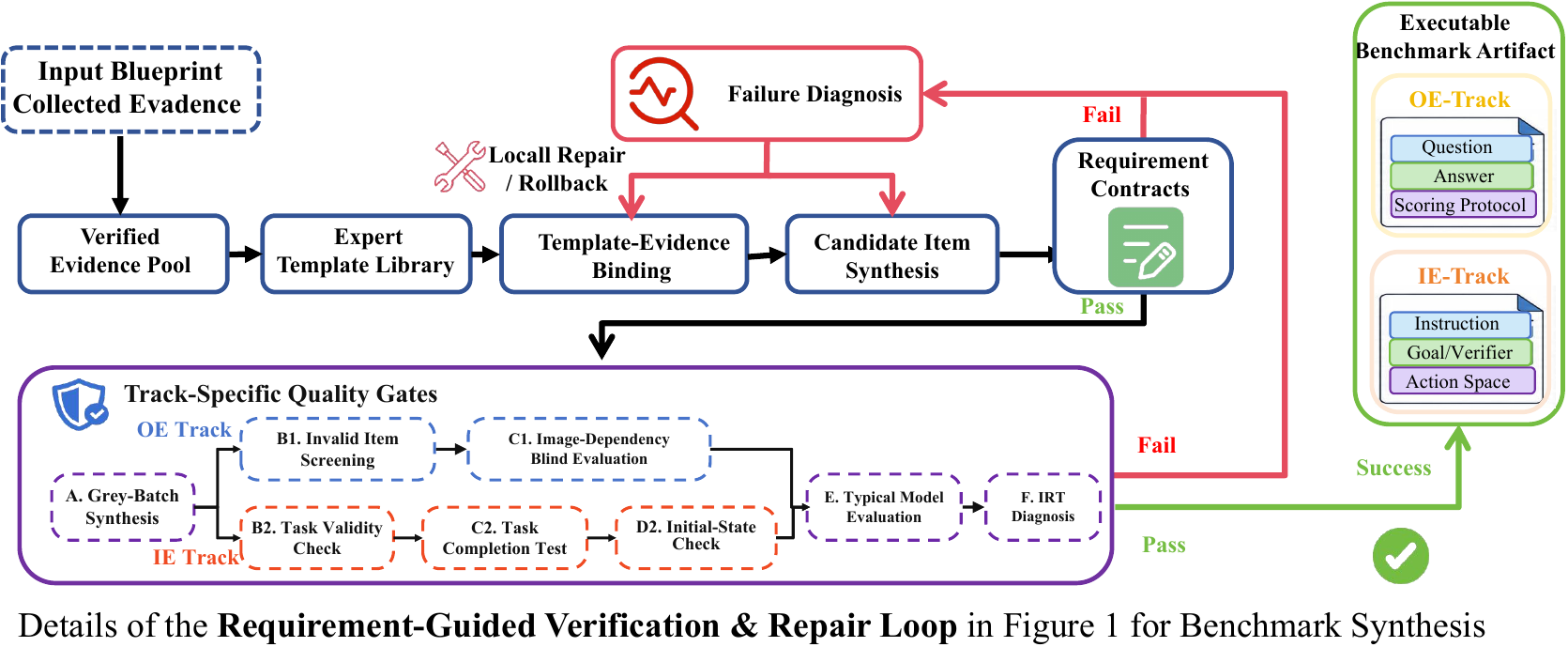}
	\caption{
		Detailed workflow of the requirement-guided verification and repair loop in Embodied-BenchForge. Starting from an input blueprint and collected evidence, the system constructs a verified evidence pool, retrieves expert templates, performs template–evidence binding, and synthesizes candidate items. Candidate items are checked against requirement contracts before entering track-specific quality gates. Following grey-batch synthesis, OE-Track applies invalid-item screening and image-dependency blind evaluation, whereas IE-Track applies task-validity, completion, and initial-state checks. Both tracks subsequently undergo typical-model evaluation and IRT diagnosis. Failures trigger diagnosis and local repair or rollback of the affected templates, bindings, or synthesized items, while successful items are packaged as executable OE-Track or IE-Track benchmark artifacts.
	}
	\label{fig:stage4_quality_control}
\end{figure*}

\paragraph{Requirement-Guided Verification and Repair.}
Each artifact type is associated with a requirement contract
\begin{equation}
\kappa_{\tau}
=
\left\{
\phi_{\tau,1},\ldots,\phi_{\tau,n}
\right\},
\end{equation}
where $\phi_{\tau,j}$ denotes a deterministic, execution-based, or
model-assisted verification rule.
The \textit{Requirement Contracts} component checks whether an artifact
satisfies all mandatory structural, grounding, semantic, and
execution requirements before it is released downstream.
The contracts are specialized to different artifact types.
Intent blueprints are checked for capability alignment and resource
feasibility.
Source and evidence records are checked for completeness, consistency,
and traceability.
Offline EQA items must be grounded in registered evidence, admit
derivable answers, and contain executable scoring protocols.
Interactive Embodied items must provide restorable initial states,
feasible instructions, valid action spaces, consistent goals, and
executable terminal verifiers.
These artifact-specific checks avoid reducing heterogeneous validity
requirements to a single generic quality score.

Artifacts that satisfy their requirement contracts enter the
\textit{Quality Gates}.
Figure~\ref{fig:stage4_quality_control} details this process for
benchmark synthesis.
An \textit{Input Blueprint} and \textit{Collected Evidence} are first
converted into a \textit{Verified Evidence Pool}.
The pool is combined with the \textit{Expert Template Library} through
\textit{Template--Evidence Binding}, after which
\textit{Candidate Item Synthesis} produces candidate benchmark
artifacts.
The candidates are first checked by \textit{Requirement Contracts} and
then routed to the \textit{Track-Specific Quality Gates}.

The quality-gate process begins with
\textit{A. Grey-Batch Synthesis}, which constructs a small candidate
batch for process-level evaluation.
For the OE-Track,
\textit{B1. Invalid Item Screening} removes malformed, inconsistent, or
unsupported items, and
\textit{C1. Image-Dependency Blind Evaluation} examines whether solving
an item actually depends on its intended visual or spatial evidence.
For the IE-Track,
\textit{B2. Task Validity Check} verifies instruction, action-space, and
goal consistency;
\textit{C2. Task Completion Test} executes the task and tests its
terminal verifier; and
\textit{D2. Initial-State Check} confirms that the required environment
state can be restored consistently.

Artifacts passing the track-specific branches are assessed by
\textit{E. Typical Model Evaluation}, which records model--item
responses and coarse difficulty or failure patterns.
\textit{F. IRT Diagnosis} then provides an auxiliary analysis of item
difficulty, discrimination, and abnormal response behavior.
OE-Track artifacts that pass the complete process contain a
\textit{Question}, \textit{Answer}, and
\textit{Scoring Protocol}, whereas IE-Track artifacts contain an
\textit{Instruction}, \textit{Goal/Verifier}, and
\textit{Action Space}.
They are released as \textit{Executable Benchmark Artifacts}.

When a requirement contract or quality gate fails,
\textit{Failure Diagnosis} identifies the violated requirements and
affected construction dependencies.
Each skill execution records its inputs, outputs, resources, templates,
parameters, and verification results, enabling an appropriate
\textit{Local Repair / Rollback} operation.
The framework may repair the current artifact, re-execute
\textit{Template--Evidence Binding} or
\textit{Candidate Item Synthesis}, or roll back to an upstream evidence,
state, or blueprint artifact.
Only affected downstream artifacts are reconstructed, while unrelated
verified artifacts are retained.
Repaired artifacts re-enter the requirement contracts and track-specific
quality gates until accepted into the executable benchmark package or
rejected.

\begin{table}[!ht]
\centering
\small
\setlength{\tabcolsep}{2.0pt}

\begin{tabular}{@{}lllc@{}}
\toprule
\textbf{Benchmark}
& \textbf{Embodiment / Scene}
& \textbf{Source}
& \textbf{Statistics} \\
\midrule

\multicolumn{4}{@{}l}{\textit{Offline EQA Track}} \\
\midrule

SHT
& Household robot
& ALFRED (Sim.)
& 8 / 6 / 22 \\

SHN
& Mobile robot
& Habitat (Sim.)
& 7 / 4 / 13 \\

SD
& Autonomous vehicle
& CARLA (Sim.)
& 8 / 6 / 80 \\

SA
& Robotic arm
& LIBERO (Sim.)
& 9 / 8 / 203 \\

SU
& UAV
& Real aerial images
& 9 / 7 / 260 \\

SQ
& Quadruped robot
& TartanGround
& 8 / 8 / 192 \\

\midrule
\multicolumn{4}{@{}l}{\textit{Interactive Embodied Track}} \\
\midrule

IE-Track
& Four household scenes
& AI2-THOR (Sim.)
& 220 / 28 / 2 \\

\bottomrule
\end{tabular}
\caption{
Data sources and composition of the benchmarks constructed by
Embodied-BenchForge.
For Offline EQA, \textit{Statistics} reports capability dimensions,
question types, and expert templates.
For IE-Track, it reports tasks, scenes, and difficulty levels.
The four interactive scene types are kitchen, living room, bedroom,
and bathroom.
}
\label{tab:constructed_benchmark_summary}

\end{table}

\section{Constructed Benchmark Suite}
\label{sec:constructed_suite}

Embodied-BenchForge instantiates two complementary benchmark tracks,
summarized in Table~\ref{tab:constructed_benchmark_summary}.
The \textit{OE-Track} contains six benchmarks spanning household robots,
mobile robots, autonomous vehicles, robotic arms, UAVs, and quadruped
robots.
They are SpatialHome-Task (SHT), SpatialHome-Nav (SHN), SpatialDrive
(SD), SpatialArm (SA), SpatialUAV (SU), and SpatialQuadruped (SQ).
Their data sources include ALFRED, Habitat, CARLA, and LIBERO simulators,
real unlabeled aerial images, and the existing TartanGround dataset.
Each instance combines embodied observations with grounded questions,
reference answers, and executable scoring protocols.
The \textit{IE-Track} contains 220 instruction-conditioned tasks
constructed from shared AI2-THOR scene pools across kitchens, living
rooms, bedrooms, and bathrooms.
Each task specifies a restorable initial state, an instruction, an action
interface, goal conditions, and an executable terminal verifier, while
retaining the interaction trajectory for subsequent EQA.
Together, the two tracks extend embodied evaluation from observation-based
reasoning to executable interaction and closed-loop task completion,
demonstrating that Embodied-BenchForge can construct heterogeneous yet
standardized benchmark artifacts.

\section{Experiments}
\label{sec:experiments}

\paragraph{Experimental Setup.} We evaluate Embodied-BenchForge from four perspectives:
(1) artifact quality and measurement reliability;
(2) the effectiveness of its construction components and closed-loop
repair;
(3) the discriminative and diagnostic utility of the constructed
benchmarks; and
(4) construction efficiency and scalability.
Unless otherwise specified, Qwen3.6-27B coordinates the
construction workflow, while GPT-5.5-Pro supports artifact synthesis and
model-assisted verification.
We evaluate 11 API and local models on both benchmark tracks.
Offline EQA scores are normalized to a 0--100 scale.
For the IE-Track, task success is determined by executable
terminal-state verifiers, and strict cross-model comparison uses the
220 exact tasks shared by all models.

\paragraph{Quality Assessment Metrics.}
Benchmark quality is assessed by four LLM/VLM judges and ten human
annotators.
The judge-based evaluation covers eight dimensions:
\textit{User-Intent Alignment} (UIA),
\textit{Format and Usability Quality} (FSQ),
\textit{Question--Answer Consistency} (QAC),
\textit{Evidence-Grounding Correctness} (EGC),
\textit{Spatial-Geometric Consistency} (SGC),
\textit{Action/Affordance Consistency} (AFC),
\textit{Target-Signal Dependency} (TSD), which measures reliance on the
intended visual or spatial evidence rather than shortcuts, and
\textit{Skill Challenge} (SSC), which reflects task difficulty and
capability discrimination.
Human annotations on sampled artifacts provide a reference for quality
and Judge--Human agreement.
The Overall score aggregates artifact quality across sampled items and
judge models.

\paragraph{Efficiency Metrics.}
Construction efficiency is measured by token cost, wall-clock time,
tokens per accepted item, and throughput.
Offline EQA costs are normalized to 10,000-item benchmarks and exclude
target-model evaluation.
Additional protocols, costs, and implementation details are provided in
the appendix.

\subsection{Construction Quality and Efficiency}

\begin{table}[t]
\centering
\small
\setlength{\tabcolsep}{1.0pt}

\begin{tabular}{@{}lccccccccc@{}}
\toprule
\textbf{Bench.}
& \textbf{UIA}
& \textbf{FSQ}
& \textbf{QAC}
& \textbf{EGC}
& \textbf{SGC}
& \textbf{AFC}
& \textbf{TSD}
& \textbf{SSC}
& \textbf{Overall} \\
\midrule
SHT & 90.25 & 93.00 & 89.50 & 91.75 & 86.25 & 88.50 & 86.00 & 83.25 & 88.56 \\
SHN & 89.75 & 87.25 & 89.00 & 91.50 & 89.25 & 90.75 & 87.75 & 87.50 & 89.09 \\
SD  & 87.25 & 88.75 & 86.50 & 91.75 & 89.25 & 89.00 & 91.50 & 87.75 & 88.97 \\
SQ  & 88.50 & 94.00 & 85.75 & 91.00 & 86.50 & 88.50 & 87.00 & 85.50 & 88.34 \\
SU  & 85.50 & 92.25 & 86.50 & 90.50 & 86.75 & 89.00 & 88.00 & 91.25 & 88.72 \\
SA  & 92.00 & 90.00 & 91.25 & 89.50 & 94.25 & 88.00 & 91.00 & 90.00 & 90.75 \\
\bottomrule
\end{tabular}
\caption{
LLM-as-a-Judge quality assessment of six
\textit{OE-Track} benchmarks.
Scores are averaged over four judge models.
}
\label{tab:qtrack_llm_as_judge_quality}
\end{table}

\paragraph{Artifact Quality.}
Table~\ref{tab:qtrack_llm_as_judge_quality} shows consistently high
artifact quality across the six Offline EQA benchmarks, with Overall
scores ranging from 88.34 to 90.75.
The small variation across heterogeneous carriers and data sources
indicates that Embodied-BenchForge maintains stable construction quality
rather than being specialized to a single environment.
The relatively lower scores on target-signal dependency and scoring
consistency also identify dimensions that remain more difficult to
control automatically.

\paragraph{Construction Efficiency.}
Embodied-BenchForge constructs a 10,000-item OE-Track benchmark in
38--160 minutes (86.0 on average).
Constructing and validating 20 executable IE-Track tasks takes 1.5 hours.
OE-Track orchestration averages 11.45M tokens, or 1,145 tokens per
accepted item, with a throughput of 116.3 items per minute.
SpatialDrive achieves the highest throughput, whereas SpatialUAV incurs
the largest cost due to extensive evidence organization and validation
for real unlabeled aerial images.
These results demonstrate scalable construction across heterogeneous
resources and tracks.
Per-benchmark costs are reported in the appendix.



\begin{table}[t]
\centering
\small
\setlength{\tabcolsep}{3pt}

\begin{tabular}{@{}lccccccc@{}}
\toprule
\textbf{Model}
& \textbf{SHT}
& \textbf{SHN}
& \textbf{SD}
& \textbf{SQ}
& \textbf{SU}
& \textbf{SA}
& \textbf{Mean} \\
\midrule

\raisebox{-0.2ex}{\includegraphics[height=1em]{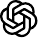}}~
GPT-5.5
& \textbf{53.88}
& \textbf{60.57}
& 44.52
& \underline{51.38}
& \textbf{63.78}
& \textbf{71.89}
& \textbf{57.67} \\

\raisebox{-0.2ex}{\includegraphics[height=1em]{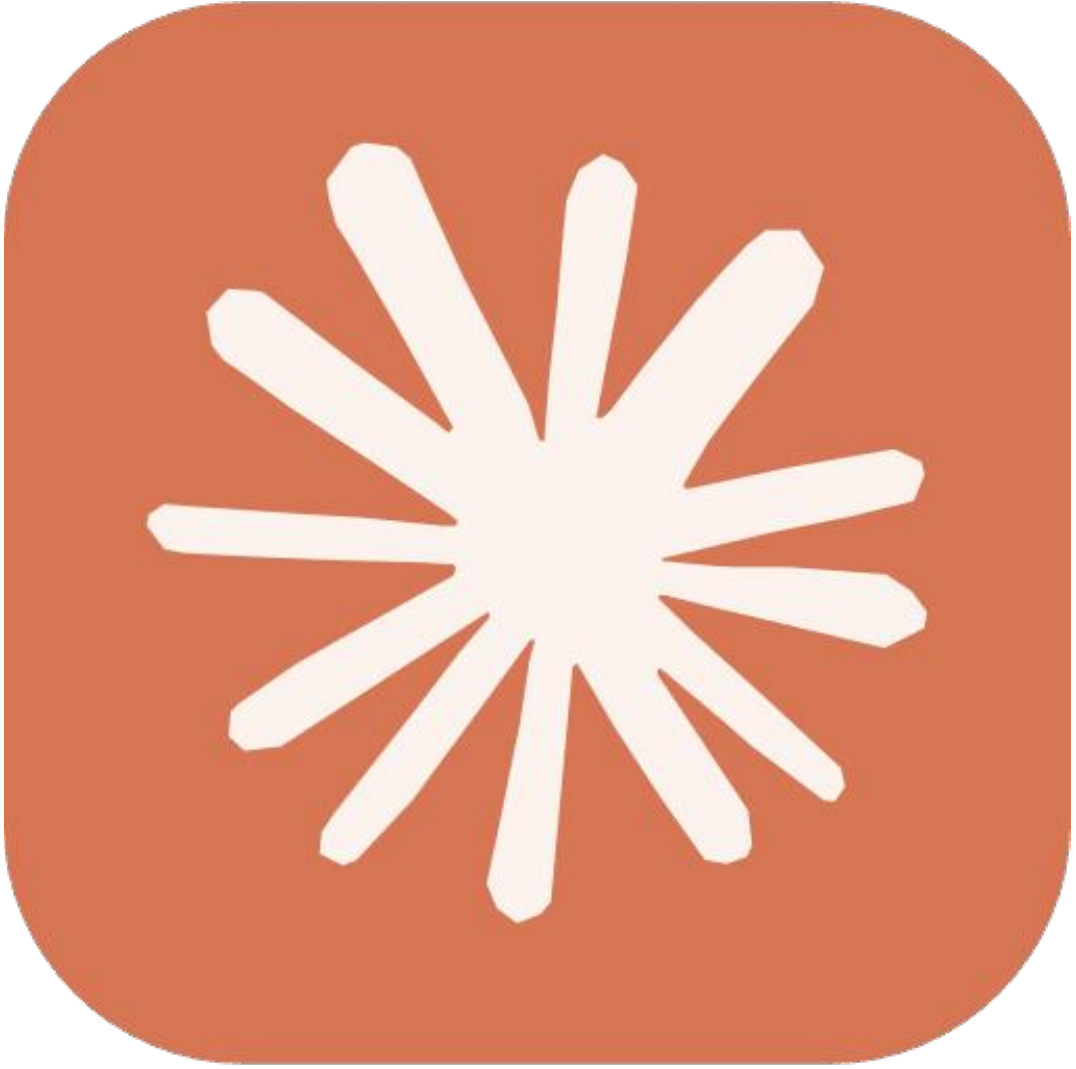}}~
Claude Opus 4.7
& 52.25
& \underline{57.57}
& 44.40
& \textbf{53.00}
& \underline{63.67}
& \underline{68.78}
& \underline{56.61} \\

\raisebox{-0.2ex}{\includegraphics[height=1em]{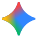}}~
Gemini-3-Flash
& 50.50
& 45.43
& \textbf{46.96}
& \underline{51.38}
& 49.89
& 67.67
& 51.97 \\

\raisebox{-0.2ex}{\includegraphics[height=1em]{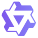}}~
Qwen3.6-35B
& \underline{52.75}
& 46.00
& 42.40
& 50.00
& 54.00
& 54.00
& 49.86 \\

\raisebox{-0.2ex}{\includegraphics[height=1em]{logos/openai.pdf}}~
GPT-5.4-Mini
& 48.25
& 40.00
& \underline{46.85}
& 42.50
& 45.78
& 60.00
& 47.23 \\

\raisebox{-0.2ex}{\includegraphics[height=1em]{logos/openai.pdf}}~
GPT-5.4-Nano
& 41.00
& 39.29
& 43.50
& 35.63
& 45.67
& 52.67
& 42.96 \\

\raisebox{-0.2ex}{\includegraphics[height=1em]{logos/claude-ai-icon.pdf}}~
Claude Sonnet 4.6
& 49.88
& 40.86
& 36.21
& 39.50
& 42.33
& 46.44
& 42.54 \\

\raisebox{-0.2ex}{\includegraphics[height=1em]{logos/qwen-color.pdf}}~
Qwen3.5-4B
& 37.88
& 31.14
& 40.08
& 27.38
& 32.11
& 47.22
& 35.97 \\

\raisebox{-0.2ex}{\includegraphics[height=1em]{logos/claude-ai-icon.pdf}}~
Claude 4.5 Haiku
& 50.25
& 32.57
& 25.83
& 28.75
& 34.67
& 36.33
& 34.73 \\

\raisebox{-0.2ex}{\includegraphics[height=1em]{logos/qwen-color.pdf}}~
Qwen3.5-2B
& 36.00
& 27.86
& 33.96
& 22.38
& 35.33
& 36.89
& 32.07 \\

\raisebox{-0.2ex}{\includegraphics[height=1em]{logos/qwen-color.pdf}}~
Qwen3.5-0.8B
& 39.88
& 29.57
& 32.79
& 23.13
& 29.22
& 27.44
& 30.34 \\

\midrule
Human
& 84.98
& 87.80
& 82.26
& 83.77
& 88.91
& 87.63
& 85.89 \\

\bottomrule
\end{tabular}

\caption{
Average performance of 11 models and humans across the six benchmarks
in the \textit{OE-Track}.
Scores are averaged over capability dimensions.
}
\label{tab:qtrack_model_avg}
\end{table}

\begin{table}[!ht]
\centering
\small
\setlength{\tabcolsep}{3pt}

\begin{tabular}{@{}lcccc@{}}
\toprule
\textbf{Model}
& \textbf{SR}
& \textbf{EQA Acc.}
& \textbf{SRL}
& \textbf{\makecell{Avg. Steps}} \\
\midrule

\raisebox{-0.2ex}{\includegraphics[height=1em]{logos/openai.pdf}}~
GPT-5.5
& \textbf{83.18}
& \textbf{73.64}
& \textbf{61.36}
& 40.13 \\

\raisebox{-0.2ex}{\includegraphics[height=1em]{logos/claude-ai-icon.pdf}}~
Claude Opus 4.7
& \underline{77.73}
& \underline{70.00}
& \underline{55.00}
& 39.65 \\

\raisebox{-0.2ex}{\includegraphics[height=1em]{logos/gemini-color.pdf}}~
Gemini-3-Flash
& 72.27
& 59.55
& 42.73
& 42.74 \\

\raisebox{-0.2ex}{\includegraphics[height=1em]{logos/qwen-color.pdf}}~
Qwen3.6-35B
& 46.36
& 50.91
& 25.45
& 30.69 \\

\raisebox{-0.2ex}{\includegraphics[height=1em]{logos/openai.pdf}}~
GPT-5.4-Nano
& 20.45
& 53.64
& 12.73
& 77.22 \\

\raisebox{-0.2ex}{\includegraphics[height=1em]{logos/openai.pdf}}~
GPT-5.4-Mini
& 16.82
& 46.82
& 8.18
& 85.30 \\

\raisebox{-0.2ex}{\includegraphics[height=1em]{logos/claude-ai-icon.pdf}}~
Claude Sonnet 4.6
& 16.82
& 42.27
& 7.73
& 81.46 \\

\raisebox{-0.2ex}{\includegraphics[height=1em]{logos/claude-ai-icon.pdf}}~
Claude 4.5 Haiku
& 5.00
& 55.91
& 1.36
& 99.76 \\

\raisebox{-0.2ex}{\includegraphics[height=1em]{logos/qwen-color.pdf}}~
Qwen3.5-4B
& 0.00
& 55.91
& 0.00
& 105.54 \\

\raisebox{-0.2ex}{\includegraphics[height=1em]{logos/qwen-color.pdf}}~
Qwen3.5-2B
& 0.00
& 46.82
& 0.00
& 35.76 \\

\raisebox{-0.2ex}{\includegraphics[height=1em]{logos/qwen-color.pdf}}~
Qwen3.5-0.8B
& 0.00
& 22.73
& 0.00
& 132.59 \\

\midrule
Human
& 97.73
& 74.09
& 71.82
& 44.00 \\

\bottomrule
\end{tabular}

\caption{
Performance of 11 models and humans on all 220 tasks in the
\textit{IE-Track}.
SR, EQA Accuracy, and SRL are reported as percentages, while Avg.\ Steps
denotes the mean number of executed steps.
SRL measures the proportion of tasks that are both successfully completed
and correctly answered in trajectory-grounded EQA.
}
\label{tab:interactive_main_results}
\end{table}

\subsection{Benchmark Evaluation Utility}

\paragraph{Offline EQA Performance.}
Table~\ref{tab:qtrack_model_avg} shows a clear capability hierarchy.
Humans achieve a mean score of 85.89, outperforming the strongest model
by 28.22 points.
Among the evaluated models, GPT-5.5 ranks first with 57.67, followed by
Claude Opus 4.7 with 56.61, while Qwen3.6-35B is the strongest local
model with 49.86.
These results demonstrate that the constructed OE-Track provides effective and discriminative evaluation of offline embodied capabilities.

\paragraph{Interactive Embodied Performance.}
Table~\ref{tab:interactive_main_results} shows that humans achieve the
highest SR and SRL, while GPT-5.5 performs best among the evaluated
models with 83.18\% SR, 73.64\% EQA accuracy, and 61.36\% SRL.
Claude Opus 4.7 and Gemini-3-Flash rank second and third, respectively.
Several smaller models retain moderate EQA accuracy despite near-zero task success, revealing a clear gap between trajectory understanding and executable embodied competence.
These results demonstrate that the constructed IE-Track effectively distinguishes trajectory understanding from executable embodied competence.



\begin{figure*}[t]
    \centering
    \includegraphics[width=0.8\textwidth]
    {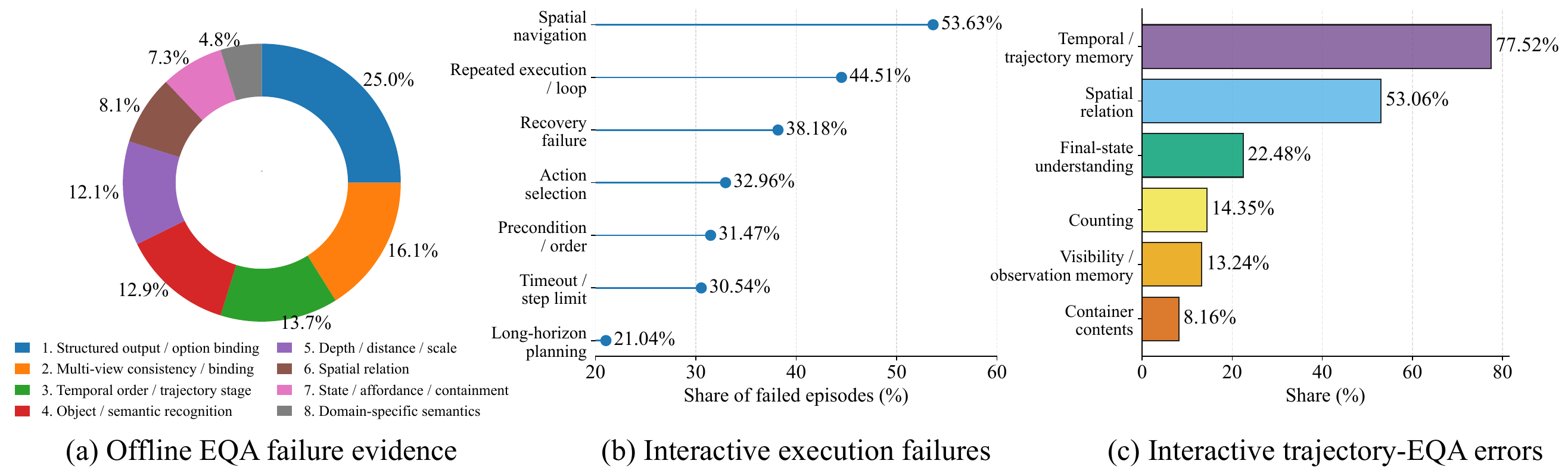}
    \caption{
    Cross-track capability diagnosis.
    (a) Diagnostic risk evidence from Offline EQA model reports.
    (b) Major execution errors measured over failed Interactive episodes.
    (c) Major trajectory-EQA errors measured over incorrect-answer
    evidence.
    Labels in (b) and (c) are non-exclusive and therefore do not sum to
    100\%.
    }
    \label{fig:cross_track_diagnosis}
\end{figure*}

\subsection{Construction Mechanisms and Ablation}

\begin{table}[t]
\centering
\small
\setlength{\tabcolsep}{3.0pt}

\begin{tabular}{@{}lcccc@{}}
\toprule
\textbf{Setting}
& \textbf{\makecell{Skills\\Used}}
& \textbf{\makecell{Reused\\Skills}}
& \textbf{Specific}
& \textbf{\makecell{Reuse\\Rate}} \\
\midrule
Simulator-based OE
& 40.3
& 39.3
& 1.0
& 97.5\% \\

SpatialUAV
& 41
& 34
& 7
& 82.9\% \\

SpatialQuadruped
& 41
& 36
& 5
& 87.8\% \\

IE-Track
& 44
& 36
& 8
& 81.8\% \\
\midrule

\textbf{Overall}
& \textbf{41.0}
& \textbf{37.6}
& \textbf{3.4}
& \textbf{91.6\%} \\
\bottomrule
\end{tabular}

\caption{
Skill reuse across representative benchmark-construction settings.
The implementation snapshot contains 64 distinct skill/card definitions.
\textit{Specific} denotes source- or track-specific skills and adapters.
The simulator-based OE row averages SHT, SHN, SD, and SA.
Reuse rates are computed from pooled module counts.
}
\label{tab:skill_reuse}
\end{table}

\paragraph{Skill Reuse.}
Table~\ref{tab:skill_reuse} shows that the four simulator-based OE
benchmarks reuse 97.5\% of their invoked modules and require only one
source-specific adapter on average.
SpatialUAV and SpatialQuadruped introduce additional modules for
real-image and dataset-specific processing, while the IE-Track requires
interaction-specific skills for state restoration, action execution, and
terminal verification.
Nevertheless, their reuse rates remain above 81\%.
Overall, 91.6\% of the invoked modules are reused, demonstrating that
Embodied-BenchForge supports heterogeneous benchmark construction through
a shared skill backbone with limited source- or track-specific extensions.

\begin{table}[t]
\centering
\small
\setlength{\tabcolsep}{3pt}
\begin{tabular}{@{}lccccc@{}}
\toprule
\textbf{Setting}
& \textbf{\makecell{LLM\\Judge}}
& \textbf{Human}
& \textbf{\makecell{Agree-\\ment ($\rho$)}}
& \textbf{\makecell{Valid\\Rate}}
& $\boldsymbol{\Delta_{\mathrm H}}$ \\
\midrule
Full
& \textbf{89.07}
& \textbf{91.28}
& \textbf{0.87}
& \textbf{93.7\%}
& -- \\

w/o Design Skill
& 82.10
& 83.76
& 0.84
& 86.2\%
& -7.52 \\

w/o Source Grounding
& 74.48
& 72.12
& 0.81
& 73.6\%
& -19.16 \\

w/o Evidence Skill
& 76.76
& 74.08
& 0.82
& 80.8\%
& -17.20 \\

w/o Verification and Repair
& 70.59
& 68.03
& 0.78
& 62.4\%
& -23.25 \\
\bottomrule
\end{tabular}
\caption{
System-level ablation averaged across the six benchmarks in the
\textit{OE-Track}.
\textit{LLM Judge} and \textit{Human} report overall artifact-quality
scores.
\textit{Agreement} is the sample-level Judge--Human Spearman correlation,
and \textit{Valid Rate} is the percentage of generated candidates
satisfying all mandatory requirements before release.
$\Delta_{\mathrm H}$ denotes the human-score change relative to the full
system.
}
\label{tab:qtrack_ablation_avg}
\end{table}

\begin{table}[t]
\centering
\small
\setlength{\tabcolsep}{3pt}
\begin{tabular}{@{}lccccc@{}}
\toprule
\textbf{Variant}
& \textbf{\makecell{Qual-\\ity $\uparrow$}}
&\textbf{\makecell{Valid\\Rate $\uparrow$}}
& \textbf{\makecell{TSD\\$\uparrow$}}
& \textbf{\makecell{High-\\Disc. $\uparrow$}}
& \textbf{\makecell{Tokens}} \\
\midrule

Final-stage inspection only
& 68.32
& 10.4\%
& 76.8
& 14.6\%
& 1.53M \\

w/o Requirement Contracts
& 78.74
& 62.2\%
& 82.5
& 40.8\%
& 2.19M \\

w/o Blind Evaluation
& 80.89
& 73.5\%
& 81.6
& 37.2\%
& 1.65M \\

w/o Gray-Box Evaluation
& 82.21
& 75.7\%
& 88.4
& 32.6\%
& 1.57M \\

w/o IRT Diagnosis
& 88.47
& 81.1\%
& 89.0
& 36.7\%
& 2.58M \\

w/o Provenance-Guided Repair
& 88.74
& 91.8\%
& 89.3
& 48.1\%
& 3.93M \\

Full Process Quality Control
& \textbf{89.21}
& \textbf{93.7\%}
& \textbf{91.2}
& \textbf{49.4\%}
& 2.62M \\

\bottomrule
\end{tabular}

\caption{
Fine-grained ablation of process quality control on SpatialUAV.
All variants receive the same request to produce 1,000 benchmark items.
\textit{Quality} and TSD are averaged over all items, while
\textit{Valid Rate} reports the proportion satisfying mandatory
requirements.
\textit{High-Disc.} is the proportion of items with \(a_i \geq 0.7\),
estimated using a common regularized 2PL fit.
Tokens include synthesis, verification, diagnosis, and repair costs
for the 1,000-item output.
}
\label{tab:process_quality_control_ablation}
\end{table}









\paragraph{System-Level Ablation.}
Table~\ref{tab:qtrack_ablation_avg} evaluates the system components
across the six OE-Track benchmarks.
Removing verification and repair causes the largest degradation, reducing
the human quality score by 23.25 points and the valid rate from 93.7\% to
62.4\%.
Source grounding and evidence construction are critical, causing
human-score drops of 19.16 and 17.20 points, respectively.
Removing the design skill produces a smaller but consistent decline.
The reduction in Judge--Human agreement from 0.87 to as low as 0.78
indicates that the complete synthesis and verification pipeline
produces more reliable and less ambiguous artifacts.

\paragraph{Process-Quality-Control Ablation.}
Table~\ref{tab:process_quality_control_ablation} evaluates the
verification-and-repair components on SpatialUAV.
Final-stage inspection yields only 10.4\% valid items and 14.6\%
high-discrimination items.
Removing requirement contracts decreases quality by 10.47 points and
valid rate by 31.5 percentage points, while removing blind evaluation,
gray-box evaluation, or IRT diagnosis mainly weakens target dependency
or item discrimination.
Without provenance-guided repair, quality remains similar but token cost
rises from 2.62M to 3.93M.
Overall, the full process achieves the best balance of quality, validity,
diagnostic value, and construction cost.


\paragraph{Cross-Track Capability Diagnosis.}
Figure~\ref{fig:cross_track_diagnosis} reveals complementary failure
patterns across the two tracks.
Offline failures are dominated by structured answer binding, multi-view
consistency, and temporal reasoning.
Interactive failures concentrate on spatial navigation, repeated
execution, recovery, and action ordering, while trajectory-EQA errors
primarily involve temporal memory and spatial relations.
The OE-Track therefore diagnoses whether models can construct a
coherent embodied representation, whereas the IE-Track
determines whether that representation supports sustained, executable,
and recoverable behavior.

\section{Conclusion}
\label{sec:conclusion}

We introduced \textbf{Closed-Loop Benchmark Synthesis}, which unifies
forward artifact synthesis with backward verification and repair, and
instantiated it in \textbf{Embodied-BenchForge}.
The framework composes typed, reusable skills into executable workflows
and uses requirement contracts, provenance tracing, and targeted repair
to construct reliable embodied benchmarks.
It produces an \textit{OE-Track} covering six embodied settings
and an \textit{IE-Track} containing 220 executable
tasks.
Experiments demonstrate strong artifact quality, model discrimination,
measurement reliability, and construction efficiency, while cross-track
analysis reveals a clear gap between embodied understanding and
closed-loop interaction.
Overall, Embodied-BenchForge provides a scalable and continually
updatable framework for grounded, executable, and traceable embodied
benchmark construction.

\bibliography{benchforge}
\bibliographystyle{iclr2027_conference}

\clearpage
\appendix

\paragraph{Overview.}
This supplementary material is organized into five appendices.
\textbf{Appendix~\ref{app:closed_loop_quality_control}} presents the
shared closed-loop quality-control framework and its realization in the
Offline EQA and Interactive Embodied Tracks, including Requirement
Contracts, quality gates, provenance-guided repair, representative repair
cases, and executable interactive-task evaluation.
\textbf{Appendix~\ref{app:resource_coverage}} examines resource coverage
and cross-resource adaptation across heterogeneous simulators, real-world
imagery, robot trajectories, and interactive environments.
\textbf{Appendix~\ref{app:skill_reuse}} summarizes the representative
Skill Library taxonomy and defines the skill-reuse accounting protocol
used across the evaluated benchmark-construction runs.
\textbf{Appendix~\ref{app:construction_cost_analysis}} reports the token,
wall-clock, throughput, and stage-wise construction-cost analysis.
Finally, \textbf{Appendix~\ref{app:experimental_details}} provides the
implementation settings and extended evaluation results, including human
and model-judge protocols, benchmark composition, LLM-as-Judge assessment,
per-capability model performance, and IRT-based diagnostic analysis.

\section{Closed-Loop Quality Control Across the OE and IE Tracks}
\label{app:closed_loop_quality_control}

This appendix first defines the shared Requirement Contracts, quality
gates, and provenance-guided repair protocol.
It then presents their track-specific realization in the Offline EQA
Track through representative before-and-after repair cases, followed by
the Interactive Embodied Track through its execution protocol, task
packages, model trajectories, and verification-and-repair cases.

\subsection{Definitions and Shared Closed-Loop Protocol}
\label{app:shared_definitions}

Embodied-BenchForge applies a shared closed-loop protocol across both
benchmark tracks rather than assigning a single overall quality score
to the final output. The protocol combines artifact-specific Requirement
Contracts, track-specific quality gates, and provenance-guided repair.
For an artifact type \(\tau\), its requirement contract is defined as
\[
\kappa_{\tau}
=
\{\phi_{\tau,1},\ldots,\phi_{\tau,n}\},
\]
where each predicate checks a structural, grounding, semantic, or
execution requirement.
An artifact proceeds downstream only when all applicable mandatory
predicates are satisfied.

The framework uses three verification types.
\textit{Deterministic} checks operate on schemas, fields, references,
option sets, provenance records, and reproducible derivations.
\textit{Execution-based} checks run scorers, state-restoration routines,
simulator actions, reference executions, and terminal verifiers.
\textit{Model-assisted} checks are used only for semantic relations that
cannot be established reliably through structure or execution alone,
such as capability alignment and instruction--goal consistency.
A model-assisted result cannot override a failed deterministic or
execution-based requirement.

\subsubsection{Artifact-Specific Requirement Contracts}
\label{app:representative_contracts}

Table~\ref{tab:requirement_contracts} summarizes the principal mandatory
contracts associated with the artifact types in the forward construction
path.
The table groups closely related artifact types for compactness.
When a contract fails, recorded provenance is used to return the artifact
to the affected construction operation.
Only its dependent artifacts are reconstructed, while unrelated verified
artifacts are retained.

\begin{table*}[!ht]
\centering
\small
\setlength{\tabcolsep}{3.0pt}
\renewcommand{\arraystretch}{1.10}
\caption{
Representative mandatory requirement contracts used by
Embodied-BenchForge.
Failure routes identify the construction operation from which the
affected artifact is repaired or reconstructed.
}
\label{tab:requirement_contracts}
\begin{tabularx}{\textwidth}{
@{}p{2.65cm}
>{\raggedright\arraybackslash}X
p{2.65cm}
p{4.05cm}@{}}
\toprule
\textbf{Artifact Group}
& \textbf{Mandatory Requirements}
& \textbf{Verification Type}
& \textbf{Failure Route} \\
\midrule

Intent Blueprint
&
Capability targets, task scope, required resources, and output artifact
types are complete and mutually compatible.
&
Deterministic and model-assisted
&
Revise the affected blueprint field or return to resource selection. \\

Source and Evidence Records
&
Required source fields are present and decodable, and evidence values are
internally consistent and traceable to observations, annotations, or
simulator states.
&
Deterministic and execution-based
&
Re-execute Data Collection or Evidence \& State Structuring. \\

Expert Template and Binding
&
Required evidence fields, task schema, answer or goal rule, and scoring
interface are declared and type-compatible with the selected evidence.
&
Deterministic and model-assisted
&
Repair the template or re-execute Template--Evidence Binding. \\

OE-Track Item and Scorer
&
The question, reference answer, visible evidence, response format, and
scoring protocol are complete.
The answer must be uniquely and reproducibly derivable from registered
evidence.
&
Deterministic, execution-based, and model-assisted
&
Repair or resynthesize the item; re-bind or reconstruct evidence when
required. \\

IE-Track Item, State, and Verifier
&
The instruction, initial state, action interface, goal conditions, and
terminal verifier are mutually consistent, restorable, and executable.
&
Deterministic and execution-based
&
Repair the task or verifier, or roll back to state construction. \\

Model Responses and Diagnostic Records
&
Each response is attributable to a model--item pair, parseable by the
declared scorer, and included in a complete response matrix.
&
Deterministic and execution-based
&
Re-run the affected model--item evaluation or enlarge the diagnostic
batch. \\

Evaluation Report and Benchmark Package
&
Reported metrics are reproducible from accepted artifacts and evaluation
records, and all required OE- or IE-Track components are present and
resolvable.
&
Deterministic and execution-based
&
Regenerate the affected report or package component. \\

\bottomrule
\end{tabularx}
\end{table*}

\subsubsection{Track-Specific Quality Gates}
\label{app:quality_gate_protocol}

Table~\ref{tab:quality_gate_protocol} defines the quality gates shown in
Figure~2 in the main paper.
Gate A evaluates a small candidate batch before full synthesis.
Gates B1--C1 apply to the OE-Track, whereas Gates B2--D2 apply to the
IE-Track.
Gate E produces the model--item response matrix, and Gate F performs
auxiliary item diagnosis.
The same gate definitions and decision criteria are used across all
process-quality-control variants.

\begin{table*}[!ht]
\centering
\small
\setlength{\tabcolsep}{2.8pt}
\renewcommand{\arraystretch}{1.10}
\caption{
Operational definitions of the quality gates in the
requirement-guided verification and repair loop.
IRT-derived discrimination is used for diagnosis rather than as the sole
criterion for artifact release.
}
\label{tab:quality_gate_protocol}
\begin{tabularx}{\textwidth}{
@{}p{2.25cm}
>{\raggedright\arraybackslash}X
p{4.15cm} 
p{3.65cm}@{}}
\toprule
\textbf{Gate}
& \textbf{Main Check}
& \textbf{Operational Criterion}
& \textbf{Failure Route} \\
\midrule

A. Grey-Batch Synthesis
&
Execute enabled templates on a small candidate batch before full-scale
construction.
&
Each retained template must produce valid candidate items, and its
generation, filtering, and scoring records must be available.
&
Repair the template or binding, or re-execute Candidate Item Synthesis. \\

B1. Invalid Item Screening
&
Check OE-Track schemas, media, questions, answers, options, scorers, and
evidence references.
&
Released items must contain no blocking format, grounding, answer, option,
or scoring error.
&
Apply local item repair, rebuild the option set, re-bind evidence, or
resynthesize the item. \\

C1. Image-Dependency Blind Evaluation
&
Compare matched full-evidence and evidence-blind evaluations.
&
Blind performance must remain near its scoring reference, while the full
condition must retain a positive dependency gap under the fixed criterion
used for all variants.
&
Rewrite shortcut-prone questions or options, strengthen visible evidence,
or regenerate the affected template family. \\

B2. Task Validity Check
&
Check consistency among the instructions, scene entities, action interfaces,
preconditions, goal conditions, and verifier.
&
All referenced objects, states, actions, and goal variables must resolve
to the declared environment interface without contradiction.
&
Repair the task fields or roll back to template binding or state
construction. \\

C2. Task Completion Test
&
Execute a verified completion sequence and evaluate the terminal
condition.
&
Repeated valid executions must complete the task, while incomplete or
incorrect controls must not satisfy the terminal verifier.
&
Repair action preconditions, resynthesize the task, or recompile the
verifier. \\

D2. Initial-State Check
&
Restore the serialized initial state and compare its task-relevant
variables with the reference state.
&
Task objects, relations, discrete states, preconditions, and goal
variables must be reproduced within the configured restoration tolerance.
&
Regenerate the state specification or return to simulator collection. \\

E. Typical Model Evaluation
&
Evaluate accepted candidates using the fixed model panel and executable
scoring protocol.
&
Every required model--item pair must produce a valid scored record after
infrastructure retries.
Correctness itself is not a hard release threshold.
&
Re-run missing pairs or repair parsing and scoring failures. \\

F. IRT Diagnosis
&
Estimate item difficulty and discrimination from the common response
matrix.
&
The regularized fit must converge on a non-degenerate response matrix.
Items with \(a_i\geq0.7\) are reported as High-Disc.
&
Audit low- or reverse-discrimination items and, where needed, strengthen
evidence, options, or task templates. \\

\bottomrule
\end{tabularx}
\end{table*}

\paragraph{IRT Diagnosis.}
IRT is applied after Typical Model Evaluation as an auxiliary
item-analysis tool rather than a stand-alone benchmark-validity criterion.
Given the common model--item response matrix, we fit a regularized
two-parameter logistic model,
\[
P(Y_{mi}=1\mid\theta_m)
=
\frac{1}
{1+\exp[-a_i(\theta_m-b_i)]},
\]
where \(\theta_m\) is model ability, \(b_i\) is item difficulty, and
\(a_i\) is item discrimination.
An item is reported as High-Disc.\ when \(a_i\geq0.7\), using the same
response construction, fitting procedure, and threshold for all
ablation variants.
Because the response panel contains 11 models, the estimated parameters
are treated as auxiliary diagnostic indicators rather than precise
psychometric estimates.
Deterministic, execution-based, and semantic validity checks remain the
primary artifact-release criteria.

\subsubsection{Provenance-Guided Repair Protocol}
\label{app:provenance-guided-repair}

This protocol specifies how Embodied-BenchForge converts a failed
requirement or quality-gate result into a targeted repair action. Repair
never relaxes a failed requirement. Instead, the affected artifact is
modified or reconstructed and then evaluated by the same Requirement
Contracts and Track-Specific Quality Gates used for its initial
construction. Provenance-guided repair therefore changes the scope of
reconstruction, not the acceptance criterion.

Each artifact records its type, parent artifact identifiers, evidence
references, template and binding identifiers, producing skill, relevant
parameters, and verification outcomes. Let \(G=(V,E)\) be the artifact
dependency graph, and let \(v_f\) denote the earliest artifact identified by
Failure Diagnosis as causal for a failed check. The invalidated region is
the downstream closure
\[
  D(v_f)=\{u\in V\mid v_f\rightarrow^{*}u\}.
\]
Only artifacts in \(D(v_f)\) are reconstructed, while verified artifacts
outside this set are retained. The repair operator is selected according to
the cause recorded in provenance rather than the final failure symptom.
This mechanism operationalizes the selective re-execution shown in
Figures~1 and~2 and explains why the full process in Table~8 retains quality
while using 2.62M tokens, compared with 3.93M tokens when
provenance-guided repair is removed.

\paragraph{Repair operator definitions.}
\label{app:repair-operators}

\paragraph{Selection principle.}
Failure Diagnosis identifies the violated requirement, the earliest
artifact that made the requirement unsatisfiable, and the smallest
reconstruction scope that can restore it. The framework then selects the
lowest-scope sufficient operator from the provenance trace. Broader
reconstruction is used only when the causal artifact lies further upstream
or when the selected scope cannot satisfy the post-repair contract.

\paragraph{Local field repair.}
This operator is selected when the evidence and template binding are
correct but a bounded Question, Answer, option, scoring, instruction, goal,
or verifier field is inconsistent. Only that field is updated. The
artifact's Requirement Contract and every downstream gate that consumes the
modified field are then re-run.

\paragraph{Template re-binding.}
This operator is selected when the template and evidence are both valid but
the evidence has been assigned to an incorrect or shortcut-prone slot. The
Template-Evidence Binding is replaced while the verified template and
evidence pool are retained. Candidate Item Synthesis, Requirement
Contracts, and the applicable OE-Track or IE-Track quality-gate branch are
then re-run.

\paragraph{Candidate re-synthesis.}
This operator is selected when the binding is valid but the realized
question, instruction, options, or visible presentation is ambiguous,
unnatural, or insufficiently image-dependent. A new candidate is generated
from the same verified template-evidence binding and assigned a new
candidate version. All item-level contracts and track-specific gates are
then re-run.

\paragraph{Skill re-execution.}
This operator is selected when a producing operation is incomplete, stale,
or repeatedly causes the same contract violation. The responsible typed
skill is re-run using the verified inputs and recorded parameters, or a
corrected implementation of that skill. Its output is revalidated, and only
its downstream closure is reconstructed.

\paragraph{Upstream rollback.}
This operator is selected when the causal defect lies in an evidence record,
environment state, source mapping, or input blueprint, so downstream edits
cannot restore traceability or correctness. The process rolls back to the
nearest invalid upstream artifact, reconstructs it from its verified parent,
and invalidates only its descendants. The affected upstream contract,
Template-Evidence Binding, Candidate Item Synthesis, and downstream gates
are then re-run.

\noindent\textbf{Final rejection.}
This operator is selected when no admissible repair can satisfy all
mandatory requirements within the repair budget, or when the necessary
evidence or state cannot be obtained. The candidate lineage is terminated
and excluded from the released benchmark. Its provenance and failure record
are retained for audit and diagnosis, but no downstream item is emitted and
no release check is waived.

\paragraph{Failure-to-operator mapping.}
Failure Diagnosis distinguishes a local symptom from a shared upstream
cause. A single incorrect answer key is locally repairable when the evidence
is correct, whereas the same symptom across many items produced by one
answer program triggers Skill Re-execution. The default routing is as
follows.

\noindent\textbf{Answer or scoring mismatch.} If the evidence record and Template-Evidence Binding are correct, use Local
Field Repair. Escalate to Skill Re-execution when the same producer causes
repeated mismatches.

\paragraph{Incorrect slot assignment.}
Use Template Re-binding when compatible evidence is bound to the wrong
template slot. Escalate to Candidate Re-synthesis, and reject the lineage if
no unique item can be constructed.

\paragraph{Ambiguous or unnatural realization.}
Use Candidate Re-synthesis when the current binding still determines a
unique intended answer or goal. Use Template Re-binding instead when the
defect originates from slot assignment.

\paragraph{Non-target shortcut.}
Use Template Re-binding when Blind Evaluation localizes the shortcut to
visible text or binding structure. Escalate to Candidate Re-synthesis, or to
Skill Re-execution if the shortcut is systematic.

\paragraph{Invalid model-visible media.}
Use Skill Re-execution when valid hidden evidence can be rendered as a
neutral processed RGB anchor. Reject the lineage if no faithful visible
anchor can be constructed.

\paragraph{Invalid evidence or state record.}
Use Upstream Rollback when provenance identifies the earliest stale,
non-traceable, inconsistent, or incorrectly joined artifact. Reject the
lineage if that artifact cannot be reconstructed.

\noindent\textbf{IE-Track field inconsistency.} Use Local Field Repair or Template Re-binding when the Action Space,
Goal/Verifier, instruction, or task preconditions conflict but the
executable state remains valid. Use Skill Re-execution for a shared compiler
or verifier defect; otherwise use Upstream Rollback.

\paragraph{Non-executable terminal verifier.}
Use Skill Re-execution when Task Completion Test fails because the terminal
verifier does not execute or encode the stated goal. Escalate to Candidate
Re-synthesis or Upstream Rollback if feasibility is no longer retained.

\paragraph{Initial-state restoration failure.}
Use Upstream Rollback when Initial-State Check traces the failure to the
state artifact rather than the policy rollout. Reject the lineage if no
restorable state can be registered.

\paragraph{Abnormal discrimination.}
Use Candidate Re-synthesis or Template Re-binding when Typical Model
Evaluation or IRT Diagnosis is corroborated by item-level response or
classical-discrimination evidence. IRT alone does not override passed
mandatory contracts.

\paragraph{Operator invariants.}
All operators satisfy four invariants. First, accepted evidence is never
silently replaced by an unregistered model guess. Second, model-visible
artifacts contain no Answer, hidden provenance, object identifiers, raw
depth, or raw segmentation state. Third, repaired items retain an audit
record linking the new version to the violated requirement and prior
version. Fourth, a repaired item is accepted only after passing the same
deterministic, execution-based, and model-assisted checks applicable to a
newly synthesized item.

\subsection{Offline EQA Track: Verification and Repair Cases}
\label{app:oe_track_quality_control}

The Offline EQA Track applies the shared contracts and the B1--C1 gate
branch to benchmark items constructed from visual observations,
trajectories, annotations, and registered spatial evidence. The cases
below illustrate the principal repair operators across the six Offline
EQA benchmarks and show which upstream artifacts are retained or
re-executed after a failed requirement.

\subsubsection{Representative Before-and-After Repairs}
\label{app:repair-examples}

\newtcolorbox{repairbeforebox}{
  colback=red!2,
  colframe=red!55!black,
  title={Before Repair: Failed Item},
  fonttitle=\bfseries,
  boxrule=0.55pt,
  arc=1pt,
  left=5pt,right=5pt,top=4pt,bottom=4pt,
  before skip=6pt,after skip=6pt
}
\newtcolorbox{repairafterbox}[1][Released Item]{
  colback=green!2,
  colframe=green!45!black,
  title={After Repair: #1},
  fonttitle=\bfseries,
  boxrule=0.55pt,
  arc=1pt,
  left=5pt,right=5pt,top=4pt,bottom=4pt,
  before skip=6pt,after skip=6pt
}
\newtcolorbox{repairrejectbox}{
  colback=orange!3,
  colframe=orange!70!black,
  title={After Repair: Rejected Item},
  fonttitle=\bfseries,
  boxrule=0.55pt,
  arc=1pt,
  left=5pt,right=5pt,top=4pt,bottom=4pt,
  before skip=6pt,after skip=6pt
}
\newtcolorbox{repairsummarybox}[1][Repair Summary]{
  colback=black!2,
  colframe=black!45,
  title={#1},
  fonttitle=\bfseries,
  boxrule=0.5pt,
  arc=1pt,
  left=5pt,right=5pt,top=4pt,bottom=4pt,
  before skip=6pt,after skip=10pt
}

The following cases pair each failed artifact with the selected repair and
its outcome. The first five cases produce released items after
re-verification. The SU candidate is rejected because its visual evidence
remains insufficient.

\newpage
\subsubsection{SHT (ALFRED): Local Field Repair}

\begin{center}
\includegraphics[width=0.96\linewidth,height=0.40\textheight,keepaspectratio]{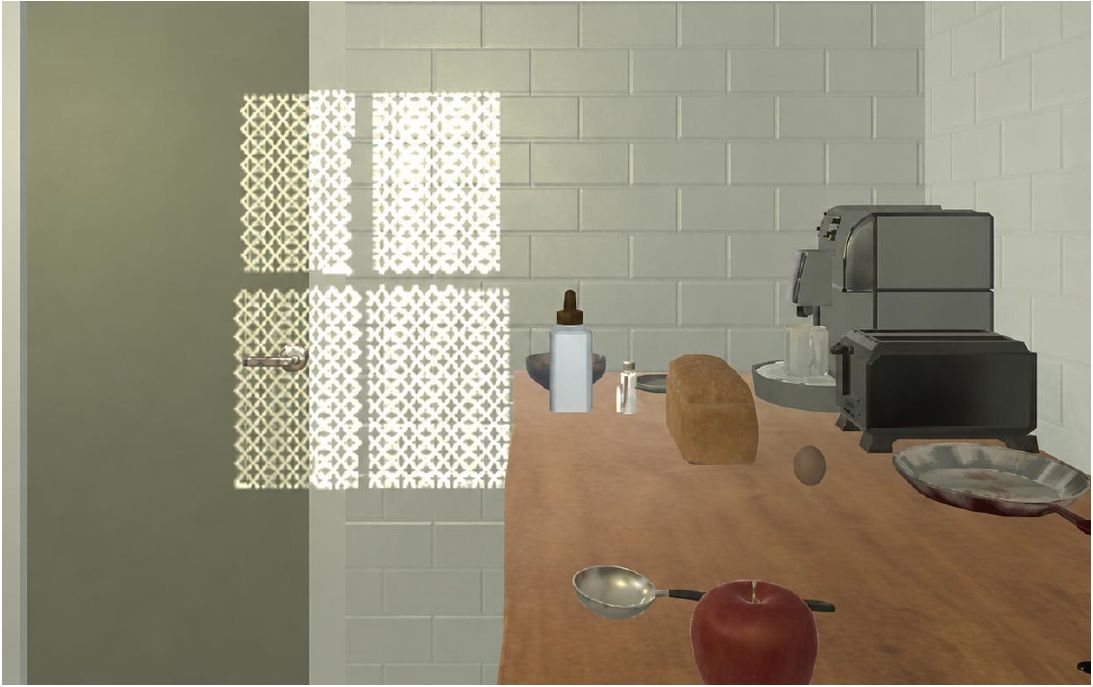}
\end{center}

\begin{repairbeforebox}
\small
\noindent\textbf{Question.} The image shows a kitchen scene with a red apple
near the front edge of the table. With left and right referring to positions
in the image, which utensil is immediately to the left of the red apple?

\noindent\textbf{Options.} (A) Ladle; (B) Spatula; (C) Fork; (D) Knife.

\noindent\textbf{Initial answer.} (B) Spatula.

\noindent\textbf{Evidence basis.} The RGB observation and its registered
object record.
\end{repairbeforebox}

\begin{repairafterbox}
\small
\noindent\textbf{Question.} The image shows a kitchen scene with a red apple
near the front edge of the table. With left and right referring to positions
in the image, which utensil is immediately to the left of the red apple?

\noindent\textbf{Options.} (A) Ladle; (B) Spatula; (C) Fork; (D) Knife.

\noindent\textbf{Repaired answer.} (A) Ladle.

\noindent\textbf{Evidence basis.} The round bowl and long handle visible
beside the apple agree with the registered \texttt{Ladle}; only the Answer
and exact-match scoring key are corrected.
\end{repairafterbox}

\begin{repairsummarybox}
\small
\textbf{Failure:} Answer derivability and evidence
grounding. Both the visible evidence and registered object record identify a
ladle, not a spatula.

\par\smallskip
\textbf{Repair:} Local Field Repair.

\par\smallskip
\textbf{Outcome:} PASS. The RGB view clearly resolves the
target relation; all four options are distinct utensils; Blind Evaluation
receives neither the object record nor the Answer.

\par\smallskip
\textbf{Re-executed scope:} Answer and Scoring Protocol fields
\(\rightarrow\) Requirement Contracts \(\rightarrow\) B1 \(\rightarrow\)
C1. Evidence, template, and binding are retained.
\end{repairsummarybox}

\newpage
\subsubsection{SHN (Habitat): Template Re-binding}

\begin{center}
\includegraphics[width=0.96\linewidth,height=0.40\textheight,keepaspectratio]{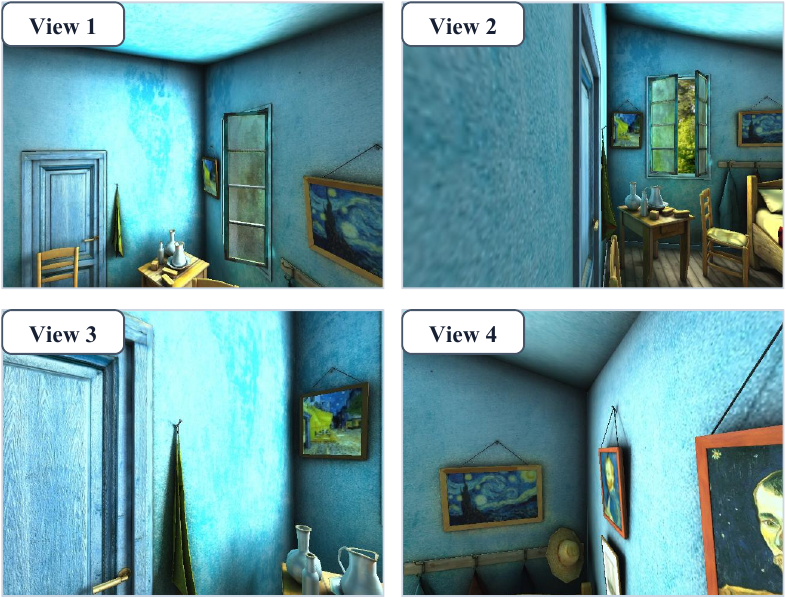}
\end{center}

\begin{repairbeforebox}
\small
\noindent\textbf{Question.} The four numbered images show the same bedroom
from different viewpoints. Which view shows the open window, the small table
beneath it, and part of the bed at the same time?

\noindent\textbf{Options.} (A) View 1; (B) View 2---the only view satisfying
all three conditions; (C) View 3; (D) View 4.

\noindent\textbf{Initial answer.} (B) View 2.

\noindent\textbf{Evidence basis.} Four registered RGB views from the same
Habitat scene.
\end{repairbeforebox}

\begin{repairafterbox}
\small
\noindent\textbf{Question.} The four numbered images show the same bedroom
from different viewpoints. Which view shows the open window, the small table
beneath it, and part of the bed at the same time?

\noindent\textbf{Options.} (A) View 1; (B) View 2; (C) View 3; (D) View 4.

\noindent\textbf{Repaired answer.} (B) View 2.

\noindent\textbf{Evidence basis.} Only View 2 contains the open window, the
table immediately beneath it, and the edge of the bed; every other view
omits at least one condition.
\end{repairafterbox}

\begin{repairsummarybox}
\small
\textbf{Failure:} Target-signal dependence. The
explanatory phrase attached only to option B reveals the answer without
requiring inspection of the four views.

\par\smallskip
\textbf{Repair:} Template Re-binding.

\par\smallskip
\textbf{Outcome:} PASS. All options now have the same
form, and the answer requires jointly checking three visual conditions.

\par\smallskip
\textbf{Re-executed scope:} Template-Evidence Binding
\(\rightarrow\) Candidate Item Synthesis \(\rightarrow\) Requirement
Contracts \(\rightarrow\) B1 \(\rightarrow\) C1. Scene collection and view
registration are retained.
\end{repairsummarybox}

\newpage
\subsubsection{SA (LIBERO): Candidate Re-synthesis}

\begin{center}
\includegraphics[width=0.96\linewidth,height=0.40\textheight,keepaspectratio]{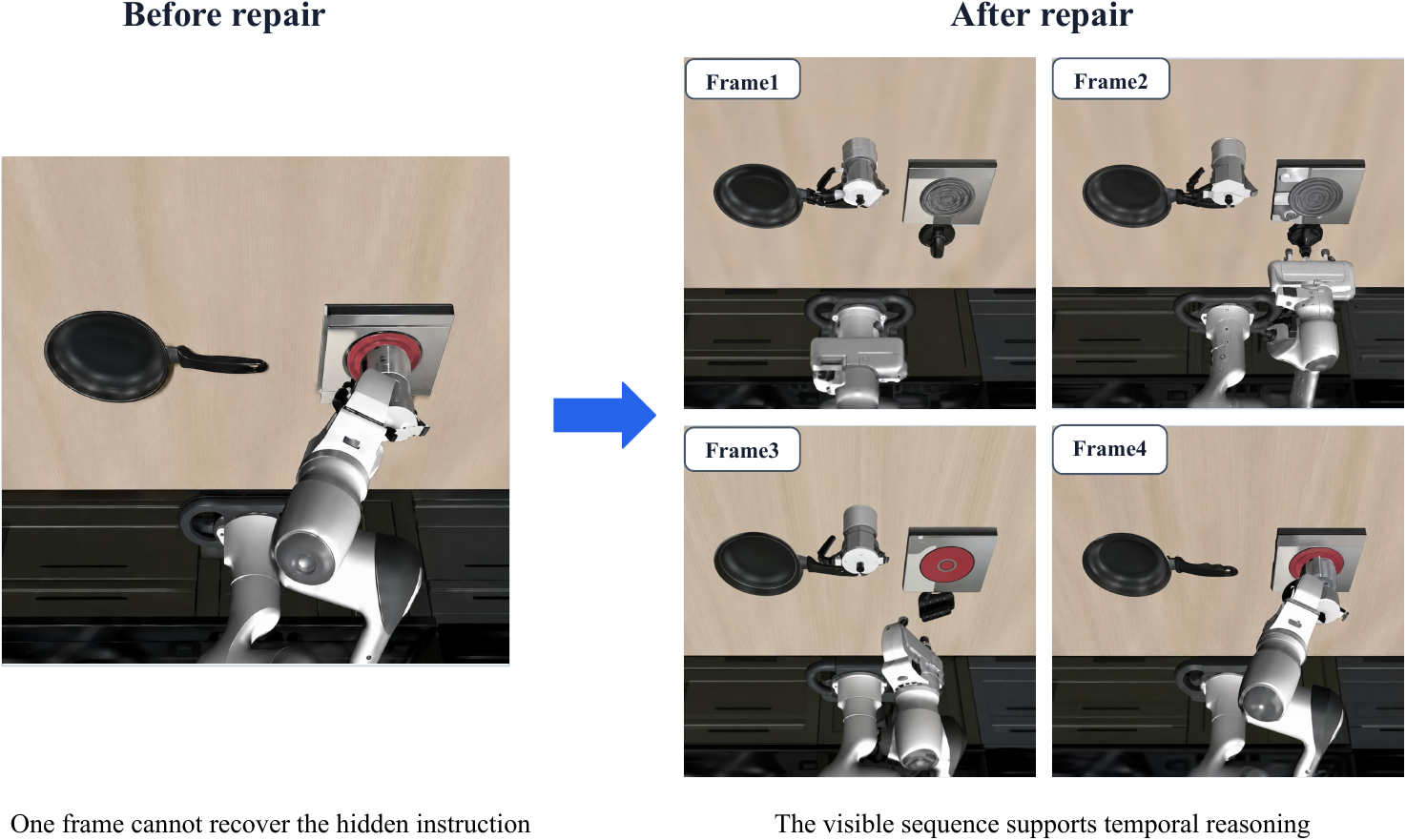}
\end{center}

\begin{repairbeforebox}
\small
\noindent\textbf{Question.} Only the final frame of the robot demonstration
is shown. Based on this image alone, which complete task did the robot most
likely perform?

\noindent\textbf{Options.} (A) Turn on the stove and place the moka pot on
it; (B) Put the black bowl on a plate; (C) Open a drawer and place a utensil
inside; (D) Move the frying pan to the sink.

\noindent\textbf{Initial answer.} (A).

\noindent\textbf{Evidence basis.} A single terminal frame; the stated
answer is derived from task metadata that is not provided to the evaluated
model.
\end{repairbeforebox}

\begin{repairafterbox}
\small
\noindent\textbf{Question.} The four frames show the robot demonstration in
chronological order. Which description best matches the observed order of
burner activation and placement of the moka pot?

\noindent\textbf{Options.} (A) The stove turns on before the moka pot is moved
onto the lit burner; (B) The moka pot is placed first and the stove
turns on afterward; (C) Both changes are already complete before Frame 1;
(D) The stove turns on, but the moka pot never changes position.

\noindent\textbf{Repaired answer.} (A).

\noindent\textbf{Evidence basis.} The burner is unlit in Frames 1 and 2 and
red in Frame 3. The moka pot is then placed on the lit burner in Frame 4.
\end{repairafterbox}

\begin{repairsummarybox}
\small
\textbf{Failure:} Answer derivability and image
dependence. A terminal frame cannot recover the complete task, so the
initial answer depends on hidden task metadata.

\par\smallskip
\textbf{Repair:} Candidate Re-synthesis.

\par\smallskip
\textbf{Outcome:} PASS. The registered frame sequence
supports the temporal relation, and every option describes the same visible
event chain.

\par\smallskip
\textbf{Re-executed scope:} Candidate Item Synthesis
\(\rightarrow\) Requirement Contracts \(\rightarrow\) B1 \(\rightarrow\)
C1. The source demonstration and registered frame order are retained.
\end{repairsummarybox}

\newpage
\subsubsection{SD (CARLA): Skill Re-execution}

\begin{center}
\includegraphics[width=0.96\linewidth,height=0.40\textheight,keepaspectratio]{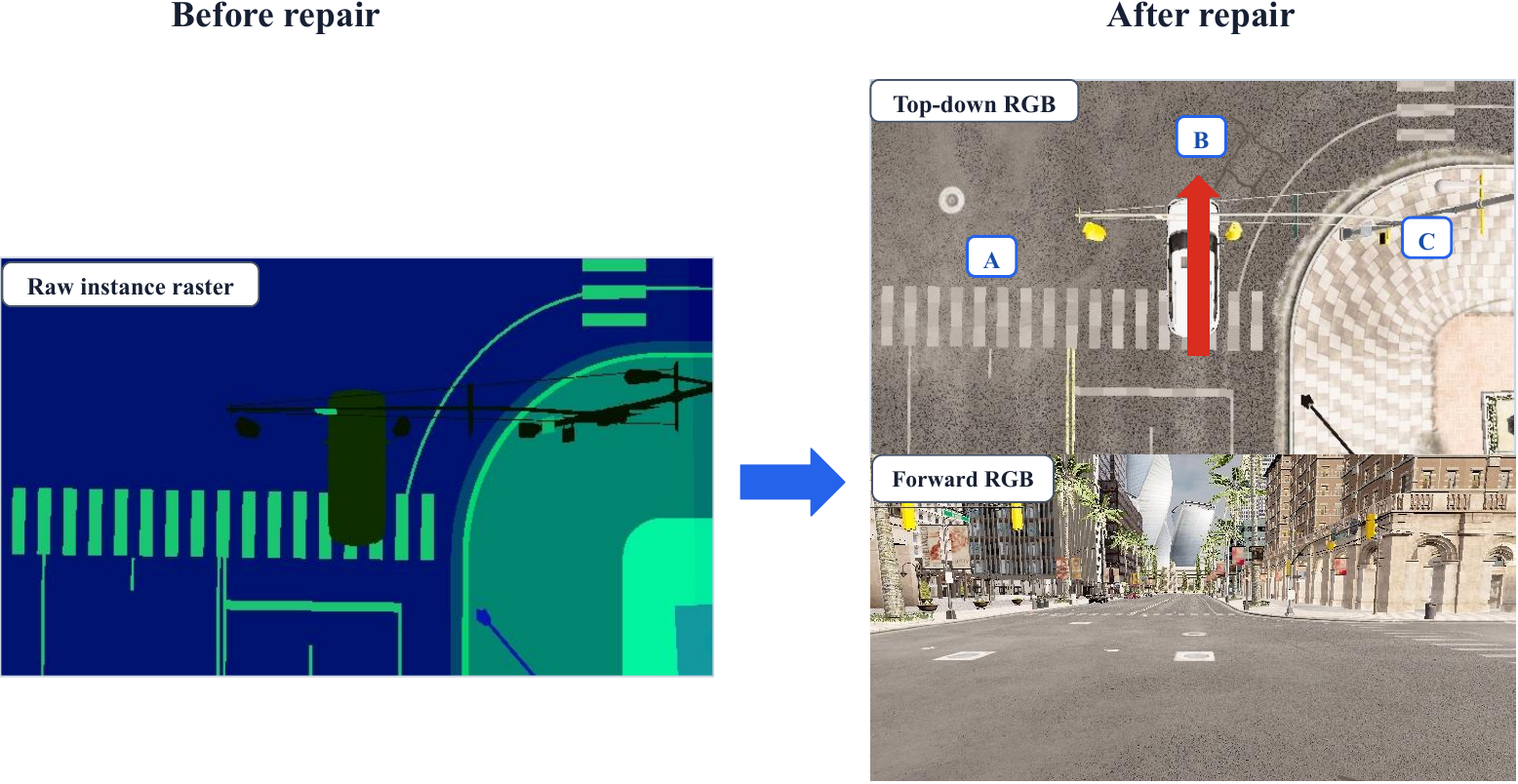}
\end{center}

\begin{repairbeforebox}
\small
\noindent\textbf{Question.} The figure is an internal instance raster whose
colors encode pipeline categories. If the vehicle must turn right at this
intersection, which labeled exit should it take?

\noindent\textbf{Options.} (A) Exit A; (B) Exit B; (C) Exit C; (D) Return
along the approach road.

\noindent\textbf{Initial answer.} (C) Exit C.

\noindent\textbf{Evidence basis.} The internal instance raster and its
paired pose and camera records.
\end{repairbeforebox}

\begin{repairafterbox}
\small
\noindent\textbf{Question.} In the repaired media, the upper panel shows the
intersection from above, and the lower panel shows the driver's forward view.
The red arrow in the upper panel marks the vehicle's heading, and A, B, and C
mark the available exits. If the vehicle turns right, which exit should it
take?

\noindent\textbf{Options.} (A) Exit A; (B) Exit B; (C) Exit C; (D) Return
along the approach road.

\noindent\textbf{Repaired answer.} (C) Exit C.

\noindent\textbf{Evidence basis.} The vehicle is aligned with the red arrow;
from that heading, Exit C is the only branch on the vehicle's right.
\end{repairafterbox}

\begin{repairsummarybox}
\small
\textbf{Failure:} Model-visible answerability and the
processed-RGB media requirement. The initial item exposes an internal
representation and omits the visual anchors needed to interpret A/B/C.

\par\smallskip
\textbf{Repair:} Skill Re-execution, applied to the
visual-anchor construction skill.

\par\smallskip
\textbf{Outcome:} PASS. The released media pair an
aligned top-down RGB view with a forward RGB view, and the route is determined
relative to the registered vehicle heading.

\par\smallskip
\textbf{Re-executed scope:} Visual-anchor construction
\(\rightarrow\) Candidate Item Synthesis \(\rightarrow\) Requirement
Contracts \(\rightarrow\) B1 \(\rightarrow\) C1. Other CARLA records are
retained.
\end{repairsummarybox}

\newpage
\subsubsection{SQ (TartanGround): Upstream Rollback}

\begin{center}
\includegraphics[width=0.96\linewidth,height=0.40\textheight,keepaspectratio]{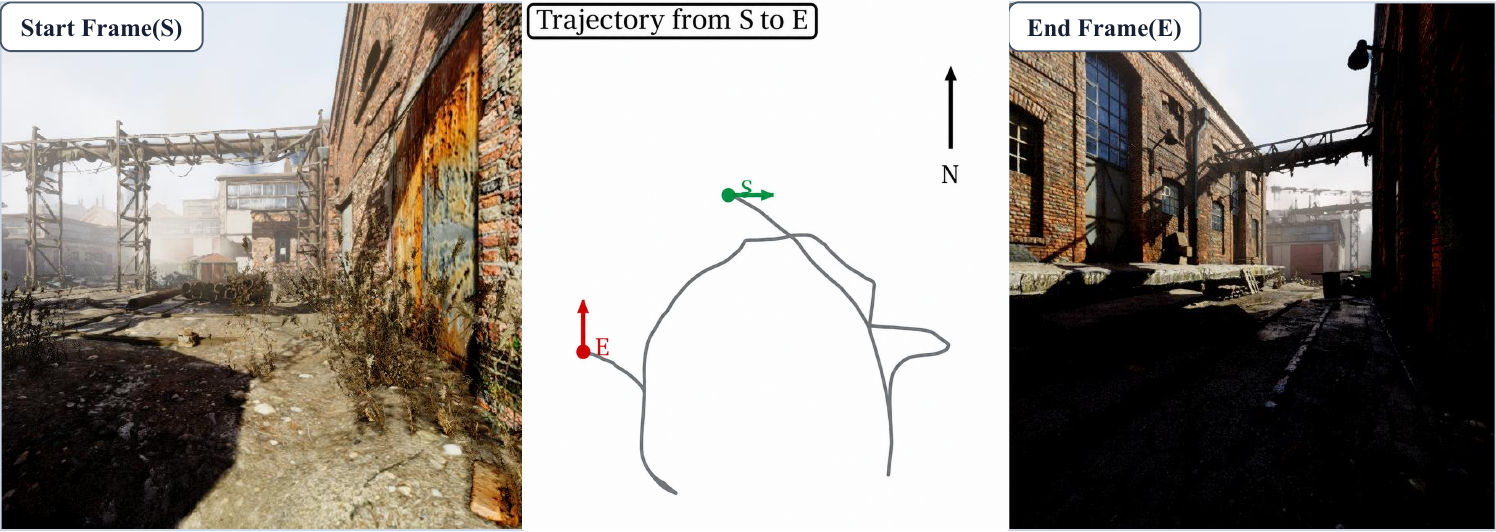}
\end{center}

\begin{repairbeforebox}
\small
\noindent\textbf{Question.} The side panels show the first and last camera
views. The center panel plots the path from start S to end E; N indicates
north, and the small arrows show the nearest cardinal camera heading at S
and E. Where is E relative to S, and which direction is the camera facing at
E?

\noindent\textbf{Options.} (A) Southwest of S, facing west; (B) Southwest
of S, facing north; (C) Southeast of S, facing west; (D) Northwest of S,
facing east.

\noindent\textbf{Initial answer.} (A) Southwest of S, facing west.

\noindent\textbf{Evidence basis.} Start/end images and the trajectory
panel; the initial end arrow is incorrectly joined to intermediate pose 600.
\end{repairbeforebox}

\begin{repairafterbox}
\small
\noindent\textbf{Question.} The side panels show the first and last camera
views. The center panel plots the path from start S to end E; N indicates
north, and the small arrows show the nearest cardinal camera heading at S
and E. Where is E relative to S, and which direction is the camera facing at
E?

\noindent\textbf{Options.} (A) Southwest of S, facing west; (B) Southwest
of S, facing north; (C) Southeast of S, facing west; (D) Northwest of S,
facing east.

\noindent\textbf{Repaired answer.} (B) Southwest of S, facing north.

\noindent\textbf{Evidence basis.} The repaired red arrow uses pose 1,182,
which matches the terminal image. E lies below and to the left of S, while
the final arrow points north.
\end{repairafterbox}

\begin{repairsummarybox}
\small
\textbf{Failure:} Source traceability and answer
derivability. The initial terminal image and terminal heading come from
different indices, so changing the Answer alone cannot restore evidence
consistency.

\par\smallskip
\textbf{Repair:} Upstream Rollback to Evidence and
State Structuring.

\par\smallskip
\textbf{Outcome:} PASS. The registered displacement is
southwest (\(\Delta x=-18.10\), \(\Delta y=-19.11\)), and the terminal
heading is closest to north. Image, pose, and rendered arrow now use the same
terminal index.

\par\smallskip
\textbf{Re-executed scope:} Evidence and State Structuring
\(\rightarrow\) Verified Evidence Pool \(\rightarrow\) Template-Evidence
Binding \(\rightarrow\) Candidate Item Synthesis \(\rightarrow\) affected
gates.
\end{repairsummarybox}

\newpage
\subsubsection{SU (UAV): Final Rejection}

\begin{center}
\includegraphics[width=0.96\linewidth,height=0.40\textheight,keepaspectratio]{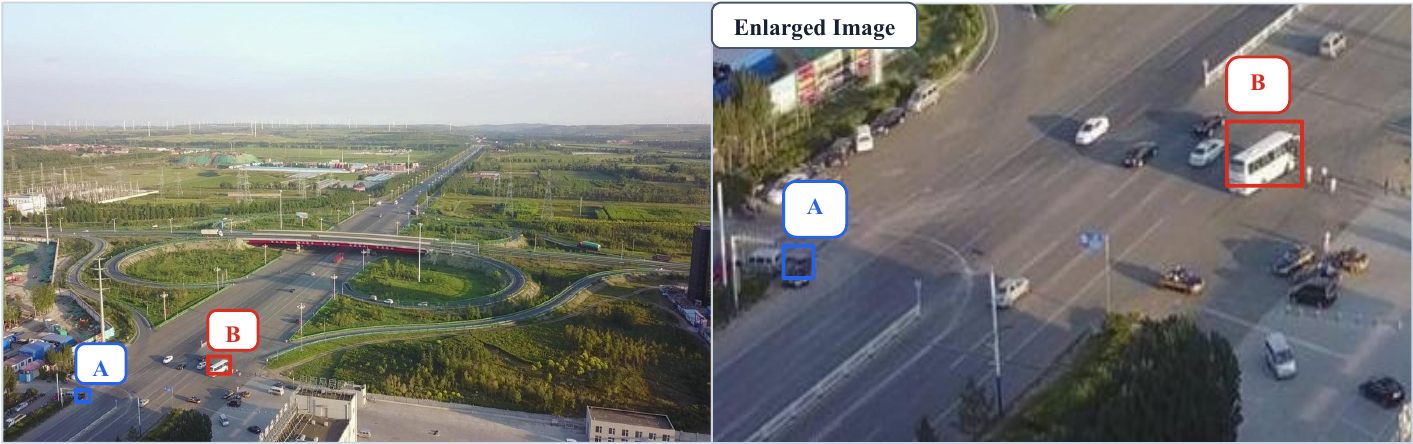}
\end{center}

\begin{repairbeforebox}
\small
\noindent\textbf{Question.} The left panel shows an aerial photograph, and
the right panel enlarges the region containing the two marked vehicles, A and
B. Based only on these images, which vehicle is farther from the camera?

\noindent\textbf{Options.} (A) Vehicle A; (B) Vehicle B.

\noindent\textbf{Initial answer.} (B) Vehicle B.

\noindent\textbf{Evidence basis.} The hidden median-depth records are
29.73~m and 30.92~m, but their detection confidences are only 0.230 and
0.226, both below the 0.25 evidence-admission threshold.
\end{repairbeforebox}

\begin{repairrejectbox}
\small
\noindent\textbf{Outcome.} The marked targets are not resolved consistently
enough in the source view and the enlarged crop to support a reliable distance
comparison. The candidate lineage is terminated, and no item is released.
\end{repairrejectbox}

\begin{repairsummarybox}[Rejection Summary]
\small
\textbf{Failure:} Evidence admissibility and human
answerability. The wording and A/B markers are clear, but the visible media
do not support a unique, reliable distance judgment; hidden depth cannot
substitute for admissible model-visible evidence.

\par\smallskip
\textbf{Decision:} Final Rejection.

\par\smallskip
\textbf{Outcome:} REJECTED. Neither target record
reaches the admission threshold, and changing the Answer or wording cannot
repair insufficient evidence.

\par\smallskip
\textbf{Re-executed scope:} Final Rejection only; the source image
and unrelated verified UAV evidence are retained.
\end{repairsummarybox}

\newpage
\subsection{Interactive Embodied Track: Execution and Verification}
\label{app:interactive_track}

The second benchmark track evaluates executable embodied interaction.
This section specifies the IE-Track composition, execution protocol,
metrics, representative task packages, GPT-5.5 execution records, and
construction-time verification and repair cases.
The protocol description applies uniformly to all evaluated models, while
the examples distinguish constructed task artifacts from target-model
execution trajectories.

\subsubsection{Track Composition and Evaluation Protocol}
\label{app:interactive_track_details}

The IE-Track contains \(220\) independent task instances constructed from
\(28\) AI2-THOR scenes across four household scene types, as summarized in
Table~\ref{tab:interactive_track_composition}.
Each task is evaluated as a separate episode with an independently
initialized environment state and does not depend on any preceding task.
All evaluated models receive the same \(220\) tasks, scene pool, initial
states, action interface, stopping rules, and evaluation protocol.

\begin{table}[!ht]
\centering
\small
\setlength{\tabcolsep}{5.0pt}
\caption{
Composition of the IE-Track by household scene and task difficulty.
}
\label{tab:interactive_track_composition}
\begin{tabular}{lrrrr}
\toprule
\textbf{Scene Type}
& \textbf{Simple}
& \textbf{Hard}
& \textbf{Total}
& \textbf{Scenes} \\
\midrule
Kitchen     & 28 & 27 & 55 & 8 \\
Living room & 27 & 28 & 55 & 7 \\
Bedroom     & 28 & 27 & 55 & 7 \\
Bathroom    & 27 & 28 & 55 & 6 \\
\midrule
\textbf{Total}
& \textbf{110}
& \textbf{110}
& \textbf{220}
& \textbf{28} \\
\bottomrule
\end{tabular}
\end{table}

Task success is determined by the executable terminal-state evaluator.
An episode is successful only when the required goal conditions are
satisfied; partial task progress is recorded separately and is not counted
as success.
The action interface supports camera rotation, directional movement,
looking up or down, object-directed navigation, object pickup and
placement, opening and closing, turning objects on or off, slicing,
dropping, waypoint operations, and episode termination.
Because actions are grounded to specific object instances and receptacles,
the evaluation records contain \(746\) instantiated action expressions.

Each episode is limited to \(150\) steps.
Invalid actions are recorded but do not immediately terminate execution;
the agent may continue until task completion, explicit termination, or the
step limit.
An episode that reaches the step limit without satisfying the goal
conditions is treated as a timeout failure.
All metrics are computed directly from the saved episode-level execution
records without manual correction or score imputation.

Trajectory-grounded EQA is performed after episode execution using the
complete interaction history.
Questions may refer to the final state, an earlier trajectory state, or
the initial environment state.
Each question has a reference answer, a model answer, and a binary
correctness label, and EQA accuracy is computed as a question-level micro
average.

Let \(N\) denote the number of evaluated episodes, \(s_i\in\{0,1\}\)
indicate whether episode \(i\) is successfully completed, \(Q_i\) denote
the number of EQA questions associated with episode \(i\),
\(c_{iq}\in\{0,1\}\) indicate whether question \(q\) is answered
correctly, and \(n_i\) denote the number of executed steps.
The reported metrics are

\[
\mathrm{SR}
=
\frac{1}{N}\sum_{i=1}^{N}s_i,
\]

\[
\mathrm{EQA\ Accuracy}
=
\frac{\sum_{i=1}^{N}\sum_{q=1}^{Q_i} c_{iq}}
     {\sum_{i=1}^{N}Q_i},
\]

\[
\mathrm{SRL}
=
\frac{\sum_{i=1}^{N}\sum_{q=1}^{Q_i}s_i c_{iq}}
     {\sum_{i=1}^{N}Q_i},
\]

and

\[
\mathrm{Avg.\ Steps}
=
\frac{1}{N}\sum_{i=1}^{N}n_i.
\]

SRL therefore measures joint task success and trajectory-EQA correctness,
while Avg.\ Steps is computed over all episodes, including successful,
failed, and timeout episodes.

\subsubsection{Representative Task Packages and GPT-5.5 Executions}
\label{app:interactive_examples}

Each IE-Track task contains an instruction, a restorable initial state,
an action interface, a goal condition, and an executable terminal-state
verifier.
This section presents selected fields from two representative task
packages together with the corresponding GPT-5.5 execution records
produced during S5 Verification \& Evaluation Reporting.
The construction-time reference trajectories used to verify task
executability are not repeated here.
The common action space, stopping rules, and evaluation metrics are
defined in Section~\ref{app:interactive_track_details}.

\begin{table*}[!ht]
\centering
\small
\setlength{\tabcolsep}{4.5pt}
\caption{
Representative IE-Track task packages and their GPT-5.5 execution
outcomes.
Success is determined by the executable terminal-state verifier, and EQA
reports the number of correctly answered trajectory-grounded questions.
}
\label{tab:app_interactive_examples}
\begin{tabular}{@{}lllrrrrr@{}}
\toprule
\textbf{Task ID}
& \textbf{Split}
& \textbf{Scene}
& \textbf{Subtasks}
& \textbf{Steps}
& \textbf{Invalid}
& \textbf{Success}
& \textbf{EQA} \\
\midrule
Kitchen27-00-slice-tomato
& \texttt{short\_hard}
& Kitchen
& 1 & 7 & 0 & 1 & 10/15 \\

Bathroom403-00-wash-stow
& \texttt{long\_hard}
& Bathroom
& 4 & 60 & 4 & 1 & 9/15 \\
\bottomrule
\end{tabular}
\end{table*}

\paragraph{Single-Subtask Slice-and-Place Task.}
\label{app:interactive_simple_example}

\noindent\textbf{Selected task-package fields.}
The task requires the agent to slice a tomato and place a tomato slice
on the countertop.
The complete package additionally stores the simulator object poses,
restorable state, action interface, and compiled terminal verifier.

\begin{figure*}[!ht]
\centering
\begin{minipage}{0.78\textwidth}

\begin{artifactbox}[Single-Subtask IE-Track Artifact]
task_id: Kitchen27-00-slice-tomato
eval_set: short_hard

instructions:
  micro: Pick up the tomato from the dining table,
    slice it, and put the slices on the counter top.

scene:
  floor_plan: FloorPlan27
  room_type: Kitchen
  random_spawn:
    seed: 0
    object_counts: {Tomato: 1}

pddl_params:
  object_target: Tomato
  parent_target: CounterTop
  object_sliced: true
  subtasks:
    - task_type: pick_slice_then_place_in_recep
      pddl_params:
        object_target: Tomato
        parent_target: CounterTop
        object_sliced: true

completion:
  mode: all
\end{artifactbox}

\end{minipage}

\caption{
Representative single-subtask IE-Track artifact produced by
Embodied-BenchForge.
The artifact specifies the natural-language instruction, scene
initialization, structured goal conditions, and completion rule for an
executable slice-and-place task.
}
\label{fig:ie_single_subtask_artifact}
\end{figure*}

\begin{figure*}[!ht]
\centering

\apptrajframefour
{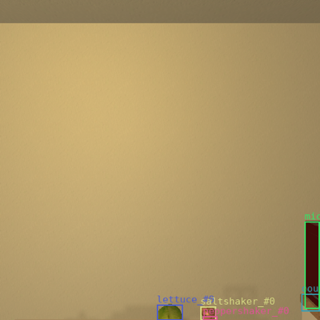}
{0}
{Restored initial state.}
\hfill
\apptrajframefour
{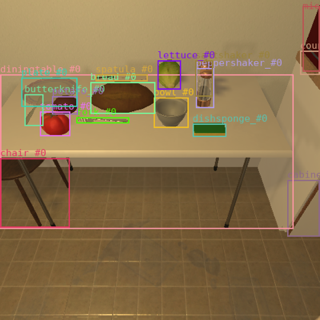}
{1}
{Look down.}
\hfill
\apptrajframefour
{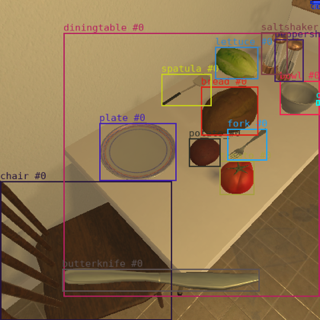}
{2}
{Pick up the butter knife.}
\hfill
\apptrajframefour
{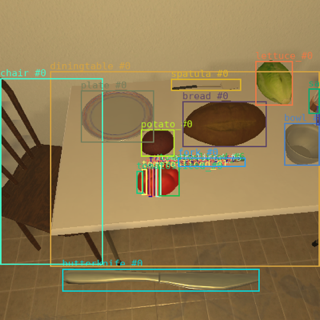}
{3}
{Slice the tomato.}

\par\smallskip

\apptrajframefour
{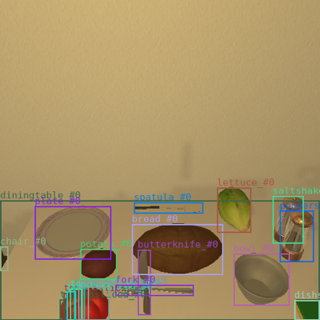}
{4}
{Put down the knife.}
\hfill
\apptrajframefour
{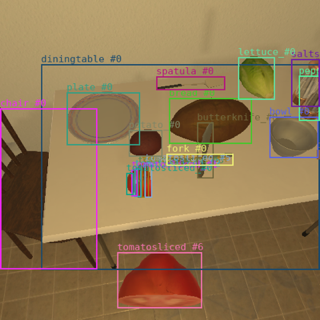}
{5}
{Pick up a tomato slice.}
\hfill
\apptrajframefour
{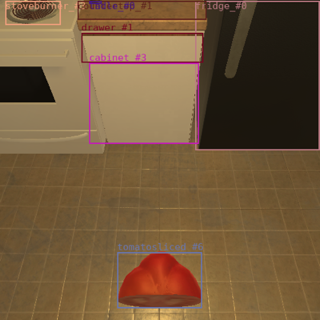}
{6}
{Turn toward the counter.}
\hfill
\apptrajframefour
{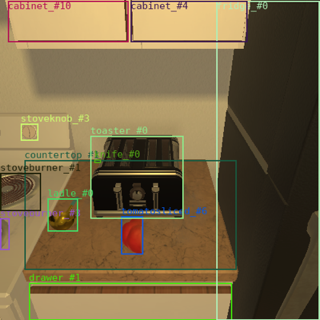}
{7}
{Place the slice on the counter.}

\caption{
Complete GPT-5.5 execution trajectory for the single-subtask
slice-and-place task.
Step~0 shows the restored initial state, and Steps~1--7 show the
observations after the actions selected by GPT-5.5.
Task success is determined by the executable terminal-state verifier.
}
\label{fig:app_simple_execution}
\end{figure*}

\noindent\textbf{GPT-5.5 execution.}
Figure~\ref{fig:ie_single_subtask_artifact} shows the corresponding
single-subtask IE-Track artifact, including its instruction, scene
configuration, structured goal conditions, and completion rule.
GPT-5.5 completes the task in seven actions without invalid actions.
The executable terminal-state verifier records the episode as successful.
The retained trajectory also yields 10 correct answers among 15
trajectory-grounded EQA questions.
Figure~\ref{fig:app_simple_execution} shows the restored initial state
and the complete GPT-5.5 execution trajectory.

\paragraph{Multi-Subtask Wash-and-Stow Task.}
\label{app:interactive_hard_example}

\noindent\textbf{Selected task-package fields.}
The task contains four goal predicates: rinsing and placing the soap bar
and cloth on the countertop, and placing the scrub brush and plunger on
the countertop.

\begin{figure*}[!ht]
\centering
\begin{minipage}{0.82\textwidth}

\begin{artifactbox}[Multi-Subtask IE-Track Artifact]
task_id: Bathroom403-00-wash-stow
eval_set: long_hard

instructions:
  micro: rinse the soapbar at the sink and put it
    on the counter top; rinse the cloth at the sink
    and put it on the counter top; put the scrubbrush
    on the counter top; put the plunger on the
    counter top

scene:
  floor_plan: FloorPlan403
  room_type: Bathroom
  random_spawn:
    seed: 0
    object_counts:
      {SoapBar: 1, Cloth: 1, ScrubBrush: 1, Plunger: 1}

pddl_params:
  subtasks:
    - task_type: pick_clean_then_place_in_recep
      pddl_params:
        {object_target: SoapBar,
         parent_target: CounterTop}

    - task_type: pick_clean_then_place_in_recep
      pddl_params:
        {object_target: Cloth,
         parent_target: CounterTop}

    - task_type: pick_and_place_simple
      pddl_params:
        {object_target: ScrubBrush,
         parent_target: CounterTop}

    - task_type: pick_and_place_simple
      pddl_params:
        {object_target: Plunger,
         parent_target: CounterTop}

completion:
  mode: all
\end{artifactbox}

\end{minipage}

\caption{
Representative multi-subtask IE-Track artifact produced by
Embodied-BenchForge.
The artifact defines four object-state and placement goals that must all
be satisfied by the executable terminal-state verifier.
}
\label{fig:ie_multi_subtask_artifact}
\end{figure*}

\begin{figure*}[!ht]
\centering

\apptrajframefour
{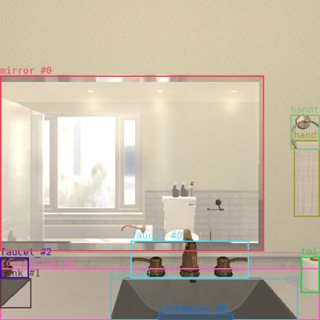}
{0}
{Restored initial state.}
\hfill
\apptrajframefour
{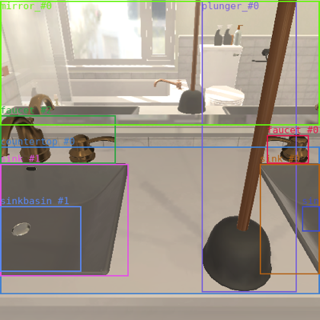}
{8}
{Place the plunger.}
\hfill
\apptrajframefour
{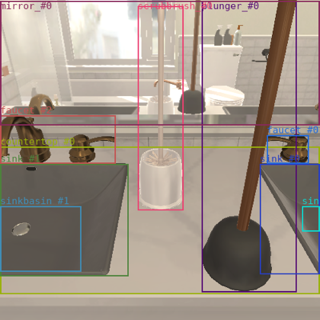}
{16}
{Place the scrub brush.}
\hfill
\apptrajframefour
{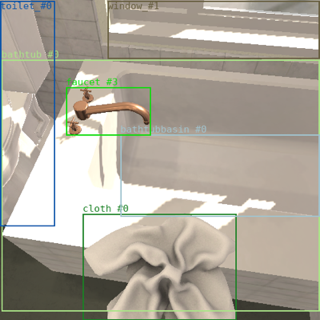}
{25}
{Pick up the cloth.}

\par\smallskip

\apptrajframefour
{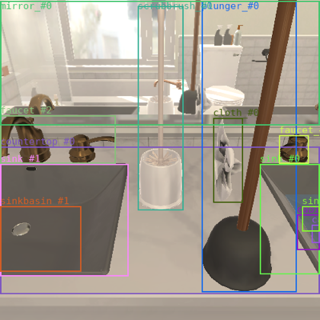}
{31}
{Place the rinsed cloth.}
\hfill
\apptrajframefour
{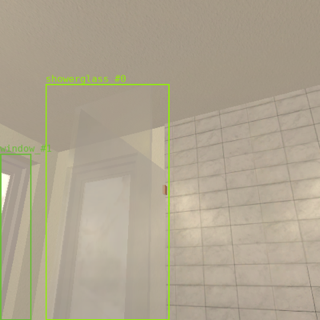}
{45}
{Search near the shower.}
\hfill
\apptrajframefour
{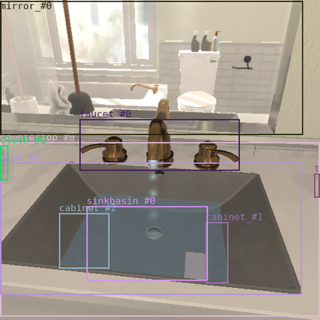}
{51}
{Rinse the soap bar.}
\hfill
\apptrajframefour
{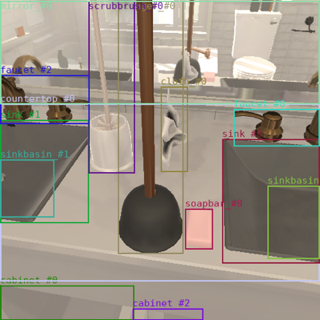}
{60}
{Place the soap bar on the counter.}

\caption{
Selected observations from the 60-step GPT-5.5 execution trajectory for
the multi-subtask wash-and-stow task.
The sequence covers object placement, cleaning, continued search, and
completion of all four goal conditions.
Task success is determined by the executable terminal-state verifier.
}
\label{fig:app_hard_execution}
\end{figure*}

\noindent\textbf{GPT-5.5 execution.}
Figure~\ref{fig:ie_multi_subtask_artifact} shows the corresponding
multi-subtask IE-Track artifact, which defines four goal conditions that
must all be satisfied.
GPT-5.5 completes all four subtasks in 60 actions.
Four invalid actions are recorded, but they do not terminate the episode;
the model continues execution and eventually satisfies all goal
conditions.
The executable terminal-state verifier records the episode as successful,
and the retained trajectory yields 9 correct answers among 15
trajectory-grounded EQA questions.
Figure~\ref{fig:app_hard_execution} shows representative states from the
GPT-5.5 execution trajectory.

These examples distinguish the constructed benchmark artifact from the
target-model execution record.
The task package defines the initial state, available actions, goal
conditions, and executable verifier, whereas S5 records the actions,
terminal outcome, and trajectory-grounded EQA responses produced by
GPT-5.5.
Both episodes are successfully completed, but their EQA results remain
imperfect, illustrating that closed-loop execution and trajectory
understanding provide complementary evaluation signals.


\subsubsection{Representative Verification and Repair Cases}
\label{app:ie_repair_cases}

Table~\ref{tab:ie_repair_case_conflict} and
Table~\ref{tab:ie_repair_case_placeholder} present two representative
construction failures detected in the IE-Track.
The five audited candidates contain two recurring failure types:
instruction--goal inconsistency and underspecified placeholder
instructions.
We therefore report one representative case for each type rather than
listing multiple instances of the same defect.

The left side of each table shows the invalid candidate, the center
summarizes the blocking requirement and repair route, and the right side
shows the repaired candidate.
Only fields related to the failure are displayed; unchanged scene and
initial-state fields are omitted.
A repaired candidate is admitted into the benchmark only after passing
Task Validity Check, Task Completion Test, and Initial-State Check again.

Kitchen21-00-hot-meal exhibits the same instruction--goal inconsistency
as the first case, while LivingRoom230-07-armchair and
LivingRoom230-07-sofa2 exhibit the same placeholder-instruction defect as
the second case.
These repeated failures are handled at the shared template or synthesis
operation when provenance indicates a common origin, rather than repaired
only as isolated task files.

The repaired candidates must re-enter the complete IE verification path.
Task Validity Check verifies instruction--action--goal consistency,
Task Completion Test confirms that a valid reference execution satisfies
the terminal verifier, and Initial-State Check confirms reproducible
state restoration.
Only after these checks pass are the affected verifier, reference
trajectory, model-execution records, trajectory-grounded EQA results, and
aggregate metrics regenerated.

\begin{table*}[!ht]
\centering
\setlength{\tabcolsep}{3.0pt}
\renewcommand{\arraystretch}{1.08}
\caption{
Representative repair of an instruction--goal inconsistency.
In the original candidate, all four destination assignments differ
between the model-visible instruction and the executable PDDL goal.
The repaired candidate aligns the PDDL goal with the mutually consistent
macro and micro instructions.
Acceptance still requires execution-based revalidation.
}
\label{tab:ie_repair_case_conflict}
\begin{tabularx}{\textwidth}{
@{}>{\raggedright\arraybackslash}p{0.34\textwidth}
>{\raggedright\arraybackslash}p{0.25\textwidth}
>{\raggedright\arraybackslash}X@{}}
\toprule
\textbf{Invalid Candidate}
& \textbf{Detection and Repair}
& \textbf{Repaired Candidate} \\
\midrule

\textbf{Kitchen26-00-hot-meal}

\medskip
\ttfamily
instruction goals:\newline
Apple $\rightarrow$ DiningTable\newline
Potato $\rightarrow$ CounterTop\newline
Egg $\rightarrow$ CounterTop\newline
ButterKnife $\rightarrow$ CounterTop

\medskip
PDDL goals:\newline
Apple $\rightarrow$ CounterTop\newline
Potato $\rightarrow$ DiningTable\newline
Egg $\rightarrow$ DiningTable\newline
ButterKnife $\rightarrow$ DiningTable
\normalfont
&
\textbf{Blocking requirement:}

Instruction, goal conditions, and terminal verifier must describe the
same intended task.

\medskip
\textbf{Detection:}

The normalized subgoal sets extracted from the instruction and PDDL do
not match.
A GPT-5.5 execution following the instruction contains no invalid action
but cannot satisfy the hidden goal.

\medskip
\textbf{Repair route:}

Roll back to task-specification synthesis, align the PDDL destinations
with the instruction, recompile the verifier, and invalidate the old
reference trajectory and evaluation records.
&
\ttfamily
task\_id: Kitchen26-00-hot-meal

instructions:\newline
\ \ micro: Heat the apple and put it on the\newline
\ \ \ \ dining table; slice the potato and\newline
\ \ \ \ put it on the counter top; put the\newline
\ \ \ \ egg and butter knife on the counter top.

pddl\_params:\newline
\ \ subtasks:\newline
\ \ \ \ - Apple, heated, DiningTable\newline
\ \ \ \ - Potato, sliced, CounterTop\newline
\ \ \ \ - Egg, placed, CounterTop\newline
\ \ \ \ - ButterKnife, placed, CounterTop

completion:\newline
\ \ mode: all
\normalfont
\\

\bottomrule
\end{tabularx}
\end{table*}

\begin{table*}[!ht]
\centering
\setlength{\tabcolsep}{3.0pt}
\renewcommand{\arraystretch}{1.08}
\caption{
Representative repair of an underspecified IE-Track instruction.
The original candidate exposes only a placeholder instruction, whereas
the operation target appears exclusively in the hidden PDDL goal.
The repair reconstructs an explicit model-visible instruction and
corrects the horizon split.
Acceptance still requires execution-based revalidation.
}
\label{tab:ie_repair_case_placeholder}
\begin{tabularx}{\textwidth}{
@{}>{\raggedright\arraybackslash}p{0.34\textwidth}
>{\raggedright\arraybackslash}p{0.25\textwidth}
>{\raggedright\arraybackslash}X@{}}
\toprule
\textbf{Invalid Candidate}
& \textbf{Detection and Repair}
& \textbf{Repaired Candidate} \\
\midrule

\textbf{Bedroom322-07-bed}

\medskip
\ttfamily
eval\_set: long\_simple

instructions:\newline
\ \ macro: Test Bed.\newline
\ \ micro: Test Bed.

pddl\_params:\newline
\ \ object\_target: Book\newline
\ \ parent\_target: Bed\newline
\ \ subtasks:\newline
\ \ \ \ - pick\_and\_place\_simple
\normalfont
&
\textbf{Blocking requirements:}

The instruction must explicitly state the required object, operation,
and destination.
The split must also be consistent with the number of subtasks.

\medskip
\textbf{Detection:}

The placeholder instruction omits both the Book and the required
pick-and-place operation.
The task contains one subtask but is assigned to a long-horizon split.

\medskip
\textbf{Repair route:}

Roll back to instruction synthesis, reconstruct the instruction from the
verified goal specification, and move the task to the corresponding
short split.
&
\ttfamily
task\_id: Bedroom322-07-bed\newline
eval\_set: short\_simple

instructions:\newline
\ \ macro: Please place the book on the bed.\newline
\ \ micro: Pick up the book and put it on\newline
\ \ \ \ the bed.

pddl\_params:\newline
\ \ object\_target: Book\newline
\ \ parent\_target: Bed\newline
\ \ subtasks:\newline
\ \ \ \ - task\_type: pick\_and\_place\_simple\newline
\ \ \ \ \ \ object\_target: Book\newline
\ \ \ \ \ \ parent\_target: Bed

completion:\newline
\ \ mode: all
\normalfont
\\

\bottomrule
\end{tabularx}
\end{table*}

\newpage
\onecolumn
\section{Resource Coverage and Cross-Resource Adaptation}
\label{app:resource_coverage}

This appendix examines the cross-resource extensibility of
Embodied-BenchForge.
The extensibility claim concerns the reuse of a common artifact schema,
skill-orchestration mechanism, and verification loop across heterogeneous
embodied resources with limited resource-specific adaptation; it does not
assume zero-shot compatibility with an arbitrary unseen simulator.
As summarized in Table~\ref{tab:resource_coverage}, the evaluated
construction settings span indoor household tasks, indoor navigation,
robotic manipulation, autonomous driving, real aerial imagery,
ground-robot trajectories, and interactive household environments.
They further cover simulator-derived and real-world resources, static and
sequential observations, privileged and non-privileged supervision, and
both offline question answering and online task execution.

\begin{table*}[!ht]
\centering
\small
\setlength{\tabcolsep}{3.0pt}
\renewcommand{\arraystretch}{1.04}
\caption{
Resource coverage and cross-resource adaptation in the evaluated
Embodied-BenchForge construction runs.
\textit{Specific Skills/Adapters} counts definitions used only in the
corresponding setting, whereas the reuse rate measures the proportion
shared with at least one other benchmark setting.
}
\label{tab:resource_coverage}
\begin{tabularx}{\textwidth}{
@{}p{2.50cm}
p{2.45cm}
>{\raggedright\arraybackslash}X
p{2.75cm}
cc@{}}
\toprule
\textbf{Resource}
& \textbf{Embodied Setting}
& \textbf{Resource Interface}
& \textbf{Constructed Benchmark}
& \textbf{\makecell{Specific Skills /\\Adapters}}
& \textbf{\makecell{Reuse\\Rate}} \\
\midrule

ALFRED
&
Indoor household tasks
&
Task trajectories, visual observations, object states, and household
task semantics
&
SpatialHome-Task, 10,000 OE items
&
1
&
97.6\% \\

Habitat
&
Indoor robot navigation
&
Egocentric observations, camera poses, navigation trajectories, and
spatial scene information
&
SpatialHome-Nav, 10,000 OE items
&
1
&
97.5\% \\

CARLA
&
Autonomous driving
&
Driving observations, temporal sequences, vehicle motion, and
traffic-scene state
&
SpatialDrive, 10,000 OE items
&
1
&
97.5\% \\

LIBERO
&
Robotic manipulation
&
Manipulation trajectories, object states, end-effector actions, and
task semantics
&
SpatialArm, 10,000 OE items
&
1
&
97.5\% \\

Real aerial imagery
&
UAV-view understanding
&
Unlabeled real-world RGB images without privileged simulator states
&
SpatialUAV, 10,000 OE items
&
7
&
82.9\% \\

TartanGround
&
Ground and quadruped robots
&
Multi-view robot trajectories with RGB, depth, semantic, and pose
information
&
SpatialQuadruped, 10,000 OE items
&
5
&
87.8\% \\

AI2-THOR
&
Interactive household execution
&
Executable simulator states, object interactions, action interfaces, and
terminal conditions
&
Interactive Embodied Track, 220 tasks
&
8
&
81.8\% \\

\bottomrule
\end{tabularx}
\end{table*}

The first four simulator- or dataset-based OE construction runs require
only one setting-specific skill or adapter each and reuse more than
\(97\%\) of their invoked skill definitions.
More substantial adaptation is required for real aerial imagery,
quadruped-robot trajectories, and interactive task execution, because
these settings introduce distinct evidence interfaces or executable
environment operations.
Nevertheless, SpatialUAV, SpatialQuadruped, and the IE-Track still reuse
\(82.9\%\), \(87.8\%\), and \(81.8\%\) of their invoked skills,
respectively.
These results indicate that extending the framework to a heterogeneous
resource primarily requires localized adapters and task-specific
operations rather than redesigning the complete construction workflow.

Two representative cases further illustrate how the shared workflow is
adapted to resources whose original interfaces differ substantially from
embodied spatial question answering.
For \textbf{SpatialUAV}, Embodied-BenchForge operates on unlabeled
real-world aerial images without privileged simulator states or
pre-existing benchmark annotations.
It structures answer-bearing visual and geometric evidence from the
images, binds the resulting evidence to capability-specific templates,
and constructs 10,000 grounded questions covering recognition, counting,
localization, scale, depth, box geometry, and category-constrained
matching.
This case demonstrates that the construction workflow can move beyond
simulator-native state and operate on externally collected visual
resources.

For \textbf{SpatialQuadruped}, Embodied-BenchForge repurposes
TartanGround, a multimodal resource originally designed for ground-robot
perception and navigation, with evaluations centered on occupancy
prediction and visual odometry/SLAM.
The framework uses its quadruped trajectories, synchronized multi-view
observations, semantic information, depth, and camera poses to construct
language-grounded questions involving semantic visibility, image-region
localization, depth ordering, cross-view consistency, trajectory
structure, and motion-orientation change.
This adaptation changes the evaluation interface from sensor-level
prediction and state estimation to evidence-bound embodied spatial
question answering while retaining \(87.8\%\) skill reuse.

Taken together, the evaluated settings provide evidence for
cross-resource extensibility through both broad resource coverage and
high skill reuse.
The same construction backbone is applied across heterogeneous embodied
resources, while resource-specific differences are isolated in a limited
number of adapters and specialized skills.

\newpage
\onecolumn

\section{Skill Taxonomy and Reuse Accounting}
\label{app:skill_reuse}

This appendix details the organization and reuse of the Skill Library. It
first groups the registered skill/card definitions into representative
functional families covering artifact synthesis, verification, repair, and
track-specific diagnosis, and then defines the counting protocol used to
measure shared and setting-specific skill usage across construction runs.

\subsection{Representative Skill Taxonomy}
\label{app:skill_taxonomy}

The analyzed implementation snapshot contains 64 registered skill/card
definitions.
For readability, Table~\ref{tab:core_skill_taxonomy} groups these
definitions into representative functional families corresponding to the
forward construction path and the backward verification-and-repair loop
in Figures~1--2.
The rows summarize the functions implemented by one or more registered
skills rather than the exact names of individual skill cards.
The complete 64-entry inventory is provided in the accompanying code
snapshot.
Inputs and outputs are expressed using the paper-level artifact types.
The backend column describes the primary implementation category:
an LLM, deterministic program, simulator API, resource adapter,
scoring/verifier compiler, or a hybrid of these components.
All 64 registered definitions, including source- and track-specific
adapters omitted from the summary table, remain included in the reuse
statistics.

\begin{table*}[!ht]
\centering
\small
\setlength{\tabcolsep}{5.0pt}
\renewcommand{\arraystretch}{1.10}
\caption{
Detailed skill-reuse statistics for the six OE-Track benchmarks and the
IE-Track.
\textit{Skills Used} counts distinct skill/card definitions invoked in
one construction run, with repeated retries, re-executions, and
repair-time invocations counted once.
\textit{Reused Skills} are shared with at least one other benchmark
setting, whereas \textit{Specific Skills/Adapters} are used only in the
corresponding setting.
The overall row is computed from pooled benchmark--skill incidence
counts.
}
\label{tab:app_skill_reuse_details}
\begin{tabular}{@{}lcccc@{}}
\toprule
\textbf{Benchmark}
& \textbf{\makecell{Skills\\Used}}
& \textbf{\makecell{Reused\\Skills}}
& \textbf{\makecell{Specific Skills /\\Adapters}}
& \textbf{\makecell{Reuse\\Rate}} \\
\midrule
SHT      & 41 & 40 & 1 & 97.6\% \\
SHN      & 40 & 39 & 1 & 97.5\% \\
SD       & 40 & 39 & 1 & 97.5\% \\
SA       & 40 & 39 & 1 & 97.5\% \\
SU       & 41 & 34 & 7 & 82.9\% \\
SQ       & 41 & 36 & 5 & 87.8\% \\
IE-Track & 44 & 36 & 8 & 81.8\% \\
\midrule
\textbf{Mean / pooled}
& \textbf{41.0}
& \textbf{37.6}
& \textbf{3.4}
& \textbf{91.6\%} \\
\bottomrule
\end{tabular}
\end{table*}

\begin{table*}[!ht]
\centering
\footnotesize
\setlength{\tabcolsep}{2.0pt}
\renewcommand{\arraystretch}{1.08}
\caption{
Representative skill families in Embodied-BenchForge.
The rows summarize functional groups rather than individual registered
skill/card names.
Inputs and outputs use the paper-level artifact types.
\textit{All} denotes SHT, SHN, SD, SA, SU, SQ, and IE-Track benchmark.
}
\label{tab:core_skill_taxonomy}
\begin{tabular}{
@{}>{\raggedright\arraybackslash}p{2.50cm}
>{\raggedright\arraybackslash}p{3.20cm}
>{\raggedright\arraybackslash}p{3.20cm}
>{\raggedright\arraybackslash}p{1.90cm}
>{\raggedright\arraybackslash}p{1.50cm}@{}}
\toprule
\textbf{Functional skill family}
& \textbf{Input artifact(s)}
& \textbf{Output artifact(s)}
& \textbf{Primary backend}
& \textbf{Used by} \\
\midrule

\multicolumn{5}{@{}l}{\textit{Skill-Orchestrated Artifact Synthesis}} \\
\midrule

Intent Blueprinting
& User Intent; Resource Descriptor
& Intent Blueprint
& LLM
& All \\

Acquire Raw Sources
& Intent Blueprint; External Resource
& Source Record
& Resource adapter
& SU \\

Adapt Existing Benchmarks
& Intent Blueprint; Existing Benchmark Record
& Adapted Source Record
& Resource adapter
& SHT, SQ \\

Collect in Simulation
& Intent Blueprint; Simulator Scenario
& Source Record; State Record
& Simulator API
& SHN, SD, SA, IE \\

Evidence \& State Structuring
& Source Record; State Record
& Evidence Record; Structured State Record
& Hybrid
& All \\

Template Design
& Intent Blueprint; Evidence/State Record
& Expert Template
& LLM
& All \\

Code Generation
& Expert Template; Artifact Schema
& Executable Synthesis Code
& LLM
& All \\

Candidate Item Synthesis
& Expert Template; Evidence/State Record; Synthesis Code
& OE Item or IE Item
& Hybrid
& All \\

Metric Production
& Candidate Item; Scoring or Goal Requirements
& Scoring Artifact or Goal-Verifier Artifact
& Scoring / verifier compiler
& All \\

Evaluation Reporting
& Verified Benchmark Package; Evaluation Record
& Evaluation Report
& Hybrid
& All \\

\midrule
\multicolumn{5}{@{}l}{\textit{Requirement-Guided Verification and Repair}} \\
\midrule

Requirement-Contract Checking
& Typed Artifact; Requirement Contract
& Verification Record
& Hybrid
& All \\

Quality-Gate Execution
& Candidate Item; Verification Record
& Gate Decision
& Hybrid
& All \\

Failure Diagnosis
& Failed Artifact; Verification Record
& Violated Requirements; Affected Dependencies
& Hybrid
& All \\

Local Repair / Rollback
& Failed Artifact; Diagnosis Record; Provenance Metadata
& Repaired Artifact or Re-execution Request
& Hybrid
& All \\

\midrule
\multicolumn{5}{@{}l}{\textit{Track-Specific Validation and Diagnosis}} \\
\midrule

Grey-Batch Synthesis
& Candidate Item Stream
& Small Candidate Batch
& Deterministic program
& All \\

Invalid Item Screening
& OE Item
& Screening Record
& Hybrid
& OE-Track \\

Image-Dependency Blind Evaluation
& OE Item; Ablated Visual Input
& Target-Dependency Record
& Hybrid
& OE-Track \\

Task Validity Check
& IE Item; Structured State Record
& Task-Validity Record
& Hybrid
& IE-Track \\

Task Completion Test
& IE Item; Simulator State; Terminal Verifier
& Execution Record
& Simulator API
& IE-Track \\

Initial-State Check
& Initial-State Specification; Simulator
& State-Restoration Record
& Simulator API
& IE-Track \\

Typical Model Evaluation
& Candidate Batch; Evaluation Models
& Item--Model Response Matrix
& Hybrid
& All \\

IRT Diagnosis
& Item--Model Response Matrix
& Difficulty and Discrimination Estimates
& Deterministic program
& All \\

\bottomrule
\end{tabular}
\end{table*}

\subsection{Skill-Reuse Counting Protocol}
\label{app:skill_reuse_protocol}

Table~\ref{tab:app_skill_reuse_details} expands the skill-reuse
statistics reported in the main paper.
The analyzed implementation snapshot contains 64 globally distinct
skill/card definitions.
For each benchmark-construction run, \textit{Skills Used} counts the
distinct definitions invoked by the workflow, with repeated retries,
re-executions, and repair-time invocations counted only once.
\textit{Reused Skills} are definitions shared with at least one other
benchmark setting, while \textit{Specific Skills/Adapters} are used only
in the current setting. Thus,
\[
\textit{Skills Used}
=
\textit{Reused Skills}
+
\textit{Specific Skills/Adapters},
\qquad
\mathrm{ReuseRate}
=
\frac{\textit{Reused Skills}}
     {\textit{Skills Used}}
\times 100\%.
\]

Rows that combine multiple benchmarks report the average numbers of used,
reused, and specific skills per construction run, while their reuse rates
are computed from pooled counts.
For example, the four simulator-based OE benchmarks contain 161
benchmark--skill incidences, including 157 reused incidences, giving
\(161/4=40.3\) skills used, \(157/4=39.3\) reused skills, and a pooled
reuse rate of \(157/161=97.5\%\).
Across all seven construction runs, the corresponding totals are 287,
263, and 24, yielding the reported averages of \(41.0\), \(37.6\), and
\(3.4\), and an overall reuse rate of \(263/287=91.6\%\).
Here, 287 is a benchmark--skill incidence count rather than the number of
globally unique definitions, since the same skill may be used by multiple
benchmarks.

\newpage
\onecolumn


\clearpage
\section{Construction Cost and Efficiency Analysis}
\label{app:construction_cost_analysis}

This appendix reports the construction-side computational cost of
Embodied-BenchForge.
It first defines the token, wall-clock, and throughput accounting protocol,
and then presents the normalized benchmark-level cost together with the
stage-wise breakdown across S1--S5.
The analysis isolates benchmark-construction overhead from target-model
inference and four-judge artifact-quality assessment.

\subsection{Token and Time Accounting}
\label{app:token_time_accounting}

For cost accounting, the construction workflow is grouped into five
stages consistent with the forward path in the main paper:
\textbf{S1} denotes \textit{Intent Blueprinting};
\textbf{S2}, \textit{Data Collection};
\textbf{S3}, \textit{Evidence \& State Structuring};
\textbf{S4}, \textit{Benchmark Synthesis}; and
\textbf{S5}, \textit{Verification \& Evaluation Reporting}.
Verification operations invoked during S1--S4 are charged to the stage
that produces the corresponding artifact, while S5 includes grey-batch
evaluation, IRT analysis, result aggregation, and report generation.

The reported token cost includes all construction-side model calls,
including initial generation, automatic retries, requirement checking,
repair or re-execution, and IRT-related processing.
It excludes the four-judge artifact-quality assessment and the inference
cost of target models answering benchmark items or executing interactive
tasks.
These costs are reported separately because they depend on the selected
judge and evaluation-model sets rather than on benchmark construction
itself.

Wall-clock time starts when the user evaluation intent is submitted and
covers artifact construction, script-based validation, retries, repair,
and final result aggregation and report generation.
Time spent running target models on the completed benchmark is excluded,
although the subsequent aggregation of their evaluation results is
included.
For a benchmark containing \(N\) accepted items and requiring
\(T\) construction minutes, throughput is computed as
\[
\mathrm{Throughput}=\frac{N}{T}
\quad\text{items per minute}.
\]
The pooled throughput across multiple benchmarks uses the total number
of accepted items divided by their total construction time.

For the IE-Track, the reported average of approximately 1.5 hours per
20-task batch includes task synthesis, state initialization, verifier
construction, and script-based execution checks used to confirm task
feasibility.
It does not include the time required for VLM-based agents to execute
and answer the resulting tasks.

\subsection{Construction Cost and Stage-Wise Breakdown}
\label{app:construction_cost_details}

Table~\ref{tab:construction_cost} summarizes the normalized construction
cost and throughput of the six Offline EQA benchmarks, while
Table~\ref{tab:app_full_stage_cost} provides the corresponding stage-wise
token and time breakdown.
Both tables follow the accounting protocol in
Section~\ref{app:token_time_accounting}; target-model evaluation and
four-judge quality assessment are excluded.

\begin{table}[!ht]
\centering
\small
\setlength{\tabcolsep}{3.2pt}
\caption{
Normalized construction cost of the six
Embodied-BenchForge-produced Offline EQA benchmarks.
All benchmarks are normalized to 10,000 final items.
SHT, SHN, SD, SQ, SU, and SA denote SpatialHome-Task,
SpatialHome-Nav, SpatialDrive, SpatialQuadruped, SpatialUAV,
and SpatialArm, respectively.
Token cost denotes the agent-orchestration cost across the complete
construction workflow and excludes target-model evaluation.
The final-row token and time values are arithmetic means across the six
construction runs, while throughput is computed from the pooled total of
60,000 final items and 516 construction minutes.
}
\label{tab:construction_cost}
\begin{tabular}{@{}lccccc@{}}
\toprule
\textbf{Benchmark}
& \textbf{Scale}
& \textbf{\makecell{Total\\Tokens}}
& \textbf{Time}
& \textbf{\makecell{Tokens/\\Item}}
& \textbf{\makecell{Items/\\Min.}} \\
\midrule
SHT & 10K & 8.90M  & 67.0 min  & 890   & 149.3 \\
SHN & 10K & 13.70M & 94.0 min  & 1,370 & 106.4 \\
SD  & 10K & 9.60M  & 38.0 min  & 960   & 263.2 \\
SQ  & 10K & 10.60M & 78.0 min  & 1,060 & 128.2 \\
SU  & 10K & 14.62M & 160.0 min & 1,462 & 62.5  \\
SA  & 10K & 11.30M & 79.0 min  & 1,130 & 126.6 \\
\midrule
\textbf{Mean / pooled}
& \textbf{10K}
& \textbf{11.45M}
& \textbf{86.0 min}
& \textbf{1,145}
& \textbf{116.3} \\
\bottomrule
\end{tabular}
\end{table}

\begin{table*}[!ht]
\centering
\footnotesize
\setlength{\tabcolsep}{3.0pt}
\renewcommand{\arraystretch}{1.05}
\caption{Stage-wise construction cost of the six
Embodied-BenchForge-produced Offline EQA benchmarks.
All results are normalized to 10,000 final items.
Input, output, and total costs are reported in million tokens.
Stage totals are consistent with Table~\ref{tab:construction_cost}.}
\label{tab:app_full_stage_cost}
\resizebox{\textwidth}{!}{
\begin{tabular}{llcccc}
\toprule
\textbf{Benchmark}
& \textbf{Construction Stage}
& \textbf{Input / M}
& \textbf{Output / M}
& \textbf{Total / M}
& \textbf{Time / min} \\
\midrule

SHT
& S1 Intent Blueprinting
& 0.98 & 0.07 & 1.05 & 7.0 \\
& S2 Data Collection
& 1.08 & 0.07 & 1.15 & 14.0 \\
& S3 Evidence \& State Structuring
& 1.42 & 0.08 & 1.50 & 12.0 \\
& S4 Benchmark Synthesis
& 4.38 & 0.32 & 4.70 & 29.0 \\
& S5 Verification \& Evaluation Reporting
& 0.46 & 0.04 & 0.50 & 5.0 \\
\midrule

SHN
& S1 Intent Blueprinting
& 1.40 & 0.10 & 1.50 & 8.0 \\
& S2 Data Collection
& 1.70 & 0.10 & 1.80 & 25.0 \\
& S3 Evidence \& State Structuring
& 2.37 & 0.13 & 2.50 & 18.0 \\
& S4 Benchmark Synthesis
& 6.82 & 0.48 & 7.30 & 37.0 \\
& S5 Verification \& Evaluation Reporting
& 0.55 & 0.05 & 0.60 & 6.0 \\
\midrule

SD
& S1 Intent Blueprinting
& 1.12 & 0.08 & 1.20 & 5.0 \\
& S2 Data Collection
& 0.84 & 0.06 & 0.90 & 8.0 \\
& S3 Evidence \& State Structuring
& 1.24 & 0.06 & 1.30 & 5.0 \\
& S4 Benchmark Synthesis
& 5.34 & 0.36 & 5.70 & 16.0 \\
& S5 Verification \& Evaluation Reporting
& 0.46 & 0.04 & 0.50 & 4.0 \\
\midrule

SQ
& S1 Intent Blueprinting
& 1.16 & 0.09 & 1.25 & 8.0 \\
& S2 Data Collection
& 0.60 & 0.05 & 0.65 & 8.0 \\
& S3 Evidence \& State Structuring
& 1.47 & 0.08 & 1.55 & 17.0 \\
& S4 Benchmark Synthesis
& 6.17 & 0.43 & 6.60 & 40.0 \\
& S5 Verification \& Evaluation Reporting
& 0.51 & 0.04 & 0.55 & 5.0 \\
\midrule

SU
& S1 Intent Blueprinting
& 1.53 & 0.12 & 1.65 & 9.0 \\
& S2 Data Collection
& 1.40 & 0.10 & 1.50 & 18.0 \\
& S3 Evidence \& State Structuring
& 2.83 & 0.17 & 3.00 & 65.0 \\
& S4 Benchmark Synthesis
& 7.30 & 0.55 & 7.85 & 60.0 \\
& S5 Verification \& Evaluation Reporting
& 0.57 & 0.05 & 0.62 & 8.0 \\
\midrule

SA
& S1 Intent Blueprinting
& 1.25 & 0.10 & 1.35 & 7.0 \\
& S2 Data Collection
& 0.98 & 0.07 & 1.05 & 13.0 \\
& S3 Evidence \& State Structuring
& 1.52 & 0.08 & 1.60 & 15.0 \\
& S4 Benchmark Synthesis
& 6.30 & 0.45 & 6.75 & 39.0 \\
& S5 Verification \& Evaluation Reporting
& 0.51 & 0.04 & 0.55 & 5.0 \\
\bottomrule
\end{tabular}
}
\end{table*}

\paragraph{Construction-Cost Analysis.}
Tables~\ref{tab:construction_cost} and
\ref{tab:app_full_stage_cost} report the normalized cost of constructing
10,000 final items for each Offline EQA benchmark.
Across the six construction runs, Embodied-BenchForge consumes an average
of 11.45M tokens, corresponding to approximately 1,145 tokens per final
item.
The mean construction time is 86.0 minutes, and the pooled throughput is
116.3 items per minute over 60,000 final items and 516 total construction
minutes.
The stage-wise breakdown shows a consistent cost structure across
heterogeneous embodied resources.
S1 Intent Blueprinting accounts for 10.9\%--12.5\% of the total token
cost, reflecting the relatively stable workflow for evaluation-intent
interpretation, capability planning, and template organization.
S2 Data Collection varies more substantially, from 6.1\% to 13.1\%,
because different benchmarks require simulator interaction, existing
trajectory loading, or real-image ingestion.
S3 Evidence \& State Structuring contributes 13.5\%--20.5\%.
Its largest share occurs for SpatialUAV, where unannotated real-world
aerial images require additional cleaning, annotation, and
answer-bearing evidence organization.
By contrast, simulator- and dataset-derived resources expose more
structured state or trajectory information and therefore require less
processing at this stage.
S4 Benchmark Synthesis is the dominant cost source for every benchmark,
accounting for 52.8\%--62.3\% of the total.
This stage executes the deepest skill hierarchy and includes
template--evidence binding, candidate-item generation, code and scoring
artifact production, grey-batch validation, repair-time re-execution, and
full-scale synthesis.
S5 Verification \& Evaluation Reporting contributes only
4.2\%--5.6\%, because the reported construction cost includes workflow
verification, diagnostic processing, result aggregation, and report
generation but excludes the large-scale inference cost of target models
answering the completed benchmark.
Input-token cost is consistently 13.8--15.0 times the output-token cost,
with a pooled ratio of approximately 14.4.
This asymmetry is expected because construction agents repeatedly consume
skill definitions, upstream artifacts, evidence records, execution
feedback, and repair context, while producing comparatively concise
structured artifacts and summaries.
It also indicates that the principal cost arises from grounded context
processing and artifact coordination rather than unrestricted
long-form generation.
The benchmark-level variation further reflects differences in resource
interfaces rather than benchmark scale, since all results are normalized
to the same 10,000-item output.
SpatialHome-Task has the lowest token cost at 8.90M, whereas SpatialUAV
has the highest cost at 14.62M.
SpatialDrive is the fastest construction run, requiring 38 minutes and
achieving 263.2 items per minute, because its structured driving records
can be converted into evidence and benchmark items with relatively low
collection and structuring overhead.
SpatialUAV requires 160 minutes and achieves 62.5 items per minute,
primarily because its S3 evidence-structuring stage alone takes 65
minutes.
These results show that Embodied-BenchForge maintains a common
construction backbone across resources, while the remaining cost
variation is concentrated in resource-specific collection, evidence
organization, and synthesis operations.

\newpage
\onecolumn

\section{Experimental Details and Extended Results}
\label{app:experimental_details}

This appendix collects the implementation and extended evaluation details
supporting the main experiments.
It reports the construction and evaluation models, runtime environment,
human and model-judge protocols, Offline EQA benchmark composition,
LLM-as-Judge quality assessment, per-capability model performance, and the
complete IRT diagnostic and stability analysis.
Construction-side token and time costs are reported separately in
Appendix~\ref{app:construction_cost_analysis}.

\subsection{Implementation and Reproducibility}

\subsubsection{Construction Models and Runtime Settings}
\label{app:models_runtime}

Embodied-BenchForge assigns workflow coordination, artifact synthesis,
process verification, and artifact-quality assessment to distinct model
roles.
Qwen3.6-27B coordinates skill selection, tool invocation, and workflow
execution, while GPT-5.5 Pro supports artifact synthesis and
model-assisted verification.
Artifact quality is assessed by four judges from the OpenAI, Anthropic,
Google, and open-weight Qwen model families.
Table~\ref{tab:construction-models} reports the exact model identifiers,
reasoning settings, sampling parameters, and output-token limits used for
benchmark construction and quality assessment.
Table~\ref{tab:evaluation-models} reports the corresponding identifiers
and inference configurations of the 11 models evaluated in the main
experiments.
Within each benchmark track, all models receive the same prompts, visual
inputs, tool interfaces, stopping rules, and output schemas.
Fields marked as \textit{omitted} were not included in the corresponding
API request, whereas explicit sampling parameters were specified for
locally served models.
Where supported, reasoning-effort levels were left at their
model-specific API defaults.
For locally served Qwen models, which do not expose a discrete effort
parameter, we instead report the checkpoint-specific default thinking
mode.
All evaluation requests use a maximum output-token cap of 32,768.

Experiments were conducted on a server with four NVIDIA H100 80\,GB GPUs,
two 64-core CPU sockets, and 512\,GB of host memory; open-weight models
were served through an OpenAI-compatible vLLM endpoint with tensor
parallelism set to 4.
The software environment used Ubuntu 22.04.5 LTS, Python 3.11.9,
CUDA 12.8, PyTorch 2.8.0, Transformers 5.1.0, and vLLM 0.19.0.

\begin{table*}[!ht]
\centering
\footnotesize
\setlength{\tabcolsep}{4.0pt}
\renewcommand{\arraystretch}{1.12}
\caption{
Models and inference configurations used for benchmark construction and
artifact-quality assessment.
Exact identifiers pin the corresponding API snapshots or open-weight
checkpoints.
\textit{Max output} denotes the request-side output-token cap rather than
the realized response length.
}
\label{tab:construction-models}
\begin{tabularx}{\textwidth}{
@{}p{2.65cm}p{5.05cm}X>{\centering\arraybackslash}p{1.55cm}@{}}
\toprule
\textbf{Role}
& \textbf{Model and exact identifier}
& \textbf{Inference configuration}
& \textbf{\makecell{Max\\output}} \\
\midrule

Workflow coordination
&
Qwen3.6-27B
(\path{Qwen/Qwen3.6-27B};
HF release 2026-04-22)
&
Default thinking mode: enabled;
no discrete effort level;
automatic tool selection;
\path{qwen3_coder} tool-call parser;
$T{=}1.0$, top-$p{=}0.95$, top-$k{=}20$,
$\mathrm{min}\_p{=}0$
&
32,768 \\

Artifact synthesis and verification
&
GPT-5.5 Pro
(\path{gpt-5.5-pro-2026-04-23})
&
Reasoning effort: high (API default);
temperature and top-$p$ omitted
&
32,768 \\

Judge 1
&
GPT-5.5
(\path{gpt-5.5-2026-04-23})
&
Reasoning effort: medium (API default);
temperature and top-$p$ omitted
&
16,384 \\

Judge 2
&
Claude Opus 4.7
(\path{claude-opus-4-7})
&
Adaptive thinking enabled;
effort: high (API default);
temperature, top-$p$, and top-$k$ omitted
&
16,384 \\

Judge 3
&
Gemini 3 Flash
(\path{gemini-3-flash-preview})
&
Thinking level: high (API default);
$T{=}1.0$, top-$p{=}0.95$
&
16,384 \\

Judge 4
&
Qwen3.6-35B-A3B
(\path{Qwen/Qwen3.6-35B-A3B};
HF release 2026-04-14)
&
Default thinking mode: enabled;
no discrete effort level;
$T{=}1.0$, top-$p{=}0.95$, top-$k{=}20$,
$\mathrm{min}\_p{=}0$
&
16,384 \\

\bottomrule
\end{tabularx}
\end{table*}

\begin{table*}[!ht]
\centering
\footnotesize
\setlength{\tabcolsep}{4.0pt}
\renewcommand{\arraystretch}{1.12}
\caption{
Exact identifiers and request-side inference configurations of the
11 models evaluated in the main experiments.
All models use a maximum output-token cap of 32,768.
\textit{Omitted} indicates that the corresponding sampling field was not
included in the request.
}
\label{tab:evaluation-models}
\begin{tabularx}{\textwidth}{
@{}p{2.35cm}p{4.85cm}p{3.05cm}X@{}}
\toprule
\textbf{Reported name}
& \textbf{Exact identifier}
& \textbf{Reasoning configuration}
& \textbf{Sampling controls} \\
\midrule

\multicolumn{4}{@{}l}{\textit{Closed-source / API models}} \\
\midrule

GPT-5.5
&
\path{gpt-5.5-2026-04-23}
&
Effort: medium (API default)
&
Temperature and top-$p$ omitted \\

Claude Opus 4.7
&
\path{claude-opus-4-7}
&
Adaptive thinking enabled;
effort: high (API default)
&
Temperature, top-$p$, and top-$k$ omitted \\

Gemini-3-Flash
&
\path{gemini-3-flash-preview}
&
Thinking level: high (API default)
&
$T{=}1.0$, top-$p{=}0.95$ \\

GPT-5.4-Mini
&
\path{gpt-5.4-mini-2026-03-17}
&
Effort: none (API default)
&
Temperature and top-$p$ omitted \\

GPT-5.4-Nano
&
\path{gpt-5.4-nano-2026-03-17}
&
Effort: none (API default)
&
Temperature and top-$p$ omitted \\

Claude Sonnet 4.6
&
\path{claude-sonnet-4-6}
&
Adaptive thinking enabled;
effort: high (API default)
&
Temperature and top-$p$ omitted \\

Claude 4.5 Haiku
&
\path{claude-haiku-4-5-20251001}
&
Extended thinking disabled (API default);
effort not supported
&
Temperature and top-$p$ omitted \\

\midrule
\multicolumn{4}{@{}l}{\textit{Open-weight / local models}} \\
\midrule

Qwen3.6-35B
&
\path{Qwen/Qwen3.6-35B-A3B}
&
Default thinking mode: enabled;
no discrete effort level
&
$T{=}1.0$, top-$p{=}0.95$, top-$k{=}20$,
$\mathrm{min}\_p{=}0$; presence penalty $=1.5$ \\

Qwen3.5-4B
&
\path{Qwen/Qwen3.5-4B}
&
Default thinking mode: enabled;
no discrete effort level
&
$T{=}0.6$, top-$p{=}0.95$, top-$k{=}20$;
presence penalty $=0$ \\

Qwen3.5-2B
&
\path{Qwen/Qwen3.5-2B}
&
Default thinking mode: disabled;
no discrete effort level
&
$T{=}0.6$, top-$p{=}0.95$, top-$k{=}20$;
presence penalty $=0$ \\

Qwen3.5-0.8B
&
\path{Qwen/Qwen3.5-0.8B}
&
Default thinking mode: disabled;
no discrete effort level
&
$T{=}0.6$, top-$p{=}0.95$, top-$k{=}20$;
presence penalty $=0$ \\

\bottomrule
\end{tabularx}
\end{table*}

\subsection{Human and Judge Evaluation Details}
\label{appendix:human_judge_eval}

We conduct human evaluation to provide an independent reference for
artifact-quality assessment.
The study involves \(10\) participants with experience in embodied
intelligence research or benchmark development, including three university
faculty members, four master's students, and three doctoral students.
Samples are stratified across the six OE-Track benchmarks and the eight
quality dimensions: UIA, FSQ, QAC, EGC, SGC, AFC, TSD, and SSC.
Each sampled item is independently assessed by at least three participants
using the same scoring rubric as the model judges.
Participants are blinded to the construction setting and ablation variant
and receive written instructions and representative examples before formal
annotation.
Final human scores are obtained by averaging repeated annotations, and
inter-annotator reliability is measured using Krippendorff's
\(\alpha\).

The four model judges evaluate the same sampled artifacts using a shared
prompt, input fields, scoring scale, and output schema.
Their identities and inference configurations are reported in
Table~\ref{tab:construction-models}.
Judge scores are averaged across models, and Judge--Human agreement is
computed as the sample-level Spearman correlation between the aggregated
judge and human scores.

\subsection{Offline EQA Benchmark Composition}
\label{app:offline_composition}

Table~\ref{tab:appendix_capability_qtype_composition} details the
capability and question-type composition of the six Offline EQA
benchmarks.
Each benchmark contains 10,000 items.
Capability identifiers are local to the corresponding benchmark, and the
capability and question-type columns are independent distributions rather
than row-wise mappings.

{\footnotesize
\setlength{\tabcolsep}{3.5pt}
\renewcommand{\arraystretch}{1.05}
\begin{longtable}{
@{}>{\raggedright\arraybackslash}p{3.10cm}
>{\raggedright\arraybackslash}p{4.80cm}
>{\raggedright\arraybackslash}p{5.30cm}@{}}
\caption{
Detailed capability and question-type composition of the six Offline EQA
benchmarks constructed by Embodied-BenchForge.
Each benchmark contains 10,000 items.
Capability dimensions are ordered by their local identifiers, while
question types are ordered by decreasing frequency.
The capability and question-type lists independently sum to 10,000 and
do not indicate a one-to-one correspondence.
}
\label{tab:appendix_capability_qtype_composition}\\
\toprule
\textbf{Benchmark (Source)}
& \textbf{Capability Dimensions}
& \textbf{Question Types} \\
\midrule
\endfirsthead
\multicolumn{3}{c}{\tablename~\thetable\ (continued)} \\
\toprule
\textbf{Benchmark (Source)}
& \textbf{Capability Dimensions}
& \textbf{Question Types} \\
\midrule
\endhead
\midrule
\multicolumn{3}{r}{Continued on next page} \\
\endfoot
\bottomrule
\endlastfoot

\textbf{SpatialHome-Task}\\
(ALFRED)
&
\texttt{C01} Object visibility (997)\newline
\texttt{C02} Counting and set size (1,040)\newline
\texttt{C03} Image-plane spatial relations (613)\newline
\texttt{C04} Distance and depth (1,953)\newline
\texttt{C05} Receptacle and containment (1,825)\newline
\texttt{C06} Reachability and egocentric reasoning (604)\newline
\texttt{C07} Temporal visibility and trajectory (1,319)\newline
\texttt{C08} Object state and affordance (1,649)
&
Marked-object choice (3,072)\newline
Multiple choice (2,719)\newline
Ordered list (1,898)\newline
Multi-view choice (974)\newline
Interval choice (847)\newline
Single choice (490)
\\
\midrule

\textbf{SpatialHome-Nav}\\
(Habitat)
&
\texttt{C01} Single-frame region/depth localization (1,497)\newline
\texttt{C02} Near-space coverage (1,456)\newline
\texttt{C03} Depth variation and openness (1,104)\newline
\texttt{C05} Camera-plane trajectory (1,786)\newline
\texttt{C06} Camera-orientation change (1,494)\newline
\texttt{C07} Image-plane region localization (1,338)\newline
\texttt{C08} Acquisition order and multi-view consistency (1,325)
&
Single choice (2,655)\newline
Ordered list (2,636)\newline
Grounding choice (2,426)\newline
Multi-view choice (2,283)
\\
\midrule

\textbf{SpatialDrive}\\
(CARLA)
&
\texttt{O1} Visible semantic categories (1,666)\newline
\texttt{O2} Visible-count intervals (1,630)\newline
\texttt{O3} Image-plane spatial relations (140)\newline
\texttt{O4} Visible-area scale (140)\newline
\texttt{M1} Cross-view visibility (1,195)\newline
\texttt{M2} Best-view selection (1,290)\newline
\texttt{R1} Traffic-infrastructure semantics (3,840)\newline
\texttt{T1} Temporal visibility dynamics (99)
&
Multiple choice (5,035)\newline
Marked-object choice (1,666)\newline
Interval choice (1,630)\newline
Multi-view choice (1,290)\newline
Ordered list (280)\newline
Temporal choice (99)
\\
\midrule

\textbf{SpatialArm}\\
(LIBERO)
&
\texttt{LC1} Object inventory (844)\newline
\texttt{LC2} Counting and set reasoning (1,299)\newline
\texttt{LC3} 3D spatial relations (839)\newline
\texttt{LC4} Distance and proximity (840)\newline
\texttt{LC5} Order and region (1,524)\newline
\texttt{LC6} Robot end-effector actions (847)\newline
\texttt{LC7} Temporal trajectory (1,645)\newline
\texttt{LC8} Task-semantic binding (1,295)\newline
\texttt{LC9} Multi-view consistency (867)
&
Multiple choice (1,971)\newline
Ordered list (1,654)\newline
Marked-object choice (1,609)\newline
Single choice (1,031)\newline
Interval choice (1,024)\newline
Yes/no (1,000)\newline
Multi-view choice (867)\newline
Temporal choice (844)
\\
\midrule

\textbf{SpatialUAV}\\
(Real aerial images)
&
\texttt{C01} Visible-category recognition (1,065)\newline
\texttt{C02} Instance counting and category distribution (985)\newline
\texttt{C03} Image-plane location and region localization (1,265)\newline
\texttt{C04} Visible-area and scale reasoning (1,092)\newline
\texttt{C05} Depth and near--far reasoning (1,115)\newline
\texttt{C06} Camera-frame 3D position and visible 3D scale (1,074)\newline
\texttt{C07} Box geometry and image-boundary relations (1,088)\newline
\texttt{C08} Category-constrained object selection and matching (1,257)\newline
\texttt{C11} Image-level aspect geometry (1,059)
&
Single choice (5,429)\newline
Ordered list (2,394)\newline
Interval choice (888)\newline
Matching (606)\newline
Multiple choice (571)\newline
Grounding choice (56)\newline
Marked-object choice (56)
\\
\midrule

\textbf{SpatialQuadruped}\\
(TartanGround)
&
\texttt{C01} Semantic-category visibility (1,343)\newline
\texttt{C02} Semantic-region area comparison (1,051)\newline
\texttt{C03} Image-region semantic localization (1,344)\newline
\texttt{C04} Region depth and near--far ordering (1,344)\newline
\texttt{C05} Joint semantic-category and depth reasoning (1,200)\newline
\texttt{C06} Multi-view semantic consistency (1,273)\newline
\texttt{C07} Trajectory order and path structure (1,201)\newline
\texttt{C08} Motion trend and orientation change (1,244)
&
Ordered list (1,527)\newline
Single choice (1,526)\newline
Interval choice (1,512)\newline
Temporal choice (1,331)\newline
Matching (1,277)\newline
Grounding choice (1,231)\newline
Multiple choice (800)\newline
Multi-view choice (796)
\\

\end{longtable}
}

\subsection{LLM-as-Judge Benchmark Quality Assessment}

We use four judge models to assess the quality of the six
Embodied-BenchForge-produced \textit{OE-Track} benchmarks.
Table~\ref{tab:llm_as_judge_metric_definitions} defines eight assessment
dimensions covering intent alignment, usability, answer consistency,
evidence grounding, spatial consistency, affordance validity,
target-signal dependency, and skill challenge.
Greater weights are assigned to evidence grounding, spatial consistency,
and answer consistency because they directly affect benchmark
answerability and scoring reliability.
As shown in Table~\ref{tab:llm_as_judge_quality}, the average Overall
scores range from 88.34 to 90.75, with SpatialArm achieving the highest
score.
The benchmarks perform consistently well in intent alignment, usability,
evidence grounding, and spatial consistency.
Relatively lower scores on skill challenge and target-signal dependency
suggest room for constructing more discriminative and evidence-dependent
items.
Overall, the results indicate that the generated benchmarks are
well-grounded, answerable, and suitable for automatic evaluation.

\begin{table*}[!ht]
\centering
\small
\setlength{\tabcolsep}{4pt}
\renewcommand{\arraystretch}{1.05}
\caption{
Definitions and weighting protocol of the LLM-as-Judge benchmark quality assessment metrics.
Each benchmark item is evaluated from multiple quality perspectives, and the final Overall score reflects a holistic judgment over these dimensions.
When AFC is not applicable to a sample, its influence is reduced or removed from the final judgment.
}
\label{tab:llm_as_judge_metric_definitions}
\resizebox{\textwidth}{!}{
\begin{tabular}{l l p{7.2cm} c}
\toprule
\textbf{Metric} & \textbf{Full Name} & \textbf{Evaluation Focus} & \textbf{Weight} \\
\midrule
UIA & User-Intent Alignment 
& Whether the sample matches the intended embodied spatial capability and avoids drifting into generic recognition, language commonsense, or irrelevant tasks. 
& 12\% \\

FSQ & Format and Usability Quality 
& Whether the question, options, answer, evidence fields, and scoring format are clear, complete, and usable for automatic or human evaluation. 
& 12\% \\

QAC & Question-Answer Consistency 
& Whether the question, candidate options, reference answer, and explanation are mutually consistent, unambiguous, and free from answer-key mismatches. 
& 14\% \\

EGC & Evidence-Grounding Correctness 
& Whether the reference answer is supported by the available evidence, such as image content, annotations, depth, bbox, mask, trajectory, metadata, or scene descriptions. 
& 18\% \\

SGC & Spatial-Geometric Consistency 
& Whether spatial relations such as left/right, front/back, near/far, containment, visibility, occlusion, viewpoint, and multi-frame geometry are correct and consistent. 
& 16\% \\

AFC & Action/Affordance Consistency 
& Whether involved actions, interactions, navigation, manipulation, reachability, placement, or physical affordances are feasible and supported by evidence. 
& 10\% \\

TSD & Target-Signal Dependency 
& Whether the sample requires the target visual/spatial evidence rather than being answerable through language priors, commonsense, option bias, or template leakage. 
& 10\% \\

SSC & Skill Challenge 
& Whether the sample provides non-trivial skill demand and can distinguish models with different spatial, embodied, temporal, or structured reasoning abilities. 
& 8\% \\

\midrule
Overall & Overall Quality 
& A holistic benchmark-item quality score considering alignment, answerability, evidence support, spatial validity, signal dependency, challenge, and scoring reliability. 
& 100\% \\
\bottomrule
\end{tabular}
}
\end{table*}

\begin{table*}[p]
\centering
\setlength{\tabcolsep}{3.5pt}
\renewcommand{\arraystretch}{0.92}
\caption{
LLM-as-Judge quality assessment results of six BenchForge-produced benchmarks.
Four judge models, including GPT-5.5, Claude-Opus-4-7, Gemini-3-Flash, and Qwen3.6-35B-A3B, are used to evaluate the overall quality of each benchmark.
UIA, FSQ, QAC, EGC, SGC, AFC, TSD, and SSC denote user-intent alignment, format and usability quality, question-answer consistency, evidence-grounding correctness, spatial-geometric consistency, action/affordance consistency, target-signal dependency, and skill challenge, respectively.
Overall denotes the average quality score across the eight dimensions.
}
\label{tab:llm_as_judge_quality}
\resizebox{\textwidth}{!}{
\begin{tabular}{llccccccccc}
\toprule
\textbf{Benchmark} & \textbf{Judge Model} & \textbf{UIA} & \textbf{FSQ} & \textbf{QAC} & \textbf{EGC} & \textbf{SGC} & \textbf{AFC} & \textbf{TSD} & \textbf{SSC} & \textbf{Overall} \\
\midrule

\multirow{5}{*}{SpatialHome-Task}
& \raisebox{-0.2ex}{\includegraphics[height=1em]{logos/openai.pdf}}~GPT-5.5 & 88 & 91 & 87 & 90 & 84 & 88 & 90 & 82 & 87.50 \\
& \raisebox{-0.2ex}{\includegraphics[height=1em]{logos/claude-ai-icon.pdf}}~Claude-Opus-4-7 & 86 & 88 & 85 & 88 & 85 & 85 & 84 & 84 & 85.63 \\
& \raisebox{-0.2ex}{\includegraphics[height=1em]{logos/gemini-color.pdf}}~Gemini-3-Flash & 95 & 98 & 96 & 95 & 88 & 92 & 85 & 82 & 91.38 \\
& \raisebox{-0.2ex}{\includegraphics[height=1em]{logos/qwen-color.pdf}}~Qwen3.6-35B-A3B & 92 & 95 & 90 & 94 & 88 & 89 & 85 & 85 & 89.75 \\
& Avg. & 90.25 & 93.00 & 89.50 & 91.75 & 86.25 & 88.50 & 86.00 & 83.25 & 88.56 \\
\midrule

\multirow{5}{*}{SpatialHome-NAV}
& \raisebox{-0.2ex}{\includegraphics[height=1em]{logos/openai.pdf}}~GPT-5.5 & 90 & 87 & 86 & 90 & 88 & 86 & 89 & 88 & 88.00 \\
& \raisebox{-0.2ex}{\includegraphics[height=1em]{logos/claude-ai-icon.pdf}}~Claude-Opus-4-7 & 88 & 92 & 85 & 88 & 87 & 92 & 85 & 88 & 88.13 \\
& \raisebox{-0.2ex}{\includegraphics[height=1em]{logos/gemini-color.pdf}}~Gemini-3-Flash & 92 & 85 & 95 & 96 & 94 & 90 & 92 & 84 & 91.00 \\
& \raisebox{-0.2ex}{\includegraphics[height=1em]{logos/qwen-color.pdf}}~Qwen3.6-35B-A3B & 89 & 85 & 90 & 92 & 88 & 95 & 85 & 90 & 89.25 \\
& Avg. & 89.75 & 87.25 & 89.00 & 91.50 & 89.25 & 90.75 & 87.75 & 87.50 & 89.09 \\
\midrule

\multirow{5}{*}{SpatialArm}
& \raisebox{-0.2ex}{\includegraphics[height=1em]{logos/openai.pdf}}~GPT-5.5 & 92 & 88 & 86 & 90 & 92 & 90 & 90 & 88 & 89.50 \\
& \raisebox{-0.2ex}{\includegraphics[height=1em]{logos/claude-ai-icon.pdf}}~Claude-Opus-4-7 & 86 & 88 & 90 & 88 & 94 & 92 & 92 & 90 & 90.00 \\
& \raisebox{-0.2ex}{\includegraphics[height=1em]{logos/gemini-color.pdf}}~Gemini-3-Flash & 95 & 92 & 94 & 90 & 96 & 85 & 92 & 94 & 92.25 \\
& \raisebox{-0.2ex}{\includegraphics[height=1em]{logos/qwen-color.pdf}}~Qwen3.6-35B-A3B & 95 & 92 & 95 & 90 & 95 & 85 & 90 & 88 & 91.25 \\
& Avg. & 92.00 & 90.00 & 91.25 & 89.50 & 94.25 & 88.00 & 91.00 & 90.00 & 90.75 \\
\midrule

\multirow{5}{*}{SpatialUAV}
& \raisebox{-0.2ex}{\includegraphics[height=1em]{logos/openai.pdf}}~GPT-5.5 & 86 & 92 & 87 & 91 & 86 & 90 & 88 & 92 & 89.00 \\
& \raisebox{-0.2ex}{\includegraphics[height=1em]{logos/claude-ai-icon.pdf}}~Claude-Opus-4-7 & 86 & 92 & 85 & 90 & 87 & 90 & 88 & 90 & 88.50 \\
& \raisebox{-0.2ex}{\includegraphics[height=1em]{logos/gemini-color.pdf}}~Gemini-3-Flash & 85 & 93 & 88 & 91 & 86 & 87 & 91 & 91 & 89.00 \\
& \raisebox{-0.2ex}{\includegraphics[height=1em]{logos/qwen-color.pdf}}~Qwen3.6-35B-A3B & 85 & 92 & 86 & 90 & 88 & 89 & 85 & 92 & 88.38 \\
& Avg. & 85.50 & 92.25 & 86.50 & 90.50 & 86.75 & 89.00 & 88.00 & 91.25 & 88.72 \\
\midrule

\multirow{5}{*}{SpatialQuadruped}
& \raisebox{-0.2ex}{\includegraphics[height=1em]{logos/openai.pdf}}~GPT-5.5 & 88 & 94 & 86 & 91 & 87 & 91 & 86 & 85 & 88.50 \\
& \raisebox{-0.2ex}{\includegraphics[height=1em]{logos/claude-ai-icon.pdf}}~Claude-Opus-4-7 & 89 & 93 & 85 & 90 & 86 & 86 & 88 & 86 & 87.88 \\
& \raisebox{-0.2ex}{\includegraphics[height=1em]{logos/gemini-color.pdf}}~Gemini-3-Flash & 85 & 95 & 86 & 92 & 85 & 87 & 88 & 86 & 88.00 \\
& \raisebox{-0.2ex}{\includegraphics[height=1em]{logos/qwen-color.pdf}}~Qwen3.6-35B-A3B & 92 & 94 & 86 & 91 & 88 & 90 & 86 & 85 & 89.00 \\
& Avg. & 88.50 & 94.00 & 85.75 & 91.00 & 86.50 & 88.50 & 87.00 & 85.50 & 88.34 \\
\midrule

\multirow{5}{*}{SpatialDrive}
& \raisebox{-0.2ex}{\includegraphics[height=1em]{logos/openai.pdf}}~GPT-5.5 & 88 & 89 & 87 & 94 & 89 & 87 & 92 & 87 & 89.13 \\
& \raisebox{-0.2ex}{\includegraphics[height=1em]{logos/claude-ai-icon.pdf}}~Claude-Opus-4-7 & 88 & 92 & 87 & 94 & 89 & 91 & 92 & 87 & 90.00 \\
& \raisebox{-0.2ex}{\includegraphics[height=1em]{logos/gemini-color.pdf}}~Gemini-3-Flash & 88 & 89 & 87 & 94 & 89 & 92 & 92 & 87 & 89.75 \\
& \raisebox{-0.2ex}{\includegraphics[height=1em]{logos/qwen-color.pdf}}~Qwen3.6-35B-A3B & 85 & 85 & 85 & 85 & 90 & 86 & 90 & 90 & 87.00 \\
& Avg. & 87.25 & 88.75 & 86.50 & 91.75 & 89.25 & 89.00 & 91.50 & 87.75 & 88.97 \\
\bottomrule
\end{tabular}
}
\end{table*}

\newpage
\subsection{Per-Capability Model Performance}
\label{app:per_capability_results}

Tables~\ref{tab:alfred_capability_perf}--\ref{tab:uav_capability_perf}
report the detailed model results for each capability dimension of the
six Offline EQA benchmarks.
The capability codes follow
Table~\ref{tab:appendix_capability_qtype_composition}, and the AVG column
summarizes performance across the capability dimensions within each
benchmark.
These tables complement the aggregate results in the main paper by
showing which spatial, temporal, grounding, and embodied reasoning skills
remain difficult for different model families.

\subsection{IRT Diagnostic Protocol and Stability Analysis}
\label{sec:irt_diagnosis}

\paragraph{Analytical objective and protocol.}
IRT Diagnosis constitutes the final auxiliary item-level analysis in
Figure~2. The preceding deterministic, execution-based, and model-assisted
gates establish artifact validity, whereas IRT quantifies the extent to which
the accepted items discriminate among models with different ability levels.
The protocol comprises four stages: constructing model--item response
matrices, fitting a regularized two-parameter logistic (2PL) model, identifying
highly discriminative or potentially reversed items, and evaluating the
stability of the resulting estimates through model bootstrap and a classical
top--bottom score gap. This separation ensures that IRT provides supplementary
measurement evidence rather than replacing the benchmark-validity checks.

\paragraph{Discrimination criterion and audit rule.}
High-Disc. follows Table~8 exactly:
\begin{equation}
  \mathrm{HighDisc}_i=\mathbf{1}[a_i\geq0.7].
  \label{eq:high_disc}
\end{equation}
Because \(\theta\) has unit variance, \(a_i=0.7\) corresponds to an odds
multiplier of \(\exp(0.7)=2.01\) for a one-standard-deviation ability
increase. Thus, High-Disc. denotes a practically strong separation effect,
not a significance test. The same threshold is fixed before analysis and
applied to every benchmark and ablation variant.

The fitted discrimination values span
[-2.60,2.62] on the unified scale. Negative slopes are subject to
inspection because they indicate lower scores among stronger models. We
therefore combine the 2PL slope with the corrected item--total correlation:
\begin{equation}
  \begin{aligned}
  r_{i,-i}
  &=\operatorname{corr}\!\left(
  y_{\cdot i},\sum_{j\neq i}y_{\cdot j}\right),\\
  \mathrm{ReverseRisk}_i
  &=\mathbf{1}[a_i<0\ \lor\ r_{i,-i}<0].
  \end{aligned}
  \label{eq:reverse_risk}
\end{equation}
ReverseRisk routes an item to answer, scoring, grounding, or ambiguity review;
it is an audit flag and does not alter the benchmark score.

\paragraph{Bootstrap stability and classical discrimination analysis.}
Each bootstrap replicate resamples the same 11 model identities across all six
benchmarks, refits both abilities and item parameters, and records the
High-Disc. proportion. The reported 95\% interval uses the 2.5th and 97.5th
percentiles. As a model-independent robustness analysis, we additionally rank
the 11 models by mean score and compute
\begin{equation}
  D_i
  =\frac{1}{3}\sum_{m\in\mathcal{G}_{\mathrm{top}}}y_{mi}
  -\frac{1}{3}\sum_{m\in\mathcal{G}_{\mathrm{bottom}}}y_{mi},
  \label{eq:classical_disc}
\end{equation}
where the top and bottom groups contain three models each. \(D_i\) is the
direct score gap on item \(i\), while \(\rho_S(a_i,D_i)\) measures agreement
between the 2PL and classical rankings.

\begin{table*}[!ht]
\centering
\small
\caption{Per-capability model performance on the SpatialHome-Task benchmark. Columns C01--C08 denote object visibility, counting and set size, image-plane spatial reasoning, distance/depth reasoning, receptacle containment, egocentric reachability, temporal visibility trajectory, and object-state affordance, respectively. Scores are reported in percentage, and AVG denotes the average score across all capability dimensions. Models are grouped into closed-source/API models and open-source/local models.}
\label{tab:alfred_capability_perf}
\resizebox{\textwidth}{!}{
\begin{tabular}{lccccccccc}
\toprule
\textbf{Model} & \textbf{C01} & \textbf{C02} & \textbf{C03} & \textbf{C04} & \textbf{C05} & \textbf{C06} & \textbf{C07} & \textbf{C08} & \textbf{AVG} \\
\midrule
\multicolumn{10}{l}{\textit{Closed-source / API Models}} \\
\midrule
\raisebox{-0.2ex}{\includegraphics[height=1em]{logos/openai.pdf}}~GPT-5.5 & 77.00 & 57.00 & 93.00 & 22.00 & 32.00 & 25.00 & 80.00 & 45.00 & 53.88 \\
\raisebox{-0.2ex}{\includegraphics[height=1em]{logos/claude-ai-icon.pdf}}~Claude Opus 4.7 & 60.00 & 52.00 & 94.00 & 22.00 & 37.00 & 33.00 & 74.00 & 46.00 & 52.25 \\
\raisebox{-0.2ex}{\includegraphics[height=1em]{logos/gemini-color.pdf}}~Gemini-3-Flash & 61.00 & 50.00 & 90.00 & 22.00 & 28.00 & 46.00 & 64.00 & 43.00 & 50.50 \\
\raisebox{-0.2ex}{\includegraphics[height=1em]{logos/openai.pdf}}~GPT-5.4-Mini & 57.00 & 45.00 & 89.00 & 23.00 & 26.00 & 49.00 & 60.00 & 37.00 & 48.25 \\
\raisebox{-0.2ex}{\includegraphics[height=1em]{logos/openai.pdf}}~GPT-5.4-Nano & 51.00 & 35.00 & 91.00 & 14.00 & 13.00 & 42.00 & 38.00 & 44.00 & 41.00 \\
\raisebox{-0.2ex}{\includegraphics[height=1em]{logos/claude-ai-icon.pdf}}~Claude Sonnet 4.6 & 61.00 & 46.00 & 95.00 & 25.00 & 27.00 & 43.00 & 57.00 & 45.00 & 49.88 \\
\raisebox{-0.2ex}{\includegraphics[height=1em]{logos/claude-ai-icon.pdf}}~Claude 4.5 Haiku & 69.00 & 44.00 & 84.00 & 11.00 & 26.00 & 65.00 & 58.00 & 45.00 & 50.25 \\
\midrule
\multicolumn{10}{l}{\textit{Open-source / Local Models}} \\
\midrule
\raisebox{-0.2ex}{\includegraphics[height=1em]{logos/qwen-color.pdf}}~Qwen3.6-35B-A3B & 66.00 & 52.00 & 93.00 & 19.00 & 19.00 & 59.00 & 71.00 & 43.00 & 52.75 \\
\raisebox{-0.2ex}{\includegraphics[height=1em]{logos/qwen-color.pdf}}~Qwen3.5-4B & 44.00 & 37.00 & 76.00 & 17.00 & 22.00 & 41.00 & 47.00 & 19.00 & 37.88 \\
\raisebox{-0.2ex}{\includegraphics[height=1em]{logos/qwen-color.pdf}}~Qwen3.5-2B & 40.00 & 19.00 & 76.00 & 9.00 & 16.00 & 45.00 & 50.00 & 33.00 & 36.00 \\
\raisebox{-0.2ex}{\includegraphics[height=1em]{logos/qwen-color.pdf}}~Qwen3.5-0.8B & 58.00 & 22.00 & 76.00 & 8.00 & 29.00 & 42.00 & 49.00 & 35.00 & 39.88 \\
\bottomrule
\end{tabular}
}
\end{table*}

\begin{table*}[!ht]
\centering
\small
\caption{Per-capability model performance on the SpatialHome-Nav benchmark. Columns C01, C02, C03, C05, C06, C07, and C08 denote single-frame region depth localization, near-space coverage, depth variation and openness, camera-plane trajectory, camera orientation change, image-plane region localization, and acquisition-order/multi-view consistency, respectively. Scores are reported in percentage, and AVG denotes the average score across all capability dimensions. Models are grouped into closed-source/API models and open-source/local models.}
\label{tab:habitat_capability_perf}
\resizebox{\textwidth}{!}{
\begin{tabular}{lcccccccc}
\toprule
\textbf{Model} & \textbf{C01} & \textbf{C02} & \textbf{C03} & \textbf{C05} & \textbf{C06} & \textbf{C07} & \textbf{C08} & \textbf{AVG} \\
\midrule
\multicolumn{9}{l}{\textit{Closed-source / API Models}} \\
\midrule
\raisebox{-0.2ex}{\includegraphics[height=1em]{logos/openai.pdf}}~GPT-5.5 & 61.00 & 36.00 & 67.00 & 69.00 & 81.00 & 63.00 & 47.00 & 60.57 \\
\raisebox{-0.2ex}{\includegraphics[height=1em]{logos/claude-ai-icon.pdf}}~Claude Opus 4.7 & 55.00 & 30.00 & 64.00 & 63.00 & 81.00 & 59.00 & 51.00 & 57.57 \\
\raisebox{-0.2ex}{\includegraphics[height=1em]{logos/gemini-color.pdf}}~Gemini-3-Flash & 49.00 & 29.00 & 56.00 & 42.00 & 66.00 & 53.00 & 23.00 & 45.43 \\
\raisebox{-0.2ex}{\includegraphics[height=1em]{logos/openai.pdf}}~GPT-5.4-Mini & 42.00 & 25.00 & 58.00 & 33.00 & 44.00 & 54.00 & 23.00 & 40.00 \\
\raisebox{-0.2ex}{\includegraphics[height=1em]{logos/openai.pdf}}~GPT-5.4-Nano & 44.00 & 25.00 & 50.00 & 35.00 & 51.00 & 44.00 & 26.00 & 39.29 \\
\raisebox{-0.2ex}{\includegraphics[height=1em]{logos/claude-ai-icon.pdf}}~Claude Sonnet 4.6 & 52.00 & 28.00 & 45.00 & 40.00 & 37.00 & 65.00 & 19.00 & 40.86 \\
\raisebox{-0.2ex}{\includegraphics[height=1em]{logos/claude-ai-icon.pdf}}~Claude 4.5 Haiku & 32.00 & 23.00 & 40.00 & 31.00 & 26.00 & 59.00 & 17.00 & 32.57 \\
\midrule
\multicolumn{9}{l}{\textit{Open-source / Local Models}} \\
\midrule
\raisebox{-0.2ex}{\includegraphics[height=1em]{logos/qwen-color.pdf}}~Qwen3.6-35B-A3B & 58.00 & 31.00 & 52.00 & 44.00 & 49.00 & 65.00 & 23.00 & 46.00 \\
\raisebox{-0.2ex}{\includegraphics[height=1em]{logos/qwen-color.pdf}}~Qwen3.5-4B & 32.00 & 24.00 & 47.00 & 28.00 & 31.00 & 36.00 & 20.00 & 31.14 \\
\raisebox{-0.2ex}{\includegraphics[height=1em]{logos/qwen-color.pdf}}~Qwen3.5-2B & 33.00 & 18.00 & 36.00 & 22.00 & 35.00 & 29.00 & 22.00 & 27.86 \\
\raisebox{-0.2ex}{\includegraphics[height=1em]{logos/qwen-color.pdf}}~Qwen3.5-0.8B & 30.00 & 22.00 & 44.00 & 33.00 & 28.00 & 44.00 & 6.00 & 29.57 \\
\bottomrule
\end{tabular}
}
\end{table*}

\begin{table*}[!ht]
\centering
\small
\caption{Per-capability model performance on the SpatialDrive benchmark. Columns M1, M2, O1, O2, O3, O4, R1, and T1 denote cross-view visibility, best-view selection, visible semantic category recognition, visible count interval reasoning, image-plane spatial relation, visible area/scale reasoning, traffic infrastructure semantics, and temporal visibility dynamics, respectively. Scores are reported in percentage, and AVG denotes the average score across all capability dimensions. Models are grouped into closed-source/API models and open-source/local models.}
\label{tab:carla_capability_perf}
\resizebox{\textwidth}{!}{
\begin{tabular}{lccccccccc}
\toprule
\textbf{Model} & \textbf{M1} & \textbf{M2} & \textbf{O1} & \textbf{O2} & \textbf{O3} & \textbf{O4} & \textbf{R1} & \textbf{T1} & \textbf{AVG} \\
\midrule
\multicolumn{10}{l}{\textit{Closed-source / API Models}} \\
\midrule
\raisebox{-0.2ex}{\includegraphics[height=1em]{logos/openai.pdf}}~GPT-5.5 & 6.00 & 38.00 & 27.00 & 27.00 & 93.67 & 84.50 & 1.00 & 79.00 & 44.52 \\
\raisebox{-0.2ex}{\includegraphics[height=1em]{logos/claude-ai-icon.pdf}}~Claude Opus 4.7 & 15.00 & 27.00 & 23.00 & 34.00 & 94.33 & 81.83 & 0.00 & 80.00 & 44.40 \\
\raisebox{-0.2ex}{\includegraphics[height=1em]{logos/gemini-color.pdf}}~Gemini-3-Flash & 12.00 & 34.00 & 36.00 & 37.00 & 94.50 & 79.17 & 4.00 & 79.00 & 46.96 \\
\raisebox{-0.2ex}{\includegraphics[height=1em]{logos/openai.pdf}}~GPT-5.4-Mini & 14.00 & 31.00 & 25.00 & 46.00 & 93.33 & 76.50 & 2.00 & 87.00 & 46.85 \\
\raisebox{-0.2ex}{\includegraphics[height=1em]{logos/openai.pdf}}~GPT-5.4-Nano & 9.00 & 45.00 & 23.00 & 38.00 & 94.50 & 73.50 & 3.00 & 62.00 & 43.50 \\
\raisebox{-0.2ex}{\includegraphics[height=1em]{logos/claude-ai-icon.pdf}}~Claude Sonnet 4.6 & 3.00 & 36.00 & 17.00 & 31.00 & 89.00 & 72.67 & 12.00 & 29.00 & 36.21 \\
\raisebox{-0.2ex}{\includegraphics[height=1em]{logos/claude-ai-icon.pdf}}~Claude 4.5 Haiku & 0.00 & 24.00 & 22.00 & 43.00 & 54.17 & 45.50 & 5.00 & 13.00 & 25.83 \\
\midrule
\multicolumn{10}{l}{\textit{Open-source / Local Models}} \\
\midrule
\raisebox{-0.2ex}{\includegraphics[height=1em]{logos/qwen-color.pdf}}~Qwen3.6-35B-A3B & 1.00 & 33.00 & 27.00 & 24.00 & 93.50 & 77.67 & 3.00 & 80.00 & 42.40 \\
\raisebox{-0.2ex}{\includegraphics[height=1em]{logos/qwen-color.pdf}}~Qwen3.5-4B & 16.00 & 44.00 & 28.00 & 25.00 & 88.67 & 61.00 & 5.00 & 53.00 & 40.08 \\
\raisebox{-0.2ex}{\includegraphics[height=1em]{logos/qwen-color.pdf}}~Qwen3.5-2B & 8.00 & 31.00 & 28.00 & 25.00 & 84.00 & 51.67 & 10.00 & 34.00 & 33.96 \\
\raisebox{-0.2ex}{\includegraphics[height=1em]{logos/qwen-color.pdf}}~Qwen3.5-0.8B & 0.00 & 41.00 & 27.00 & 40.00 & 50.50 & 49.83 & 9.00 & 45.00 & 32.79 \\
\bottomrule
\end{tabular}
}
\end{table*}

\begin{table*}[!ht]
\centering
\small
\caption{Per-capability model performance on the SpatialArm benchmark. Columns LC1--LC9 denote object inventory, count and set reasoning, 3D spatial relation, distance/proximity reasoning, order and region reasoning, robot end-effector action, temporal trajectory, task semantic binding, and multi-view consistency, respectively. Scores are reported in percentage, and AVG denotes the average score across all capability dimensions. Models are grouped into closed-source/API models and open-source/local models.}
\label{tab:libero_capability_perf}
\resizebox{\textwidth}{!}{
\begin{tabular}{lcccccccccc}
\toprule
\textbf{Model} & \textbf{LC1} & \textbf{LC2} & \textbf{LC3} & \textbf{LC4} & \textbf{LC5} & \textbf{LC6} & \textbf{LC7} & \textbf{LC8} & \textbf{LC9} & \textbf{AVG} \\
\midrule
\multicolumn{11}{l}{\textit{Closed-source / API Models}} \\
\midrule
\raisebox{-0.2ex}{\includegraphics[height=1em]{logos/openai.pdf}}~GPT-5.5 & 88.00 & 75.00 & 66.00 & 68.00 & 67.00 & 70.00 & 84.00 & 63.00 & 66.00 & 71.89 \\
\raisebox{-0.2ex}{\includegraphics[height=1em]{logos/claude-ai-icon.pdf}}~Claude Opus 4.7 & 83.00 & 92.00 & 62.00 & 70.00 & 67.00 & 71.00 & 60.00 & 62.00 & 52.00 & 68.78 \\
\raisebox{-0.2ex}{\includegraphics[height=1em]{logos/gemini-color.pdf}}~Gemini-3-Flash & 86.00 & 90.00 & 56.00 & 73.00 & 60.00 & 65.00 & 53.00 & 63.00 & 63.00 & 67.67 \\
\raisebox{-0.2ex}{\includegraphics[height=1em]{logos/openai.pdf}}~GPT-5.4-Mini & 78.00 & 81.00 & 65.00 & 67.00 & 49.00 & 58.00 & 58.00 & 64.00 & 20.00 & 60.00 \\
\raisebox{-0.2ex}{\includegraphics[height=1em]{logos/openai.pdf}}~GPT-5.4-Nano & 70.00 & 67.00 & 46.00 & 65.00 & 41.00 & 51.00 & 44.00 & 57.00 & 33.00 & 52.67 \\
\raisebox{-0.2ex}{\includegraphics[height=1em]{logos/claude-ai-icon.pdf}}~Claude Sonnet 4.6 & 63.00 & 71.00 & 66.00 & 59.00 & 16.00 & 39.00 & 35.00 & 51.00 & 18.00 & 46.44 \\
\raisebox{-0.2ex}{\includegraphics[height=1em]{logos/claude-ai-icon.pdf}}~Claude 4.5 Haiku & 51.00 & 41.00 & 30.00 & 72.00 & 2.00 & 31.00 & 35.00 & 34.00 & 31.00 & 36.33 \\
\midrule
\multicolumn{11}{l}{\textit{Open-source / Local Models}} \\
\midrule
\raisebox{-0.2ex}{\includegraphics[height=1em]{logos/qwen-color.pdf}}~Qwen3.6-35B-A3B & 73.00 & 77.00 & 71.00 & 71.00 & 44.00 & 45.00 & 43.00 & 57.00 & 5.00 & 54.00 \\
\raisebox{-0.2ex}{\includegraphics[height=1em]{logos/qwen-color.pdf}}~Qwen3.5-4B & 64.00 & 57.00 & 47.00 & 52.00 & 31.00 & 48.00 & 35.00 & 64.00 & 27.00 & 47.22 \\
\raisebox{-0.2ex}{\includegraphics[height=1em]{logos/qwen-color.pdf}}~Qwen3.5-2B & 56.00 & 46.00 & 39.00 & 38.00 & 24.00 & 24.00 & 18.00 & 61.00 & 26.00 & 36.89 \\
\raisebox{-0.2ex}{\includegraphics[height=1em]{logos/qwen-color.pdf}}~Qwen3.5-0.8B & 34.00 & 32.00 & 16.00 & 35.00 & 1.00 & 33.00 & 19.00 & 28.00 & 49.00 & 27.44 \\
\bottomrule
\end{tabular}
}
\end{table*}

\begin{table*}[!ht]
\centering
\small
\caption{Per-capability model performance on the SpatialQuadruped benchmark. Columns C01--C08 denote semantic category visibility, semantic region area comparison, image-region semantic localization, region depth and near-far ordering, semantic-category/depth joint reasoning, multi-view semantic consistency, trajectory order and path structure, and motion trend/orientation change, respectively. Scores are reported in percentage, and AVG denotes the average score across all capability dimensions. Models are grouped into closed-source/API models and open-source/local models.}
\label{tab:dog_capability_perf}
\resizebox{\textwidth}{!}{
\begin{tabular}{lccccccccc}
\toprule
\textbf{Model} & \textbf{C01} & \textbf{C02} & \textbf{C03} & \textbf{C04} & \textbf{C05} & \textbf{C06} & \textbf{C07} & \textbf{C08} & \textbf{AVG} \\
\midrule
\multicolumn{10}{l}{\textit{Closed-source / API Models}} \\
\midrule
\raisebox{-0.2ex}{\includegraphics[height=1em]{logos/openai.pdf}}~GPT-5.5 & 57.00 & 51.00 & 52.00 & 60.00 & 46.00 & 37.00 & 61.00 & 47.00 & 51.38 \\
\raisebox{-0.2ex}{\includegraphics[height=1em]{logos/claude-ai-icon.pdf}}~Claude Opus 4.7 & 51.00 & 48.00 & 54.00 & 57.00 & 40.00 & 43.00 & 59.00 & 72.00 & 53.00 \\
\raisebox{-0.2ex}{\includegraphics[height=1em]{logos/gemini-color.pdf}}~Gemini-3-Flash & 48.00 & 47.00 & 46.00 & 56.00 & 40.00 & 43.00 & 54.00 & 77.00 & 51.38 \\
\raisebox{-0.2ex}{\includegraphics[height=1em]{logos/openai.pdf}}~GPT-5.4-Mini & 44.00 & 40.00 & 46.00 & 39.00 & 36.00 & 38.00 & 50.00 & 47.00 & 42.50 \\
\raisebox{-0.2ex}{\includegraphics[height=1em]{logos/openai.pdf}}~GPT-5.4-Nano & 36.00 & 42.00 & 45.00 & 38.00 & 30.00 & 23.00 & 35.00 & 36.00 & 35.63 \\
\raisebox{-0.2ex}{\includegraphics[height=1em]{logos/claude-ai-icon.pdf}}~Claude Sonnet 4.6 & 46.00 & 43.00 & 51.00 & 51.00 & 35.00 & 41.00 & 21.00 & 28.00 & 39.50 \\
\raisebox{-0.2ex}{\includegraphics[height=1em]{logos/claude-ai-icon.pdf}}~Claude 4.5 Haiku & 39.00 & 29.00 & 48.00 & 37.00 & 26.00 & 16.00 & 20.00 & 15.00 & 28.75 \\
\midrule
\multicolumn{10}{l}{\textit{Open-source / Local Models}} \\
\midrule
\raisebox{-0.2ex}{\includegraphics[height=1em]{logos/qwen-color.pdf}}~Qwen3.6-35B-A3B & 50.00 & 48.00 & 64.00 & 59.00 & 44.00 & 45.00 & 47.00 & 43.00 & 50.00 \\
\raisebox{-0.2ex}{\includegraphics[height=1em]{logos/qwen-color.pdf}}~Qwen3.5-4B & 37.00 & 35.00 & 32.00 & 24.00 & 24.00 & 18.00 & 24.00 & 25.00 & 27.38 \\
\raisebox{-0.2ex}{\includegraphics[height=1em]{logos/qwen-color.pdf}}~Qwen3.5-2B & 27.00 & 35.00 & 32.00 & 17.00 & 17.00 & 18.00 & 21.00 & 12.00 & 22.38 \\
\raisebox{-0.2ex}{\includegraphics[height=1em]{logos/qwen-color.pdf}}~Qwen3.5-0.8B & 21.00 & 21.00 & 39.00 & 34.00 & 23.00 & 17.00 & 21.00 & 9.00 & 23.13 \\
\bottomrule
\end{tabular}
}
\end{table*}

\begin{table*}[!ht]
\centering
\small
\caption{Per-capability model performance on the SpatialUAV benchmark. Columns C01, C02, C03, C04, C05, C06, C07, C08, and C11 denote visible category recognition, instance counting and category distribution, image-plane location, visible area/scale reasoning, depth reasoning, camera-frame 3D position and scale, box geometry and image-boundary relation, category-constrained object selection/matching, and image-level aspect geometry, respectively. Scores are reported in percentage, and AVG denotes the average score across all capability dimensions. Models are grouped into closed-source/API models and open-source/local models.}
\label{tab:uav_capability_perf}
\resizebox{\textwidth}{!}{
\begin{tabular}{lcccccccccc}
\toprule
\textbf{Model} & \textbf{C01} & \textbf{C02} & \textbf{C03} & \textbf{C04} & \textbf{C05} & \textbf{C06} & \textbf{C07} & \textbf{C08} & \textbf{C11} & \textbf{AVG} \\
\midrule
\multicolumn{11}{l}{\textit{Closed-source / API Models}} \\
\midrule
\raisebox{-0.2ex}{\includegraphics[height=1em]{logos/openai.pdf}}~GPT-5.5 & 48.00 & 49.00 & 79.00 & 45.00 & 54.00 & 74.00 & 66.00 & 74.00 & 85.00 & 63.78 \\
\raisebox{-0.2ex}{\includegraphics[height=1em]{logos/claude-ai-icon.pdf}}~Claude Opus 4.7 & 52.00 & 49.00 & 93.00 & 60.00 & 59.00 & 78.00 & 86.00 & 74.00 & 22.00 & 63.67 \\
\raisebox{-0.2ex}{\includegraphics[height=1em]{logos/gemini-color.pdf}}~Gemini-3-Flash & 58.00 & 42.00 & 63.00 & 33.00 & 47.00 & 62.00 & 65.00 & 47.00 & 22.00 & 49.89 \\
\raisebox{-0.2ex}{\includegraphics[height=1em]{logos/openai.pdf}}~GPT-5.4-Mini & 60.00 & 38.00 & 53.00 & 36.00 & 43.00 & 56.00 & 54.00 & 42.00 & 20.00 & 45.78 \\
\raisebox{-0.2ex}{\includegraphics[height=1em]{logos/openai.pdf}}~GPT-5.4-Nano & 50.00 & 36.00 & 39.00 & 45.00 & 27.00 & 48.00 & 51.00 & 43.00 & 72.00 & 45.67 \\
\raisebox{-0.2ex}{\includegraphics[height=1em]{logos/claude-ai-icon.pdf}}~Claude Sonnet 4.6 & 49.00 & 30.00 & 43.00 & 36.00 & 39.00 & 62.00 & 47.00 & 41.00 & 34.00 & 42.33 \\
\raisebox{-0.2ex}{\includegraphics[height=1em]{logos/claude-ai-icon.pdf}}~Claude 4.5 Haiku & 47.00 & 32.00 & 28.00 & 25.00 & 24.00 & 50.00 & 29.00 & 32.00 & 45.00 & 34.67 \\
\midrule
\multicolumn{11}{l}{\textit{Open-source / Local Models}} \\
\midrule
\raisebox{-0.2ex}{\includegraphics[height=1em]{logos/qwen-color.pdf}}~Qwen3.6-35B-A3B & 53.00 & 43.00 & 58.00 & 52.00 & 57.00 & 64.00 & 64.00 & 48.00 & 47.00 & 54.00 \\
\raisebox{-0.2ex}{\includegraphics[height=1em]{logos/qwen-color.pdf}}~Qwen3.5-4B & 40.00 & 36.00 & 28.00 & 20.00 & 33.00 & 48.00 & 38.00 & 25.00 & 21.00 & 32.11 \\
\raisebox{-0.2ex}{\includegraphics[height=1em]{logos/qwen-color.pdf}}~Qwen3.5-2B & 38.00 & 26.00 & 31.00 & 21.00 & 22.00 & 47.00 & 32.00 & 21.00 & 80.00 & 35.33 \\
\raisebox{-0.2ex}{\includegraphics[height=1em]{logos/qwen-color.pdf}}~Qwen3.5-0.8B & 41.00 & 15.00 & 24.00 & 24.00 & 24.00 & 38.00 & 31.00 & 25.00 & 41.00 & 29.22 \\
\bottomrule
\end{tabular}
}
\end{table*}

\begin{table*}[!ht]
\centering
\small
\setlength{\tabcolsep}{5.0pt}
\renewcommand{\arraystretch}{1.10}
\caption{Specification of the auxiliary regularized 2PL diagnosis. The same
scale, thresholds, and optimization settings are applied to all six OE-Track
benchmarks and the Table~8 ablation variants.}
\label{tab:irt_protocol}
\begin{tabular}{@{}p{0.16\textwidth}p{0.38\textwidth}p{0.39\textwidth}@{}}
\toprule
\textbf{Quantity or step} & \textbf{Statistical interpretation} & \textbf{Definition or setting} \\
\midrule
Model ability \(\theta_m\)
& Relative capability of model \(m\), with larger values indicating stronger
benchmark performance.
& Standardized after every update to mean \(0\) and variance \(1\), then
oriented to agree with raw mean scores. Ability updates are bounded to
\([-6,6]\). \\
Item discrimination \(a_i\)
& Sensitivity of item \(i\) to differences in model ability; a larger positive
slope indicates greater discriminative capacity.
& \(a_i<0\): reversed; \(0\leq a_i<0.3\): low;
\(0.3\leq a_i<0.7\): moderate; \(a_i\geq0.7\): High-Disc. \\
Item difficulty \(b_i\)
& Ability level at which the expected item score reaches \(0.5\).
& \(-2\leq b_i\leq2\) is the central range. \(b_i=-\beta_i/a_i\) is omitted
when \(|a_i|<0.05\), because an almost-flat curve has no stable crossing point. \\
Regularization
& Constrains unstable parameter estimates while preserving meaningful
difficulty estimates and model ordering.
& \(\lambda_a=1.0\), \(\lambda_\beta=0.05\), and
\(\lambda_\theta=0.05\). \\
Optimization
& Alternating estimation of item parameters and model abilities.
& Initialization uses mean-response logits and item--ability correlations.
Four alternating updates; L-BFGS-B for \((a_i,\beta_i)\), at most 200
iterations and tolerance \(10^{-9}\); bounded scalar minimization for
\(\theta_m\), tolerance \(10^{-5}\). \\
Stability
& Sensitivity of the High-Disc. proportion to the composition of the model
evaluation panel.
& 1,000 joint model-bootstrap replicates with seed 20260731; 11 model
identities are sampled with replacement and the full 2PL model is refitted. \\
Information curve
& Total measurement information and its concentration across the ability
scale.
& \(\theta\in[-4,4]\) at increments of \(0.05\); item information is summed
within each 10,000-item benchmark. \\
\bottomrule
\end{tabular}
\end{table*}

\begin{table*}[!ht]
\centering
\small
\setlength{\tabcolsep}{4.8pt}
\renewcommand{\arraystretch}{1.08}
\caption{Item-discrimination stability on the six 10,000-item OE-Track
benchmarks. High-Disc. is the proportion with \(a_i\geq0.7\); \(D_i\) is the
top-three minus bottom-three score gap; \(\rho_S\) measures agreement between
the two discrimination rankings; and \(\theta_{\mathrm{peak}}\) is the ability
at which the benchmark provides maximum information. Red entries are
data-anchored estimates retained for author verification.}
\label{tab:irt_stability}
\begin{tabular}{@{}lrrrrrr@{}}
\toprule
\textbf{Benchmark}
& \textbf{Items}
& \textbf{High-Disc.}
& \textbf{Bootstrap 95\% CI}
& \(\mathbf{Mean\ }D_i\)
& \(\boldsymbol{\rho_S(a_i,D_i)}\)
& \(\boldsymbol{\theta_{\mathrm{peak}}}\) \\
\midrule
SHT & 10,000 & 36.1\% & [21.0, 38.1]\% & 0.150 & 0.929 & -0.25 \\
SHN & 10,000 & 42.9\% & [25.3, 49.1]\% & 0.252 & 0.906 & -0.10 \\
SD  & 10,000 & 36.0\% & [22.6, 39.3]\% & 0.153 & 0.913 & -0.75 \\
SQ  & 10,000 & 43.3\% & [28.7, 45.1]\% & 0.269 & 0.891 & -0.10 \\
SU  & 10,000 & 48.8\% & [28.7, 52.7]\% & 0.285 & 0.845 & -0.35 \\
SA  & 10,000 & 47.9\% & [30.8, 57.6]\% & 0.361 & 0.925 & -0.55 \\
\midrule
\textbf{OE-Track pooled}
& \textbf{60,000}
& \textbf{42.5\%}
& \textbf{[31.3, 43.7]\%}
& \textbf{0.245}
& \textbf{0.906}
& -- \\
\bottomrule
\end{tabular}
\end{table*}

\paragraph{Item-discrimination results.}
The pooled estimate indicates that 42.5
discriminative. The corresponding proportions exceed 35\% for all six
benchmarks and 40\% for four benchmarks. SpatialUAV reaches 48.8\%, which is
consistent with the 49.4\% Full Process Quality Control result reported for
SpatialUAV in Table~8. These results extend the main-text evidence obtained on
SpatialUAV to the heterogeneous settings represented by the OE-Track
benchmarks. Moreover, all mean \(D_i\) values are positive, and the six rank
correlations range from 0.845 to 0.929, indicating substantial agreement
between the discriminative-item rankings produced by the 2PL model and the
direct top--bottom comparison.

\paragraph{Benchmark information analysis.}
Item \(i\)'s Fisher information at ability \(\theta\), and the total
information of benchmark \(B\), are
\begin{equation}
  I_i(\theta)
  =a_i^2p_i(\theta)\bigl(1-p_i(\theta)\bigr),
  \qquad
  I_B(\theta)=\sum_{i\in\mathcal{I}_B}I_i(\theta).
  \label{eq:test_information}
\end{equation}
Larger \(I_B(\theta)\) means more precise model separation at that ability
level, with conditional standard error
\(\mathrm{SE}_B(\theta)=I_B(\theta)^{-1/2}\).

\paragraph{Interpretation of the benchmark information curves.}
Each color denotes one benchmark. The horizontal axis represents standardized
model ability: \(\theta=0\) corresponds to the evaluation-panel mean, whereas
negative and positive values indicate lower and higher relative ability,
respectively. Panel (a) retains the absolute sum defined in
Eq.~\eqref{eq:test_information}; curve height therefore permits comparison of
total measurement precision across benchmarks. Panel (b) normalizes each curve
by its maximum, such that every peak equals \(1\), thereby identifying the
ability region in which each benchmark is most informative independently of
its total magnitude. The shaded interval \([-2,2]\) denotes the central ability
range, the dashed line indicates the panel mean, and the open circles denote
the \(\theta_{\mathrm{peak}}\) values reported in
Table~\ref{tab:irt_stability}.

\begin{figure*}[!ht]
\centering
\includegraphics[width=\textwidth]{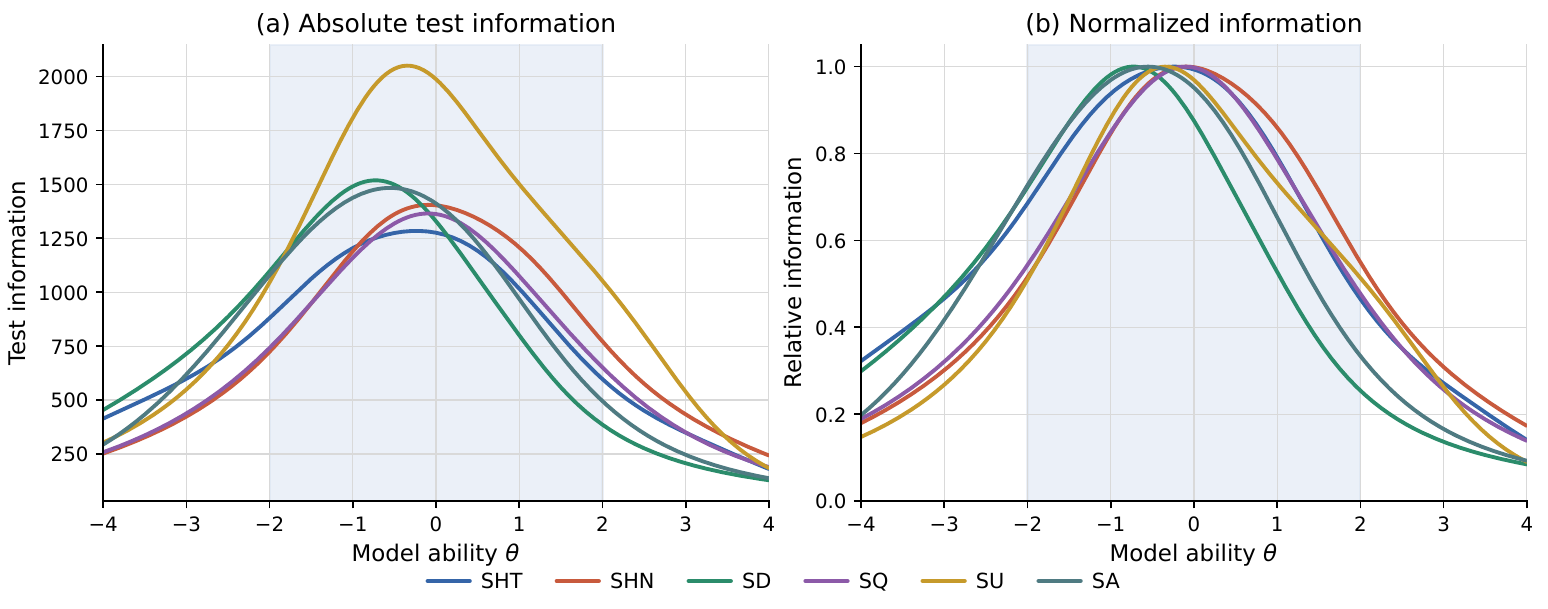}
\caption{Benchmark-level information curves for the six 10,000-item OE-Track
benchmarks. Panel (a) reports the absolute measurement information provided by
each benchmark; higher curves imply smaller conditional measurement error.
Panel (b) reports the distribution of within-benchmark normalized information
across the ability scale after removing magnitude differences. All peaks fall
between
\(\theta=-0.75\) and \(-0.10\), while every benchmark retains information
throughout the central \([-2,2]\) range.}
\label{fig:irt_information_curves}
\end{figure*}



\end{document}